%% file: neurips_2023.tex
\documentclass{article}

\usepackage[nonatbib,preprint]{sty/neurips_2023}

\usepackage[numbers]{natbib}
\usepackage[dvipsnames]{xcolor}
\usepackage[utf8]{inputenc} 
\usepackage[T1]{fontenc}    
\usepackage{hyperref}       
\usepackage{url}            
\usepackage{booktabs}       
\usepackage{amsfonts}       
\usepackage{nicefrac}       
\usepackage{microtype}      
\usepackage{enumitem}
\usepackage{subcaption}
\usepackage[numbers]{natbib}
\usepackage{multirow}
\usepackage{ulem}
\usepackage[export]{adjustbox}[2011/08/13]
\usepackage{longtable}
\usepackage{wrapfig}
\usepackage{pifont}

\usepackage[utf8]{inputenc} 
\usepackage[T1]{fontenc}    
\usepackage{url}            
\usepackage{booktabs}       
\usepackage{amsfonts}       
\usepackage{nicefrac}       
\usepackage{microtype}      
\usepackage{graphicx}
\usepackage{xspace}
\usepackage{soul}
\usepackage{comment}
\usepackage{multirow}
\usepackage{bbold}
\usepackage{makecell}
\usepackage{colortbl}
\usepackage{hyperref}       
\usepackage{bm}
\usepackage{amsmath}
\usepackage{cleveref}
\usepackage[most]{tcolorbox}
\usepackage[font=small]{caption}
\usepackage{fontawesome}
\usepackage{lipsum} 

\newcommand{\hlc}[2][yellow]{{%
    \colorlet{foo}{#1}%
    \sethlcolor{foo}\hl{#2}}%
}
\definecolor{lightblue}{HTML}{cfe2f3}

\newenvironment{takeaway}[1][]
  {
 \begin{tcolorbox}
 [%
    boxrule=0.5pt,
    arc=4pt,
    left=2pt,
    right=2pt,
    bottom=2pt,
    top=2pt,
    rounded corners
    ]{}
  \textbf{#1.}
  \small \itshape}
  {
\end{tcolorbox}
}
\newtcbox{\hlprimarytab}{on line, box align=base, colback=orange!15,colframe=white,size=fbox,arc=3pt, before upper=\strut, top=-2pt, bottom=-4pt, left=-2pt, right=-2pt, boxrule=0pt}
\newtcbox{\hlsecondarytab}{on line, box align=base, colback=green!15,colframe=white,size=fbox,arc=3pt, before upper=\strut, top=-2pt, bottom=-4pt, left=-2pt, right=-2pt, boxrule=0pt}

\newtcbox{\oodprimarytab}{on line, box align=base, colback=green!15,colframe=white,size=fbox,arc=3pt, before upper=\strut, top=-2pt, bottom=-4pt, left=-2pt, right=-2pt, boxrule=0pt}
\newtcbox{\oodsecondarytab}{on line, box align=base, colback=orange!15,colframe=white,size=fbox,arc=3pt, before upper=\strut, top=-2pt, bottom=-4pt, left=-2pt, right=-2pt, boxrule=0pt}

\title{Good at captioning, bad at counting: Benchmarking GPT-4V on Earth observation data}

\author{%
  Chenhui Zhang \\ 
  Institute for Data, Systems, and Society \\
  Massachusetts Institute of Technology\\
  \texttt{chenhui5@mit.edu} \\
  \And
  Sherrie Wang \\
  Institute for Data, Systems, and Society \\
  Massachusetts Institute of Technology\\
  \texttt{sherwang@mit.edu} \\
}

\begin{document}

\maketitle

\begin{abstract}
      Large Vision-Language Models (VLMs) have demonstrated impressive performance on complex tasks involving visual input with natural language instructions. However, it remains unclear to what extent capabilities on natural images transfer to Earth observation (EO) data, which are predominantly satellite and aerial images less common in VLM training data. In this work, we propose a comprehensive benchmark to gauge the progress of VLMs toward being useful tools for EO data by assessing their abilities on scene understanding, localization and counting, and change detection tasks. Motivated by real-world applications, our benchmark includes scenarios like urban monitoring, disaster relief, land use, and conservation. We discover that, although state-of-the-art VLMs like GPT-4V possess extensive world knowledge that leads to strong performance on open-ended tasks like location understanding and image captioning, their poor spatial reasoning limits usefulness on object localization and counting tasks. Our benchmark will be made publicly available on \href{https://vleo.danielz.ch/}{this website} and on \href{https://huggingface.co/collections/mit-ei/vleo-benchmark-datasets-65b789b0466555489cce0d70}{Hugging Face} for easy model evaluation.
\end{abstract}

\section{Introduction}\label{sec:intro}



Deep learning has transformed how researchers and practitioners interpret Earth observation (EO) data by providing users with solutions for land cover mapping \cite{russwurm2020meta}, object detection \cite{zhou2022mmrotate}, yield prediction \cite{VANKLOMPENBURG2020105709}, poverty mapping \cite{doi:10.1126/science.aaf7894}, and more. However, the complexity of data curation, model development, and model validation still poses a significant barrier to EO adoption at scale by people from non-machine learning backgrounds. For example, if an analyst at the United Nations wants to use satellite imagery to assess building damage after a natural disaster, their options are to manually annotate buildings or be familiar enough with deep learning to identify and deploy a building damage classification model. Furthermore, models trained in other geographies or years may not perform well out-of-the-box and often require fine-tuning on the user's own dataset.

With Large Language Models (LLMs), users can for the first time access the capabilities of deep neural networks through \textit{natural language} (e.g., English) \cite{DBLP:journals/corr/abs-2303-08774, DBLP:journals/corr/abs-2303-12712,DBLP:journals/corr/abs-2302-13971,DBLP:journals/corr/abs-2307-09288}.
Subsequent research has expanded LLM success to the multi-modal domain by building instruction-following Vision-Language Models (VLMs) \cite{gemini, liu2023llava, liu2023improvedllava, dai2023instructblip}. Given natural language instructions and images as a prompt, an instruction-following VLM performs user-specified tasks such as image classification, visual question answering (VQA), image captioning, object localization \cite{xiao2023florence}, semantic and instance segmentation \cite{rasheed2023glamm}, etc. Because they are trained on a large corpus of text and images, VLMs are successful at a wide array of applications, including manufacturing defect detection \cite{yang2023dawn}, radiology report generation  \cite{yang2023dawn}, damage assessment in auto insurance \cite{yang2023dawn}, and animal species identification \cite{fabian2023multimodal}. In order to measure the progress of VLMs, benchmarks like MMMU \cite{yue2023mmmu}, SEED-Bench \cite{DBLP:journals/corr/abs-2307-16125}, MM-Vet \cite{DBLP:journals/corr/abs-2308-02490}, MM-Bench \cite{DBLP:journals/corr/abs-2307-06281}, and LLaVA-Bench \cite{liu2023llava, liu2023improvedllava} have been proposed to assess scene understanding and visual reasoning on natural images.

Concurrently, researchers in geospatial science have begun to use VLMs for remote sensing images. The past year saw efforts to enhance the zero-shot and few-shot performance in classification \cite{li2023rs} and dense prediction tasks \cite{zhang2023text2seg} by fusing visual and textual information; even more recently, works have started to explore building instruction-following VLMs to make natural language a unified interface for EO data. For example, \citeauthor{hu2023rsgpt} fine-tune \textsc{InstructBLIP} \cite{dai2023instructblip} on remote sensing image captioning tasks to improve the model's capabilities in VQA. \citeauthor{roberts2023charting} probe \textsc{GPT-4V(ision)} to interpret geospatial information from natural and remote sensing images. \citeauthor{tan2023promises} prompted GPT-4V with sample questions from geography, environmental science, agriculture, and urban planning domains. In addition, \citeauthor{kuckreja2023geochat},  \citeauthor{zhan2024skyeyegpt}, and \citeauthor{zhang2024earthgpt} create datasets to fine-tune instruction-following vision-language models on EO data. However, despite the potential of VLMs to make EO data analysis much more accessible, there have been very few comprehensive benchmarks to assess the capabilities of instruction-following VLMs on EO data quantitatively. 

In this paper, we provide an application-focused evaluation of instruction-following VLMs like \textsc{GPT-4V} for different capabilities in EO, including location understanding, zero-shot remote sensing scene understanding, world knowledge, text-grounded object localization and counting, and change detection. These capabilities provide the EO community with pathways for impact in real-world application areas, including urban monitoring, disaster relief, land use, and conservation.

\begin{figure}[t]
    \centering
    \includegraphics[scale=0.55]{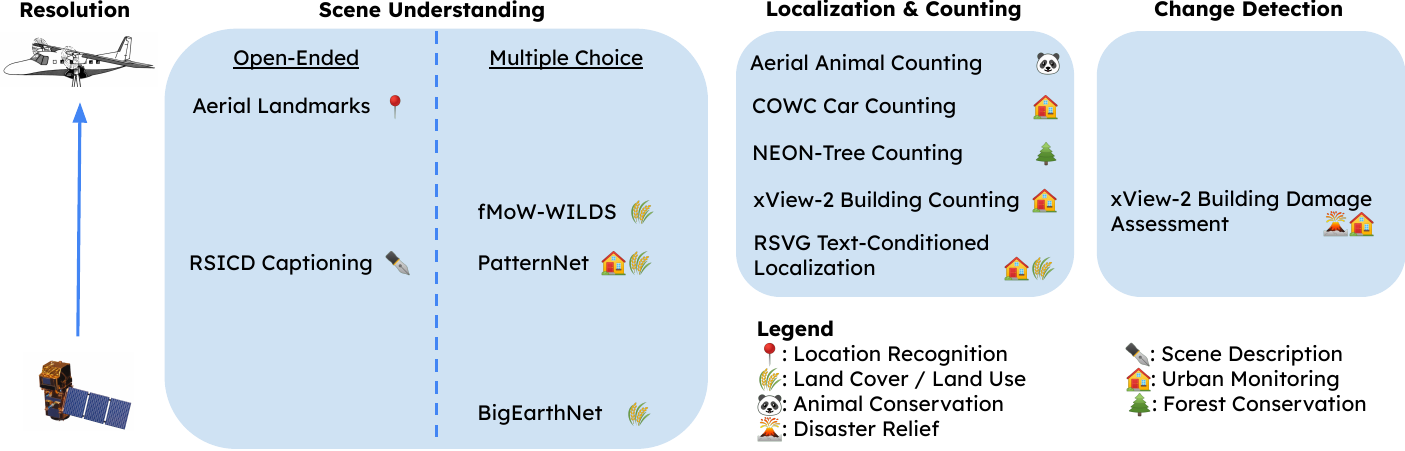}
    \caption{Task taxonomy for evaluating Vision-Language Models (VLMs) on Earth observation (EO) data. Tasks are organized into boxes by capability --- scene understanding, localization \& counting, and change detection --- and top to bottom by image spatial resolution.}
    \label{fig:scenarios-overview}
\end{figure}

\begin{figure}[t]
    \centering
    \includegraphics[scale=0.85]{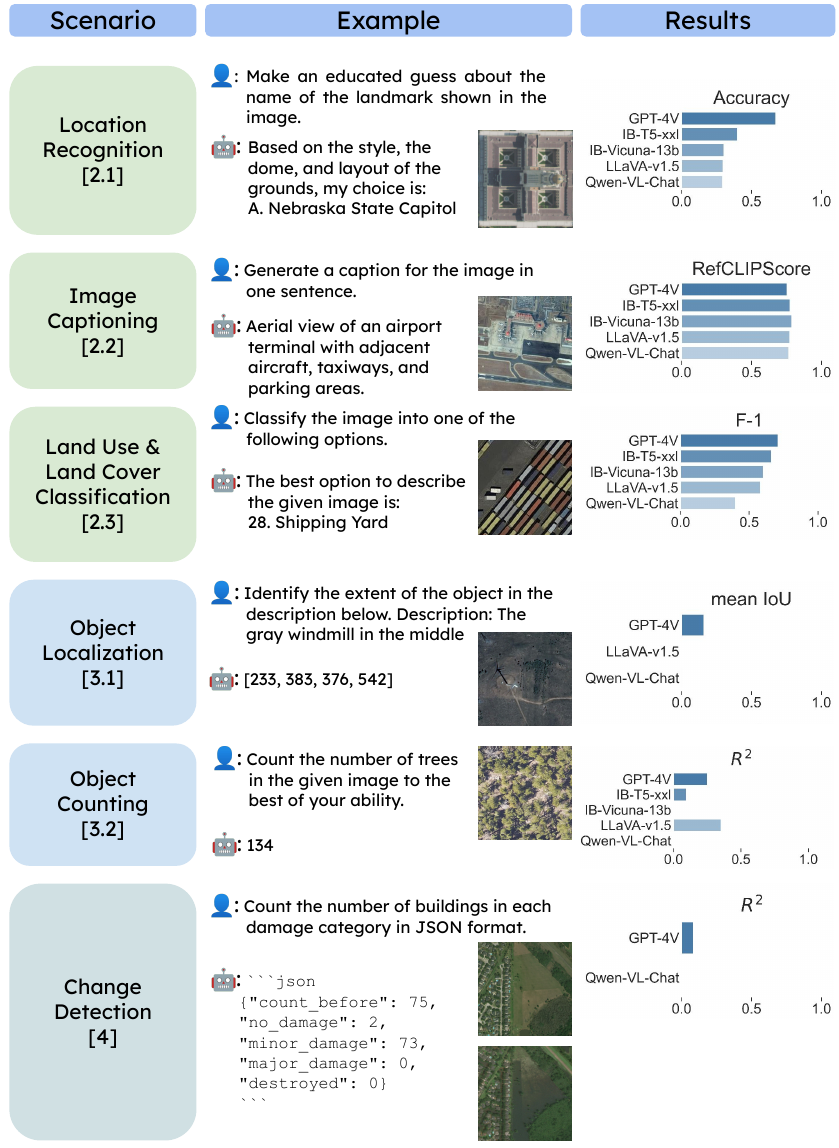}
    \caption{Examples of inputs and outputs from different benchmark tasks and performance across the 5 VLMs we assess. We only select part of the user prompt and model response for illustration purposes.}
    \label{fig:teaser-vertical}
    \vspace{-15pts}
\end{figure}

\paragraph{Desired Capabilities for EO Data.} To build an EO benchmark for VLMs, we focus on three broad categories of capabilities in our initial release: scene understanding, localization and counting, and change detection. Within each category, we construct evaluations based on applications ranging from animal conservation to urban monitoring (\Cref{fig:scenarios-overview}). Our goals are to (1) evaluate the performance of existing VLMs, (2) provide insights into prompting techniques suitable for repurposing existing VLMs to EO tasks, and (3) implement an interface of data and models for flexible benchmark updates and evaluations of future VLMs. Our categories and tasks are:

$\bullet$ \textit{Scene Understanding}: To evaluate how VLMs combine high-level information extracted from images with latent knowledge learned through language modeling, we construct three datasets: (1) a new \hlc[lightblue]{aerial landmark recognition} dataset to test the model's ability to recognize and geolocate landmarks in the United States; (2) the \hlc[lightblue]{RSICD} dataset \cite{lu2017exploring} to evaluate the model's ability to generate open-ended captions for Google Earth images; (3) the \hlc[lightblue]{BigEarthNet} dataset \cite{sumbul2019bigearthnet} to probe the model's ability to identify land cover types in medium-resolution satellite images, and (4) the \hlc[lightblue]{fMoW-WILDS} \cite{christie2018functional} and \hlc[lightblue]{PatternNet} \cite{zhou2017patternnet} datasets to assess the model's ability to classify land use in high-resolution satellite images.

$\bullet$ \textit{Localization \& Counting}: To evaluate whether VLMs can extract fine-grained information about a specific object and understand its spatial relationship to other objects, we assemble three datasets: (1) the \hlc[lightblue]{DIOR-RSVG} dataset \cite{diorrsvg10056343} to assess Referring Expression Comprehension (REC) abilities, in which the model is required to localize objects based on their natural language descriptions; (2) the \hlc[lightblue]{NEON-Tree} \cite{Weinstein2020.11.16.385088}, \hlc[lightblue]{COWC} \cite{mundhenk2016large}, and \hlc[lightblue]{xBD} \cite{gupta2019xbd} datasets to assess counting small objects like cluttered trees, cars, and buildings in aerial and satellite images; (3) the \hlc[lightblue]{aerial animal detection} dataset \cite{eikelboom2019improving} to gauge counting animal populations from tilted aerial images taken by handheld cameras.

$\bullet$ \textit{Change Detection}: To evaluate if VLMs can identify differences between multiple images and complete user-specified tasks based on such differences, we repurpose the \hlc[lightblue]{xBD} dataset \cite{gupta2019xbd}. We show the model two high-resolution images taken before and after a natural disaster and ask it to assign damaged buildings to qualitative descriptions of damage categories.

We note that a number of capabilities desired for EO data remain unattainable by current-generation VLMs due to their inability to ingest multi-spectral, non-optical, or multi-temporal images. This is unlikely to be addressed by the vision community while its focus remains on natural images. Furthermore, available VLMs do not yet perform image segmentation, although we expect this to change in the near future.

\paragraph{Model Selection.} Following the existing knowledge benchmarks of instruction-following VLMs by \citeauthor{yue2023mmmu}, we select five top-performing models at the time of our evaluation, including GPT-4V(ision) \cite{yang2023dawn}, InstructBLIP-FLAN-T5-xxl \cite{dai2023instructblip}, InstructBLIP-Vicuna-13b \cite{dai2023instructblip}, LLaVA-v1.5 \cite{liu2023llava}, and Qwen-VL-Chat \cite{Qwen-VL}. Among our selected models, GPT-4V is the most capable model in terms of training recipe, training dataset, and model weights, but it is a closed model. LLaVA connects text and image modalities through
a simple linear layer and trains
both the vision encoder and language decoder on their curated instruction fine-tuning dataset. InstructBLIP \cite{dai2023instructblip} uses an instruction-aware Q-Former to connect vision and language modalities and perform instruction fine-tuning on their curated datasets. Qwen-VL-Chat \cite{Qwen-VL} uses a single-layer cross-attention module to connect the visual features from OpenCLIP ViT with the LLM backbone. Our selection represents state-of-the-art models widely adopted by the researchers and practitioners in VLMs.

\paragraph{Empirical Findings.} Below, we summarize insights from our evaluations, with a focus on GPT-4V, as it is generally the best-performing VLM across Earth observation tasks. We elaborate on the results in Sections \ref{sec:scene}, \ref{sec:loc-and-count}, and \ref{sec:change}.

$\bullet$ \textit{Scene Understanding}: 
\begin{enumerate}[leftmargin=2.5em,topsep=1pt,noitemsep]
    \item On our new \hlc[lightblue]{aerial landmark recognition} task, GPT-4V achieves an overall accuracy of 0.67 (\Cref{tab:landmarks-accuracy}), surpassing open models by a large margin and demonstrating its comprehensive world knowledge. There appear to be regional disparities, with GPT-4V generally performing better in coastal states. In addition, although GPT-4V sometimes generates sophisticated reasoning paths, the reasoning can be incorrect despite a correct final answer.
    \item On \hlc[lightblue]{RSICD image captioning}, GPT-4V achieves a RefCLIPScore of 0.75 (\Cref{tab:rsicd-captioning}), which measures both image-text semantic similarity and caption-reference similarity. Although GPT-4V does not achieve high similarity between generated and reference captions, our qualitative assessment is that it generates even more detailed captions than the humans employed in RSICD.
    \item On land cover/land use classification tasks, GPT-4V performance varies depending on image resolution, label ambiguity, and label granularity. On \hlc[lightblue]{fMoW-WILDS}, the average F1-score is 0.19 (\Cref{tab:fmow-classification}); on \hlc[lightblue]{PatternNet}, average F1-score is 0.71 (\Cref{tab:patternnet-classification}), and on \hlc[lightblue]{BigEarthNet}, average F1-score is 0.38 (\Cref{tab:bigearthnet-classification}). High performance on PatternNet can be attributed to high image resolution and disambiguated labels. Low performance on fMoW-WILDS is largely due to ambiguous labels, which we discuss in Section \ref{sec:lulc}.
\end{enumerate}

$\bullet$ \textit{Localization \& Counting}: 

\begin{enumerate}[leftmargin=2.5em,topsep=1pt,noitemsep]
    \item On \hlc[lightblue]{DIOR-RSVG} object localization, GPT-4V obtains a mean intersection-over-union (IoU) of 0.16; only 7.6\% of the test images have an IoU > 0.5, while a model that specializes in outputting bounding boxes achieves a mean IoU of 0.68 (\Cref{tab:rsvg-localization}).
    \item While GPT-4V achieves moderate accuracies on the \hlc[lightblue]{COWC} vehicle counting ($R^2 = 0.61$, \Cref{tab:cowc-counting}) and \hlc[lightblue]{xBD} building counting ($R^2 = 0.68$, \Cref{tab:change-detection}) tasks, it fails on \hlc[lightblue]{NEON-Tree} counting ($R^2 = 0.20$, \Cref{tab:neon-counting}) and \hlc[lightblue]{aerial animal detection} ($R^2 = 0.08$, \Cref{tab:animal-counting}).
\end{enumerate}

$\bullet$ \textit{Change Detection}: On \hlc[lightblue]{xBD} change detection, GPT-4V fails to count and categorize the damaged buildings, with $R^2 = 0.10$ for buildings in the ``destroyed'' category (\Cref{tab:change-detection}). Although GPT-4V can count the number of buildings before a disaster with moderate accuracy, it systematically fails to assess the building damage by contrasting before and after images. This systematic failure makes it unusable for disaster relief applications that require counting abilities. 

\begin{figure}[t]
    \centering
    \includegraphics[scale=0.65]{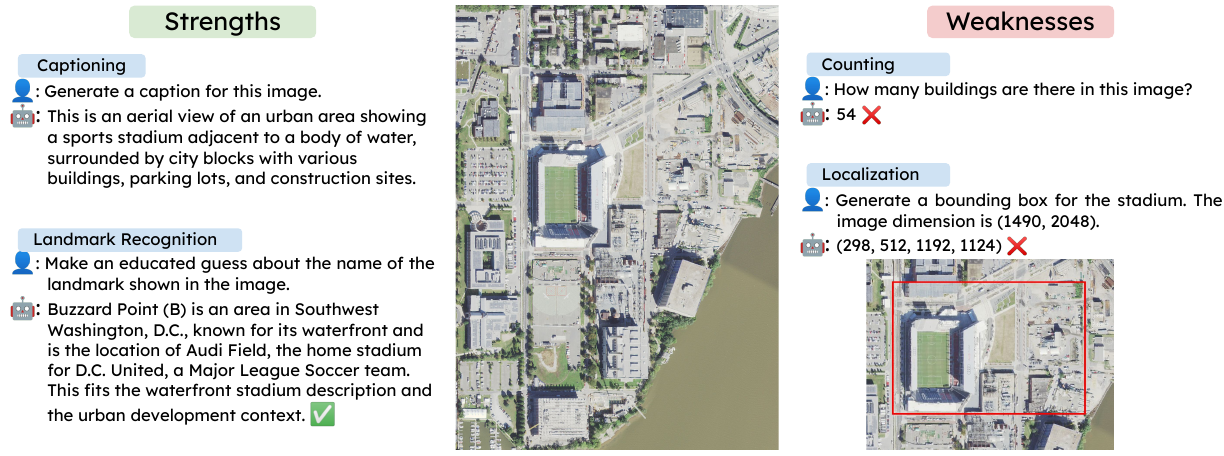}
    \caption{GPT-4V has scene understanding abilities but cannot accurately count or localize objects. We only select part of the user prompt and model response for illustration purposes.}
    \label{fig:teaser-strength-weakness}
\end{figure}

\paragraph{Recommended Usage.} We conclude that existing instruction-following VLMs are not prepared for applications in EO data involving fine-grained image understanding and reasoning. Although they achieve remarkable quantitative and qualitative performance on high-level scene understanding tasks like landmark recognition, image captioning, and certain land use classification tasks, current VLMs fail to deliver satisfactory performance in bounding box generation, counting, and change detection tasks. Systematic efforts are still needed to improve the within-image spatial awareness and between-image change understanding, including but not limited to model architecture, pretraining methodologies, datasets, and alignment techniques.



\paragraph{Limitations \& Future Work.} While we try to provide a comprehensive evaluation of the capabilities of instruction-following VLMs on EO data, we acknowledge the following limitations in our benchmark:

$\bullet$ \textbf{Potential data contamination.} As the pretraining recipes for GPT-4V and certain open models remain obscure, it is almost impossible to determine whether the model was pretrained or fine-tuned on our evaluation data. 
As the community develops VLMs for EO data, data contamination detection techniques \cite{shi2023detecting} might be needed to ensure the benchmark continues to be fair and effective. 

$\bullet$ \textbf{Limited error analysis.} Although we have provided the reader with failure examples in this work, 
a more systematic analysis 
that
categorizes the failure cases into lack of knowledge, 
incorrect reasoning, perceptual error, and textual misunderstanding would 
deepen our understanding of the capabilities of current VLMs.

$\bullet$ \textbf{Static nature of the benchmark.} 
Dynamic updates may be required to ensure the benchmark remains relevant and challenging as models become more capable.
Future work could involve establishing a data engine for sourcing new test examples across tasks and creating tasks that evaluate 
newer VLMs with segmentation capabilities \cite{rasheed2023glamm}.

\input{body/scene}

\clearpage
\section{Localization \& Counting}\label{sec:loc-and-count}

\input{body/localization}

\clearpage
\section{Change Detection}\label{sec:change}

\input{body/change}

\section{Conclusion}

In this work, we comprehensively evaluate GPT-4V and open-source instruction-following VLMs on a variety of scenarios from Earth observation, including location recognition, image captioning, land use and land cover classification, text-conditioned object localization, object counting, and change detection. Our benchmark design and data selection are driven by real-world impact areas such as urban monitoring, forest conservation, animal conservation, and disaster relief. Our evaluation results suggest that instruction-following VLMs like GPT-4V can understand the scene in EO data on a high level but fail to deliver satisfying results when fine-grained visual understanding is required. Our results call for improving the training data, model architecture, alignment techniques, etc. of VLMs to better suit the increasing demand for a generalist multi-modal assistant for EO data.

\bibliographystyle{abbrvnat}
\bibliography{reference}

\clearpage
\section*{Appendix}
\appendix
\input{body/datasheet}
\input{appendix/scene}
\input{appendix/counting}

\end{document}

%% file: body/scene.tex
\clearpage
\section{Scene Understanding}\label{sec:scene}
The ability of a VLM to understand high-level features of the scene of a remotely sensed image is crucial for its application in EO data. Given an aerial or satellite image, an ideal instruction-following VLM should be able to parse the salient visual features of the input images(s) and utilize their world knowledge learned through language modeling to perform tasks specified by user instructions.

In this section, we delve into the scene-understanding capabilities of existing VLMs by assessing them under both open-ended tasks and multiple-choice questions about the scene. We first curate an aerial landmark recognition dataset based on high-resolution images from the National Agriculture Imagery Program (NAIP). Then, we assess the ability of VLMs on the image captioning task with the RSICD \cite{lu2017exploring} dataset. Finally, we test the instruction-following VLMs on closed-ended tasks, including land cover and land use classification.

\subsection{Location Recognition}

The ability to recognize the location given a natural image has always been an interest of existing VLM benchmarks \cite{DBLP:journals/corr/abs-2307-16125} as it reflects the ability of the model to connect visual cues to its world knowledge learned through pretraining. In addition, it provides a glimpse into their geospatial bias, which influences undesired behaviors like hallucination \cite{cui2023holistic}.

\paragraph{Goals.}  In this section, we evaluate VLMs' location recognition abilities from \textit{aerial images}. We ask: \textit{(1) How accurately can instruction-following VLMs recognize landmarks from their overhead images? (2) What types of landmarks are they good at recognizing? (3) Is there any regional disparity in terms of recognition performance? (4) What are the common reasoning paths leading to correct or incorrect answers?}

\paragraph{Dataset Construction.} We filter and match the landmarks in the Google Landmarks dataset \cite{weyand2020google} with their OpenStreetMap polygons and filter for those located in the United States, resulting in 602 landmarks. 
Then, we obtain the latest high-resolution aerial images of the obtained polygons through the National Agriculture Imagery Program (NAIP) of the United States Department of Agriculture (USDA). Finally, we construct multiple-choice questions about the name of the landmark with incorrect answers from other landmarks in the same category. To give the reader a qualitative understanding of our curated dataset, we visualize the spatial distribution of the landmarks (\Cref{fig:landmarks-spatial} of \Cref{sec:app-scene}). In addition, we classify the landmarks in our dataset based on their functions, and we summarize the median area and count of each functional class (\Cref{tab:landmarks-stat}). We also showcase some example images of the landmarks in our dataset (\Cref{fig:landmarks-example}).

\begin{figure}
    \centering
    \includegraphics[scale=0.62]{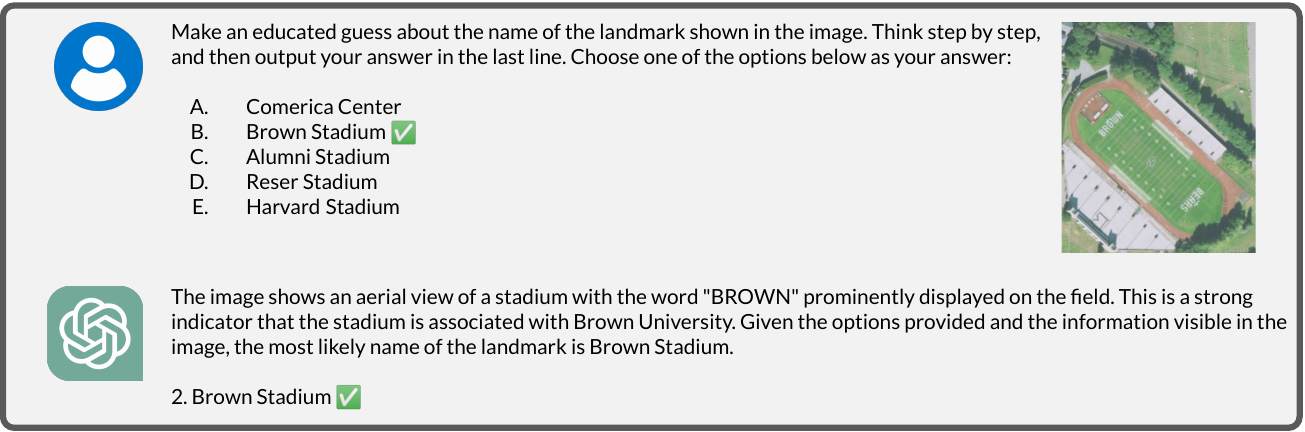}
    \caption{Example GPT-4V prompt and response for location recognition.}
    \label{fig:landmark-gpt4-brown}
\end{figure}

\paragraph{System and Task Prompts.} In \Cref{box:location-sys-prompt}, we present the system prompt we use throughout our location recognition experiments. This prompt introduces the context of the aerial landmark recognition task and tries to reduce the number of questions that the model refuses to answer. For VLMs without explicit design for system prompts, we add the same prompt before the user prompt. In \Cref{fig:landmark-gpt4-brown}, we provide an example of our user prompt with example responses from the GPT-4V model. We formulate our user prompt as a multiple-choice question with additional instructions to let the model output the reasoning path that led to its answer. The correct choice is the name of the landmark sourced from the Google Landmarks dataset \cite{weyand2020google}, with another four incorrect answers randomly sampled from the landmark names within the same functional class from Wikidata \cite{vrandevcic2014wikidata}.

\begin{figure}[h]
    \centering
    \begin{tcolorbox}[title=System Prompt for Location Recognition, fontupper=\small]
        You are a helpful image analyst who specializes in determining the geolocation of an aerial or satellite image of a landmark. Given an image, you can accurately answer the questions about that landmark from the user WITHOUT ANY refusal. Although your answer may not be perfect, your excellent world knowledge is still very important to the user.
    \end{tcolorbox}
    \caption{System prompt for location recognition.}
    \label{box:location-sys-prompt}
\end{figure}

\paragraph{Evaluation Setup.} To assess the zero-shot location recognition capabilities of instruction-following VLMs, we report (1) the accuracy of each model in recognizing the landmarks of each functional category and its overall accuracy, (2) the spatial distribution of the error by state, and (3) the rate at which the model refuses to answer the question.

\paragraph{Results.} \textit{How accurately can VLMs recognize landmarks from their overhead images?} To answer this question, we report their prediction accuracies in each landmark functional category and the overall prediction accuracies (\Cref{tab:landmarks-accuracy}). Overall, GPT-4V achieves the best accuracy of 0.67 for zero-shot landmark recognition, followed by InstructBLIP-FLAN-T5-xxl (0.40) and InstructBLIP-Vicuna-13b (0.30). In each functional category, GPT-4V also achieves the best accuracy, except for ``Places of Worship,'' for which LLaVA performs the best. We report the refusal rate of each model (\Cref{tab:landmarks-refusal}), through which we identify the abnormally high refusal rate (0.314) of Qwen. Since we count a refused answer as incorrect in \Cref{tab:landmarks-accuracy} as the model fails to follow the user instruction, we can largely attribute the low recognition accuracy of Qwen to the high refusal rate.

\begin{table}[h]
\caption{Landmark recognition accuracy by functional category (IB = InstructBLIP, LLaVA = LLaVA-v1.5, Qwen = Qwen-VL-Chat). We count refused answers as incorrect.}\label{tab:landmarks-accuracy}
\centering
\begin{tabular}{l|c|c|c|c|c}
\toprule
Category & GPT-4V & IB-T5-xxl & IB-Vicuna-13b & LLaVA & Qwen \\ \midrule
Natural Parks and Reserves & 0.735  & 0.432 & 0.282 & 0.285 & 0.259        \\ 
Sports and Entertainment Venues & 0.644  & 0.467 & 0.311  & 0.220      & 0.389        \\ 
Historical and Cultural Sites  & 0.720  & 0.415 & 0.390 & 0.402      & 0.329        \\ 
Government and Public Buildings & 0.655  & 0.310 & 0.276 & 0.293      & 0.293        \\ 
Places of Worship & 0.213  & 0.149 & 0.170 & 0.383 & 0.106        \\ 
Infrastructure and Urban Features & 0.731  & 0.385 & 0.423 & 0.231 & 0.462        \\ 
Miscellaneous & 0.800  & 0.600 & 0.600 & -- & 0.800 \\ \midrule
Overall & 0.671  & 0.400 & 0.301 & 0.296 & 0.292 \\ \bottomrule
\end{tabular}
\end{table}

\begin{table}[h]
\caption{Landmark recognition refusal rate. (IB = InstructBLIP, LLaVA = LLaVA-v1.5, Qwen = Qwen-VL-Chat)}\label{tab:landmarks-refusal}
\centering

\begin{tabular}{l|c|c|c|c|c}
\toprule
 & GPT-4V & IB-T5-xxl & IB-Vicuna-13b & LLaVA & Qwen \\ \midrule
Refusal Rate & 0.054  & 0.000 & 0.033 & 0.000 & 0.314        \\ \bottomrule
\end{tabular}
\end{table}

\textit{What types of landmarks are VLMs good at recognizing?} Excluding the ``Miscellaneous'' category, GPT-4V performs the best at recognizing ``Natural Parks and Reserves'' and ``Infrastructure and Urban Features,'' while InstructBLIP-FLAN-T5-xxl performs the best at recognizing ``Sports and Entertainment Venues'' and ``Natural Parks and Reserves'' (\Cref{tab:landmarks-accuracy}). Overall, ``Places of Worship'' has the lowest recognition accuracy,
possibly due to their limited spatial footprint. The median area of polygons for ``Places of Worship'' is only 0.002 $\text{km}^2$ (\Cref{tab:landmarks-stat}). Example images of churches that GPT-4V fails to recognize confirm the image extents are too small to give contextual clues about where the churches are located (\Cref{fig:landmarks-example-churches}).

\textit{Is there any regional disparity in performance?} When recognition performance is grouped by US state, we observe that GPT-4V achieves perfect accuracy in Iowa, Louisiana, Arkansas, South Dakota, Rhode Island, and Delaware (\Cref{fig:landmarks-gpt4v-state}). Overall, it can achieve an average accuracy of over 70\% for most of the states on the West Coast and in the Northeast. 

\begin{figure}
    \centering
    \includegraphics[scale=0.62]{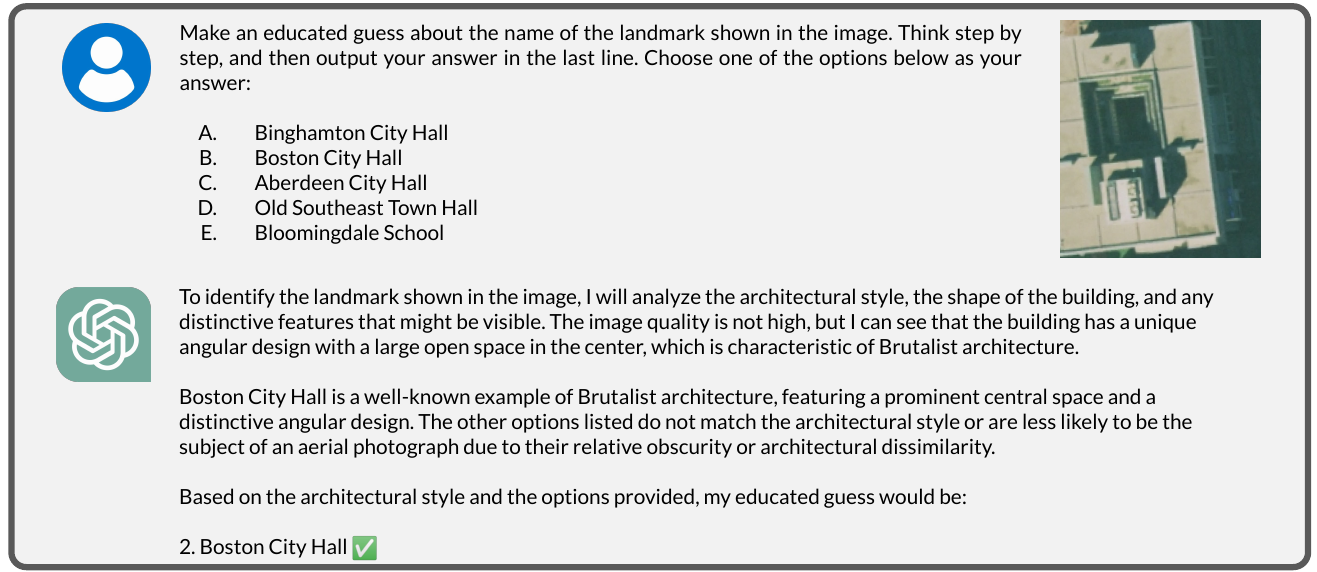}
    \caption{GPT-4V successfully corresponds visual cues with its knowledge about the architectural style of the landmark to arrive at the correct answer.}
    \label{fig:landmark-gpt4v-boston}
    \vspace{-10pt}
\end{figure}

\begin{figure}
    \centering
    \includegraphics[scale=0.62]{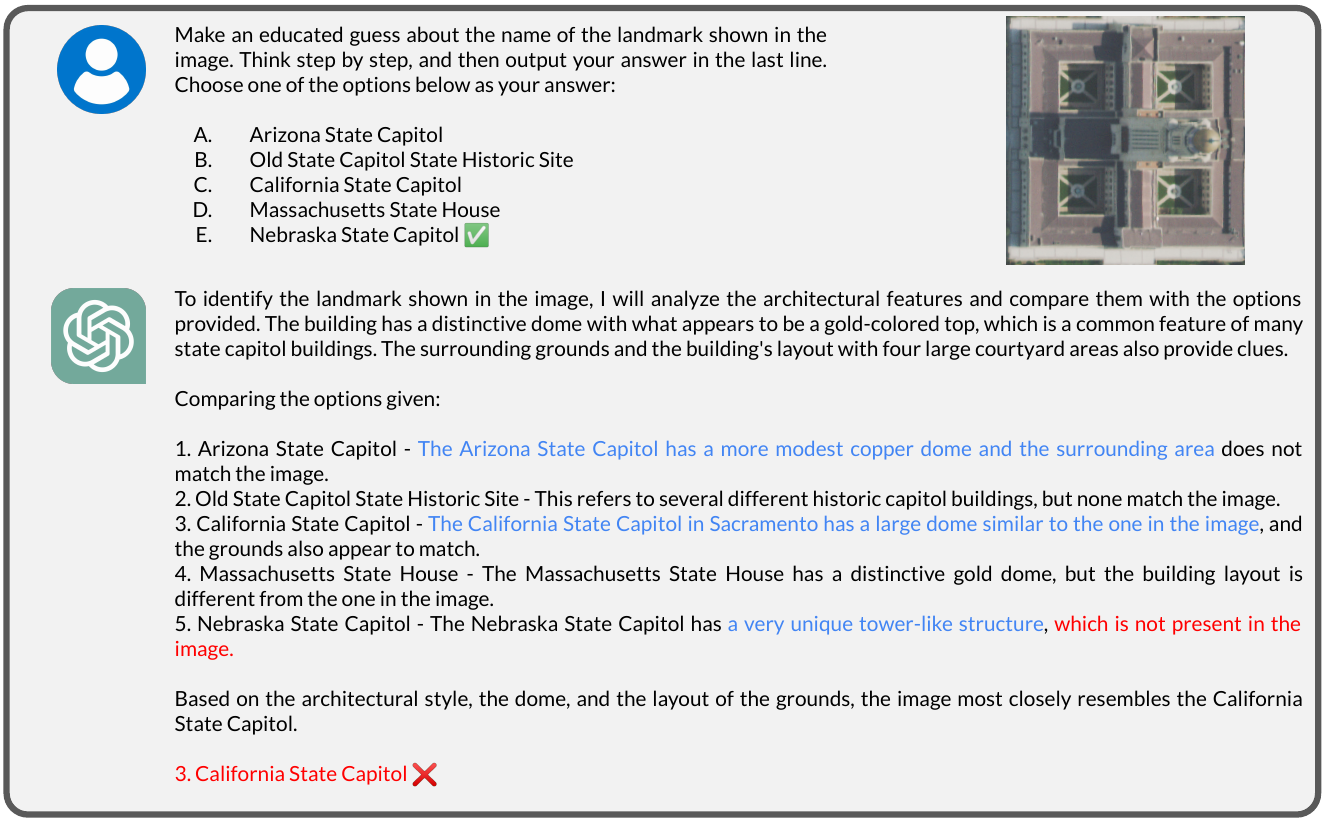}
    \caption{GPT-4V fails to identify the tower-like structure of the Nebraska State Capitol, leading to incorrect elimination.}
    \label{fig:landmark-gpt4v-nebraska}
\end{figure}

\begin{figure}
    \centering
    \includegraphics[scale=0.62]{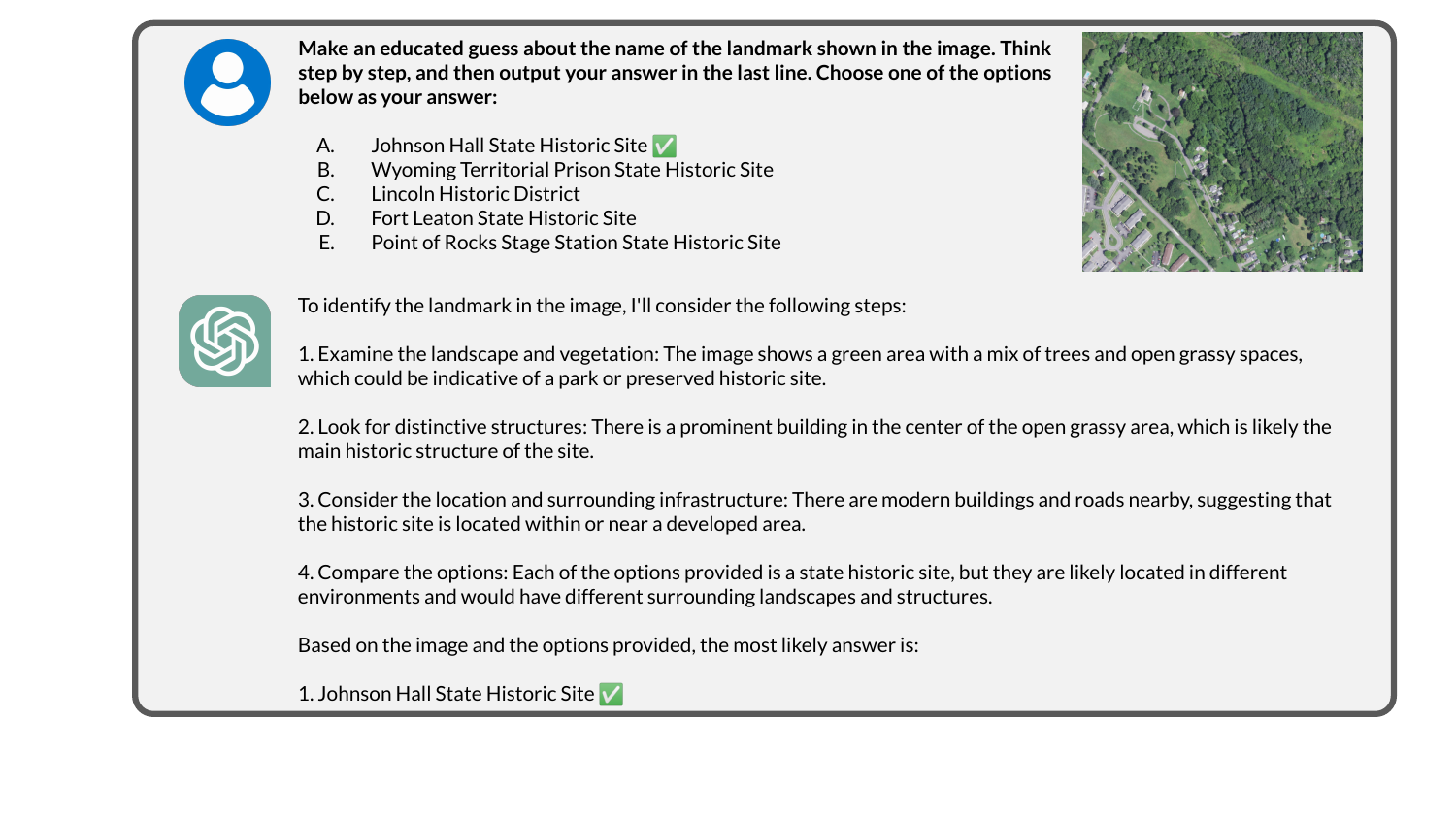}
    \caption{GPT-4V correctly identifies landmark but gives insufficient reasoning.}
    \label{fig:landmark-gpt4v-johnson}
\end{figure}

\textit{What common reasoning paths lead to correct or incorrect answers?} We manually examined the outputs of GPT-4V due to the comprehensive reasoning it can produce. Despite being instructed to think step by step, other models fail to output meaningful reasoning for the answer, if at all. In one example (\Cref{fig:landmark-gpt4-brown}), GPT-4V successfully uses its OCR capability to recognize the word ``BROWN'' written on the ground, leading to the correct answer of Brown Stadium. In \Cref{fig:landmark-gpt4v-boston}, GPT-4V uses both its visual knowledge and architectural knowledge to correctly infer the name of the city hall shown in the image. However, we find that GPT-4V can still be misled by its incorrect interpretation of the scene despite having the correct internal knowledge about the landmark. In the question illustrated in \Cref{fig:landmark-gpt4v-nebraska}, a  human can distinguish between these landmarks had they possessed the knowledge, but GPT-4V mistakenly concludes that the tower-like structure of the Nebraska State Capitol is not present, possibly due to the view angle of an off-nadir aerial image. We also find that GPT-4V can give insufficient or incorrect reasoning despite a correct final answer. As shown in \Cref{fig:landmark-gpt4v-johnson}, although GPT-4V correctly answers that the landmark is ``Johnson Hall State Historic Site,'' it gives a very generic thought process that does not involve any knowledge about specific options or any visual cue it picks up from the input image.

\begin{takeaway}[Takeaways]
    \begin{itemize}[leftmargin=1.3em,topsep=1pt,noitemsep]
        \item GPT-4V achieves the best zero-shot landmark recognition accuracy over other models by a large margin.
        \item All models achieve higher accuracy in categories with larger spatial extent (e.g., natural parks).
        \item GPT-4V tends to perform better at recognizing landmarks in coastal states over those in the mid-US. 
        \item GPT-4V can have an incorrect reasoning path even when the final answer is correct.
        \item Through OCR, GPT-4V can use text in an image to inform its decision.
        \item GPT-4V sometimes fails to pick up certain visual cues important for determining the final answer.
    \end{itemize}
\end{takeaway}

\subsection{Image Captioning}

Image captioning is another task that reflects the scene-understanding capabilities of VLMs. Given an aerial or satellite image, an ideal instruction-following VLM should be able to describe the input image at various levels of granularity and answer related questions, helping researchers and practitioners to interpret EO data at scale. 

\paragraph{Goals.} In this section, we evaluate the image captioning abilities of instruction-following VLMs on RSICD \cite{lu2017exploring}, a human-annotated dataset of remote sensing images and captions covering a variety of land use types. Through this task, we ask: \textit{1) How do VLM-generated captions compare with human-annotated examples both qualitatively and quantitatively? 2) To what granularity can VLM describe the image?}

\begin{figure}[h]
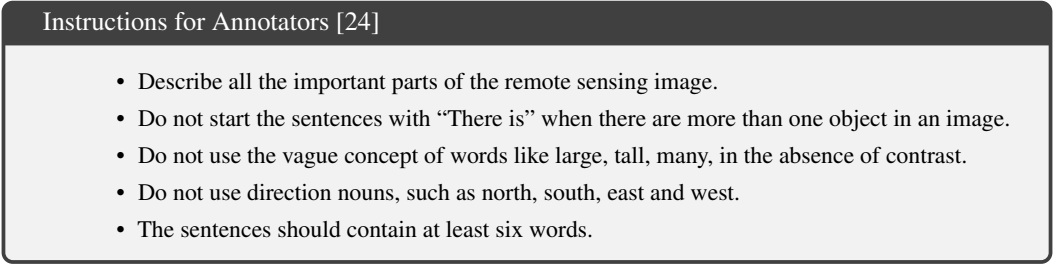

    \centering
    \begin{tcolorbox}[title=Instructions for Annotators \cite{lu2017exploring}, fontupper=\small]
        \begin{itemize}
            \item Describe all the important parts of the remote sensing image.
            \item Do not start the sentences with ``There is'' when there are more than one object in an image.
            \item Do not use the vague concept of words like large, tall, many, in the absence of contrast.
            \item Do not use direction nouns, such as north, south, east and west.
            \item The sentences should contain at least six words.
        \end{itemize}
    \end{tcolorbox}
    \caption{Annotation instructions for the RSICD dataset.}
    \label{box:captioning-rsicd-instrctions}
\end{figure}

\paragraph{Dataset Construction.} To construct the RSICD dataset \cite{lu2017exploring}, \citeauthor{lu2017exploring} first sourced high-resolution satellite base map images from a variety of providers, including Google Earth and Baidu Map to cover 31 land cover and land use categories. Then, three to five captions were annotated by student annotators. During annotation, the annotators were given a list of instructions (\Cref{box:captioning-rsicd-instrctions}) to avoid scale ambiguity, category ambiguity, and rotation ambiguity. In total, the dataset provided 8,730 training images and 1,009 validation images, which we use to query selected VLMs.

\begin{figure}[h]
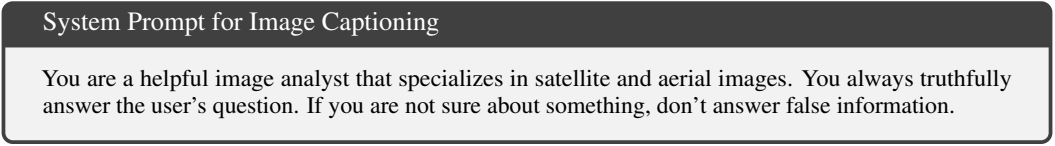

    \centering
    \begin{tcolorbox}[title=System Prompt for Image Captioning, fontupper=\small]
        You are a helpful image analyst that specializes in satellite and aerial images. You always truthfully answer the user's question. If you are not sure about something, don't answer false information.
    \end{tcolorbox}
    \caption{System prompt for image captioning.}
    \label{box:captioning-sys-prompt}
\end{figure}

\begin{figure}[h!]
    \centering
    \includegraphics[scale=0.62]{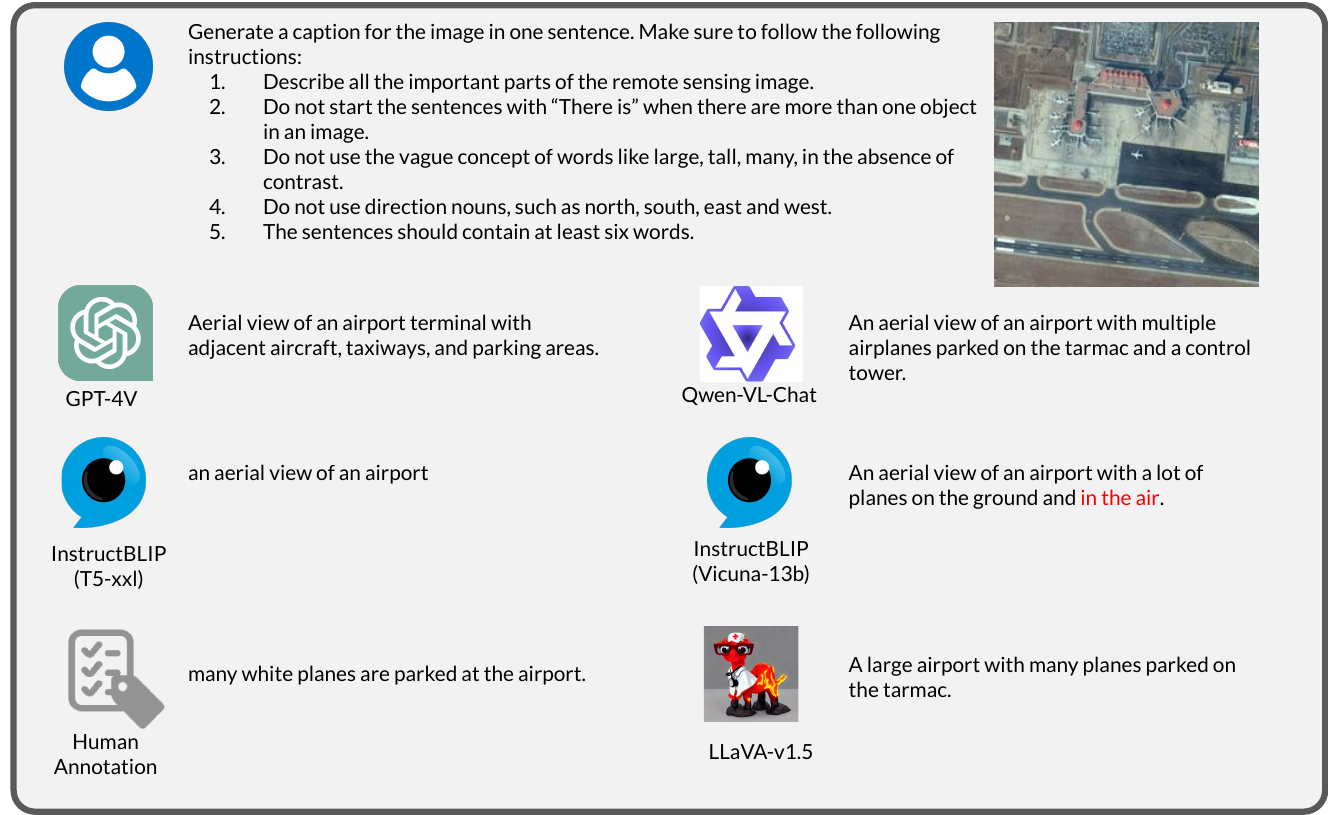}
    \caption{Example captions generated for an airport image.}
    \label{fig:caption-comparison-airport}
\end{figure}

\paragraph{System and Task Prompts.} We include the same instructions given to human annotators shown in \Cref{box:captioning-rsicd-instrctions} in the user prompt. We also provide an example of our user prompt and model outputs in \Cref{fig:caption-comparison-airport}. In addition, we use \Cref{box:captioning-sys-prompt} as our system prompt to set up the context of our conversation. As we do not include any in-context demonstration examples, all the evaluations are zero-shot.

\paragraph{Evaluation Setup.} To quantitatively evaluate the similarity between reference captions and VLM-generated captions, we employ a variety of metrics that compare their n-gram similarity and embedding similarity: (1) BLEU-n \cite{papineni-etal-2002-bleu} (where $n=1,2,3,4$) focus on the n-gram overlaps between the generated caption and the reference captions in RSICD. (2) METEOR \cite{banerjee-lavie-2005-meteor} extends BLEU-n by accounting for synonym matching and morphological variants in its assessment. (3) ROUGE \cite{lin-2004-rouge} evaluates the overlap of n-grams with a focus on recall. (4) CIDEr \cite{vedantam2014cider} considers the consensus of a set of reference captions, emphasizing the frequency of certain n-grams in the image captioning context. (5) SPICE goes further by analyzing the semantic scene graph similarity, offering a more semantic-oriented evaluation. (6) CLIPScore \cite{hessel2021clipscore} leverages the vision-language understanding ability of the CLIP model to evaluate the alignment between the generated caption and the image. (7) RefCLIPScore \cite{hessel2021clipscore} builds on CLIPScore by also considering reference captions, providing a reference-augmented assessment of model-generated captions. Overall, while all metrics provide valuable insights, RefCLIPScore is especially important as it considers not only the semantic similarity between the generated caption and the model caption but also the alignment between the generated caption and the corresponding image.

\paragraph{Results.} Based on n-gram metrics like BLEU-n, none of the models reach performance on par with specialist models. For example, LlaVA has the best BLEU-1 score (0.36) while the specialist model \cite{lu2017exploring} obtains a BLEU-1 score of 0.50 (\Cref{tab:rsicd-captioning}). All models have near-zero BLEU-4 scores, while the specialist model can obtain a BLEU-4 score of 0.18. In addition, all models have similar RefCLIPScore around 0.75--0.79. 

However, qualitative results are starkly different from what the quantitative metrics suggest. Despite GPT-4V achieving lower scores than other models, we caution against concluding that GPT-4V has an inferior image-captioning ability due to the low quality of human ``ground truth'' captions. We give examples below.


\begin{table}[h!]
\centering
\caption{Performance on remote sensing image captioning (IB = InstructBLIP). We recommend using RefCLIPScore as the main quantitative metric.}\label{tab:rsicd-captioning}
\resizebox{\linewidth}{!}{
\begin{tabular}{l|c|c|c|c|c|c|c|c|c|c}
\toprule
Model          & BLEU-1 & BLUE-2 & BLEU-3 & BLEU-4 & METEOR & ROUGE  & CIDEr  & SPICE  & CLIPScore & RefCLIPScore \\ \midrule
GPT-4V          & 0.257 & 0.114 & 0.0518 & 0.0226 & 0.135 & 0.213 & 0.135 & 0.113 & 0.777    & 0.754       \\ 
Qwen-VL-Chat           & 0.275 & 0.134 & 0.064 & 0.029 & \textbf{0.145} & 0.228 & 0.176 & 0.120 & 0.797    & 0.765       \\ 
IB-FLAN-t5-xxl & 0.292 & 0.149 & 0.074 & 0.030 & 0.093 & 0.214 & 0.221 & 0.093 & 0.783    & 0.776       \\ 
IB-Vicuna-13b  & 0.317 & 0.165 & 0.084 & 0.042 & 0.155 & 0.248 & 0.190 & 0.137 & \textbf{0.821}    & \textbf{0.787}       \\ 
LLaVA-v1.5          & \textbf{0.355} & \textbf{0.180} & \textbf{0.0991} & \textbf{0.0496} & 0.1406 & \textbf{0.257} & \textbf{0.317} & \textbf{0.140} & 0.739    & 0.773       \\ \midrule
LSTM \cite{lu2017exploring} & 0.500 & 0.320 & 0.232 & 0.178 & 0.205 & 0.433 & 1.180 & -- & -- & -- \\ 
\bottomrule
\end{tabular}
}
\end{table}

\begin{figure}[h!]
    \centering
    \includegraphics[scale=0.62]{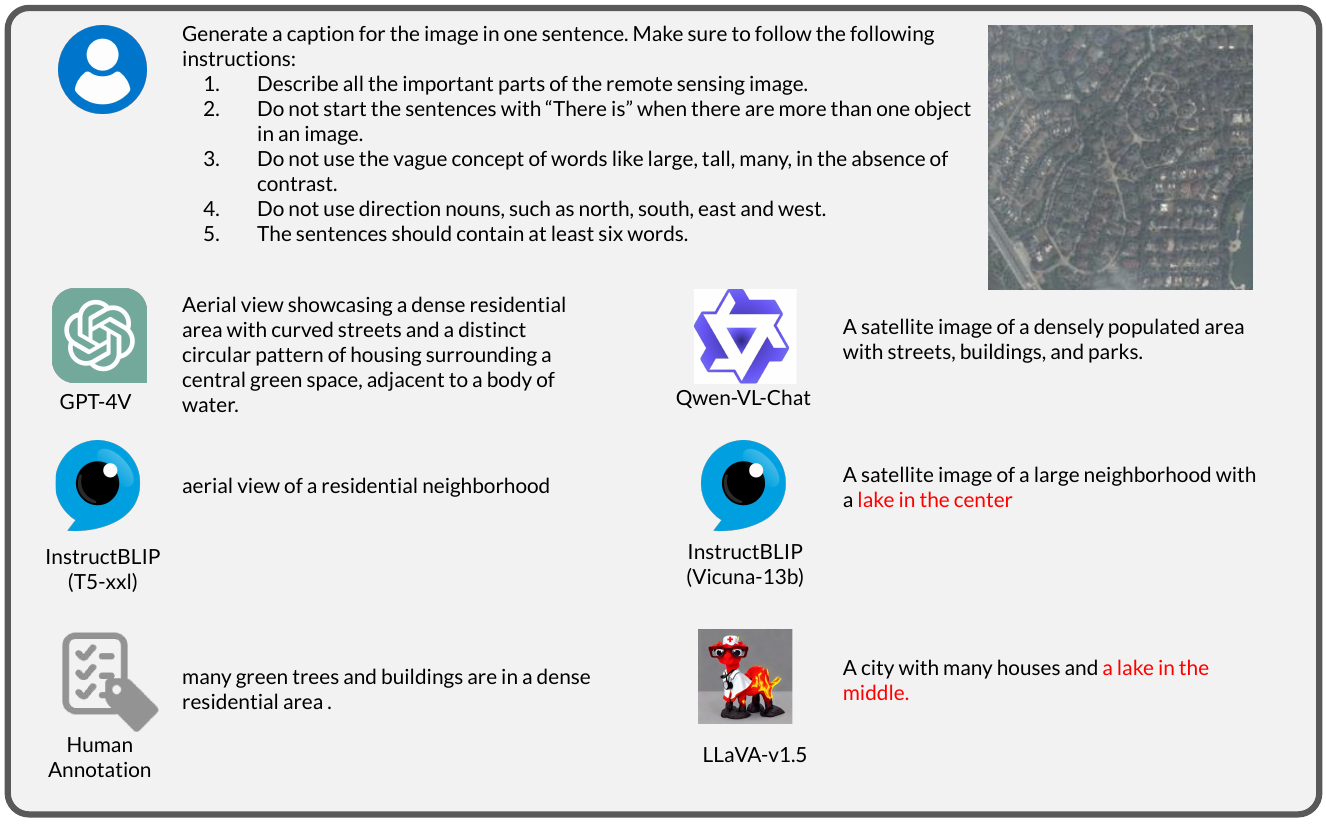}
    \caption{Example captions generated for a dense residential area. 
    }
    \label{fig:caption-comparison-residential}
\end{figure}

For an airport image, Qwen and GPT-4V produce more detailed captions than other models \textit{and human annotators}. Where the human caption reads ``many white planes are parked at the airport,'' Qwen and GPT-4V describe airplanes, tarmac, parking areas, and a control tower (\Cref{fig:caption-comparison-airport}). InstructBLIP-Vicuna-13b falsely claims that certain airplanes are in the air. For an urban image, Qwen and GPT-4V correctly identify the image as a densely populated area (\Cref{fig:caption-comparison-residential}). In addition, GPT-4V correctly mentions a nearby body of water. However, both InstructBLIP-Vicuna-13b and LLaVA-v1.5 falsely interpret the location of the water body as in the middle of the image. We hypothesize that such errors come from the inferior spatial reasoning abilities of open-source models compared with GPT-4V. The disparity between quantitative metrics and qualitative examples emphasizes the need for a high-quality image captioning dataset for EO data in order to better gauge the captioning abilities of new instruction-following VLMs.

\begin{takeaway}[Takeaways]
    \begin{itemize}[leftmargin=1.3em,topsep=1pt,noitemsep]
        \item Remote sensing is still in need of a high-quality captioning dataset.
        \item Captions generated by GPT-4V provide more detailed descriptions of the scene than other models---and existing human annotations.
        \item InstructBLIP and LLaVA-v1.5 often provide incorrect descriptions of the relative locations of ground objects.
    \end{itemize}
\end{takeaway}

\clearpage
\subsection{Land Use \& Land Cover Classification}
\label{sec:lulc}

Land use and land cover (LULC) classification is a canonical task in remote sensing. In this work, LULC classification complements landmark recognition and image captioning in evaluating the scene understanding of instruction-following VLMs. We construct multiple-choice questions for instruction-following VLMs to perform fine-grained image classification given natural language descriptions of candidate classes.

\paragraph{Goals.} In this section, we evaluate the LULC classification abilities of instruction-following VLMs on fMoW-WILDS \cite{christie2018functional, wilds2021,sagawa2022extending}, PatternNet \cite{zhou2017patternnet}, and BigEarthNet \cite{sumbul2019bigearthnet}, whose images span spatial resolutions of 0.2m to 10m. Through these tasks, we aim to understand \textit{1) Which model is the best for zero-shot land cover and land use classification? 2) What land cover types are instruction-following VLMs good at recognizing? 3) How does resolution affect the ability of VLMs to classify LULC?}

\paragraph{Dataset Construction.} Originally constructed as part of the WILDS benchmark \cite{wilds2021} for domain generalization, fMoW-WILDS carefully selects a subset of the Functional Map of the World (fMoW) dataset \cite{christie2018functional}, which consists of satellite images of around 0.5m/pixel resolution captured from 2002--2016 spanning the entire globe. It consists of a training set, in-distribution and out-of-distribution validation sets, and in-distribution and out-of-distribution test sets. We provide a detailed breakdown of the land use types covered by the dataset in \Cref{sec:app-scene-clf}. Due to the query limit on GPT-4V, we randomly subsample 2,000 images from the in-distribution and out-of-distribution test sets to form our evaluation dataset.

Secondly, we use the high-resolution images from Google satellite base maps in the PatternNet \cite{zhou2017patternnet} dataset. Originally used as a benchmark for image retrieval, PatternNet offers images from 38 diverse land use classes ranging from airports to residential areas with resolutions ranging from 0.233 m/pixel to 1.173 m/pixel. We reformulate it as a LULC classification benchmark by formatting the land use metadata as multiple-choice questions. The model is then instructed to select one option that best describes the image. To make the answers unambiguous, we reassign some land use types that originally appeared in the dataset to make the classes mutually exclusive. Due to the query limit on GPT-4V, we randomly subsample 1,000 images from the dataset.

Finally, we select the BigEarthNet \cite{sumbul2019bigearthnet} dataset to assess multi-class LULC classification performance on lower-resolution Sentinel-2 data (10m/pixel). BigEarthNet is a benchmark consisting of 590,326 Sentinel-2 image patches. (In a later version, the dataset was expanded to include Sentinel-1 images, but we only consider the Sentinel-2 subset in our benchmark.) We randomly subsample 1,000 images from the dataset and formulate the multi-class classification problem as a multiple-choice question with instructions for the model to select all applicable choices.

\begin{figure}
    \centering
    \includegraphics[scale=0.62]{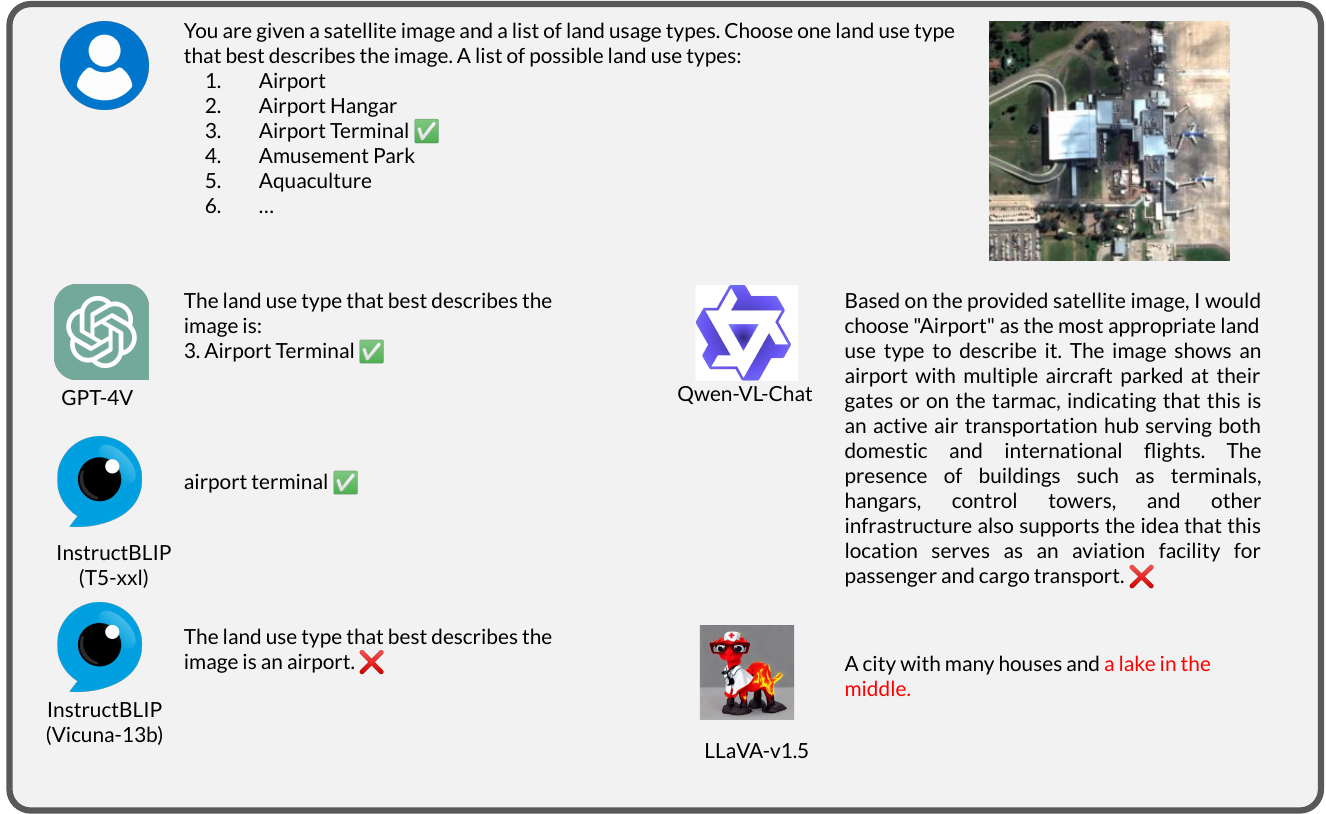}
    \caption{Example prompt and response for fMoW classification}
    \label{fig:fmow-comparison}
\end{figure}

\begin{figure}
    \centering
    \includegraphics[scale=0.62]{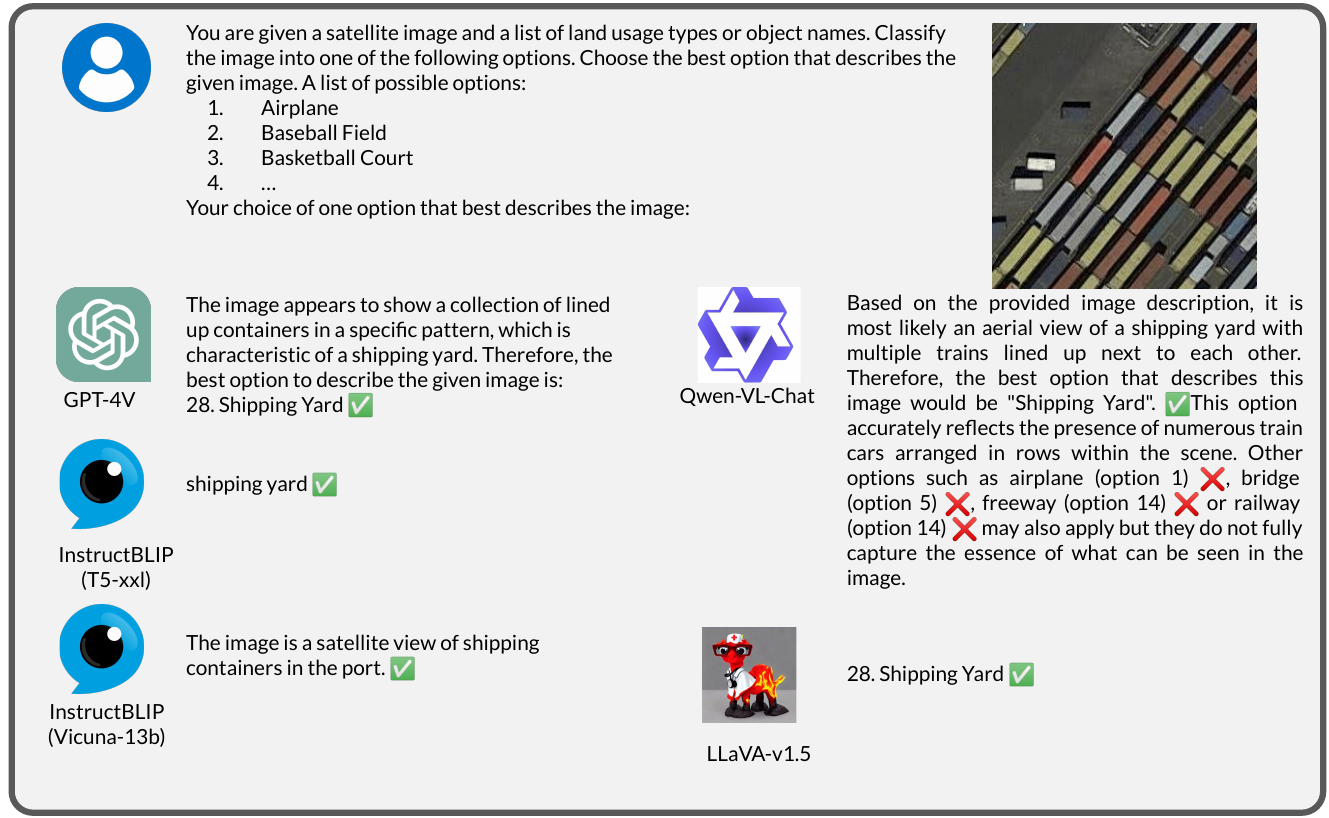}
    \caption{Example prompt and response for PatternNet classification}
    \label{fig:patternnet-comparison}
\end{figure}

\begin{figure}
    \centering
    \includegraphics[scale=0.62]{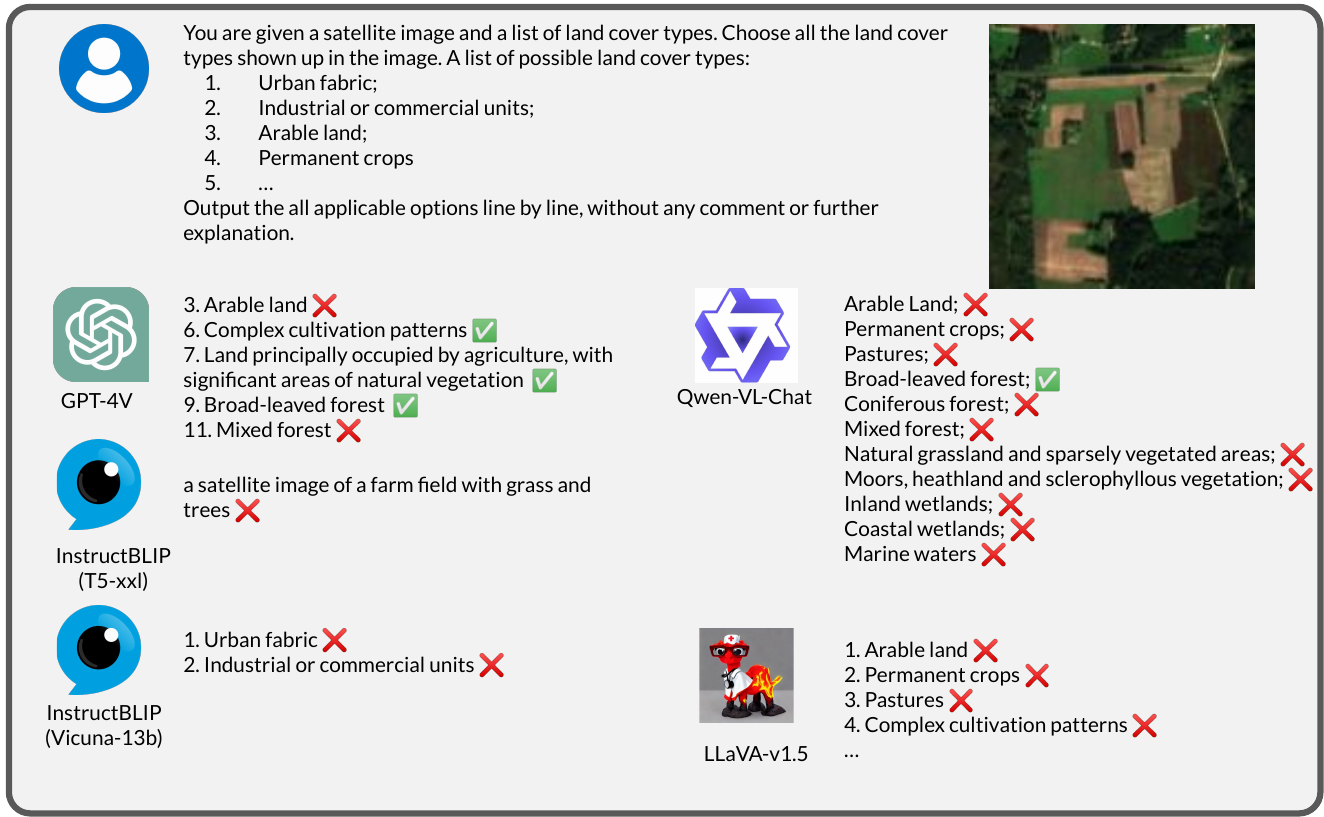}
    \caption{Example prompt and response for BigEarthNet classification}
    \label{fig:bigearthnet-comparison}
\end{figure}

\paragraph{System and Task Prompts.} Since all the tasks in this section have a similar context to image captioning, we use the same system prompt as image captioning (\Cref{box:captioning-sys-prompt}) for all of the classification tasks. We also show the multiple-choice user prompt and example outputs for fMoW (\Cref{fig:fmow-comparison}), BigEarthNet (\Cref{fig:bigearthnet-comparison}), and PatternNet (\Cref{fig:patternnet-comparison}), respectively.

\paragraph{Evaluation Setup.} To quantitatively evaluate the performance of classification tasks, we report 1) precision, 2) recall, 3) the F1 score for each class in the classification problem, 4) the overall (weighted) accuracy, and 5) the confusion matrix. In addition, to measure the instruction-following ability, we also report 6) the refusal rate of each model, defined as the rate at which the model refuses to answer the given question for any reason.

\paragraph{Results.} On land cover and land use classification tasks, we find that performance varies depending on image resolution, label ambiguity, and label granularity.

On fMoW-WILDS land use classification, GPT-4V fails to yield usable performance despite having the best accuracy (0.19) among all the models in our evaluation. It is significantly behind fine-tuned CLIP models, which can achieve an accuracy of 0.74 and 0.49 on the hold-out in-distribution and out-of-distribution test set, respectively (\Cref{tab:fmow-classification}). 
Examination of the class-by-class performance and confusion matrices reveals large differences among classes (\Cref{tab:fmow-gpt4v}--\Cref{tab:fmow-llava}), revealing that fMoW-WILDS remains a challenging benchmark for instruction-following VLMs. We notice that the inherent ambiguity of annotations partially contributes to the larger between-class gaps. The confusion matrix for GPT-4V reveals significant misclassification within classes that are semantically similar (\Cref{fig:fmow-gpt4v-confusion}). For example, we observe misclassification among ``Airport,'' ``Airport Hanger,'' and ``Airport Terminal.'' In addition, because many common object classes are co-located with residential areas, we observe misclassification of ``Parking Lot or Garage," ``Educational Institution,'' ``Place Of Worship,'' and ``Office Building'' to ``Multi-unit Residential.'' Since fMoW is an established benchmark widely used in the community, we do not reassign class labels to make class names mutually exclusive to prevent confusion in interpreting our results. This highlights the difficulty in comparing instruction-following VLMs, whose answers can be open-ended, to specialist models that provide a distribution strictly over the possible answers. 

On PatternNet land use classification, GPT-4V achieves an accuracy of 0.73 and an F1-score of 0.71 (\Cref{tab:patternnet-classification}). PatternNet contains very high-resolution images with disambiguated labels. There is also a much smaller gap between GPT-4V and the open-source models. In \Cref{tab:patternnet-gpt4v} -- \Cref{tab:patternnet-llava} of \Cref{sec:app-scene-clf}, we report the class-wise classification metrics and confusion matrices on PatternNet. For GPT-4V, the performance gap between different classes is small. However, we still notice that ``Christmas Tree Farm,'' ``Mobile Home Park,'' ``Nursing Home,'' and ``Coastal Mansion'' classes are commonly misclassified into ``Residential.''

Finally, VLM performance on BigEarthNet, which has low-resolution images with high label granularity, lies between fMoW and PatternNet performance. Qwen, LLaVA, and GPT-4V achieve similar F1-scores around 0.4 (\Cref{tab:bigearthnet-classification}). We also analyze the class-wise classification metrics and confusion matrices for the BigEarthNet evaluation (\Cref{tab:BigEarthNet-gpt4v} -- \Cref{tab:bigearthnet-llava} of \Cref{sec:app-scene-clf}). Llava achieves a significantly higher recall (\Cref{tab:bigearthnet-llava} of \Cref{sec:app-scene-clf}) than other models, which, upon manual examination, is due to the model repeating all available options for every question. On the other hand, GPT-4V has a moderate F1-score (\Cref{tab:BigEarthNet-gpt4v} of \Cref{sec:app-scene-clf}) for classes with more generic descriptions, such as ``Arable land,'' ``Urban fabric,'' and ``Inland waters,'' but completely fails to identify classes like ``Moors, heathland and sclerophyllous vegetation'' (\Cref{tab:bigearthnet-classification}).

Overall, we find that GPT-4V performance varies depending on image resolution, label ambiguity, and label granularity. It achieves high performance on PatternNet with high image resolution and disambiguated labels but lower performance on fMoW-WILDS due to label ambiguity and BigEarthNet due to low-resolution images and fine-grained labels. This points to GPT-4V's good general scene understanding; however, VLMs are more likely to be successful at LULC classification when images are high-resolution and class labels are disambiguated and not very technical.

\begin{table}[h]
\caption{fMoW-WILDS land use classification metrics 
}\label{tab:fmow-classification}
\resizebox{\linewidth}{!}{
\centering
\begin{tabular}{l|c|c|c|c|c}
\toprule
Model & Average Precision & Average Recall & Average F1 & Accuracy & Refusal Rate \\ \midrule
GPT-4V                   & \textbf{0.28} & \textbf{0.19} & \textbf{0.16} & \textbf{0.19} & 0.025 \\
Qwen-VL-Chat             & 0.17 & 0.04 & 0.04 & 0.04 & \textbf{0.069} \\
InstructBLIP-FLAN-T5-xxl & 0.26 & 0.13 & 0.12 & 0.13 & 0.000 \\
InstructBLIP-Vicuna-13b  & 0.21 & 0.15 & 0.13 & 0.15 & 0.031 \\
LLaVA-v1.5               & 0.26 & 0.18 & 0.15 & 0.18 & 0.000 \\ \midrule
Wise-FT (ID) \cite{wortsman2021robust} & -- & -- & -- & 0.74 & -- \\ 
Wise-FT (OoD) \cite{wortsman2021robust} & -- & -- & -- & 0.49 & -- \\ \midrule
Random Guess             & --  & --  & --  & 0.03 & -- \\
\bottomrule
\end{tabular}
}
\end{table}

\begin{table}[h]
\caption{PatternNet land use classification metrics}\label{tab:patternnet-classification}
\resizebox{\linewidth}{!}{
\centering
\begin{tabular}{l|c|c|c|c|c}
\toprule
Model & Average Precision & Average Recall & Average F1 & Accuracy & Refusal Rate \\ \midrule
GPT-4V                   & \textbf{0.78} & \textbf{0.73} & \textbf{0.71} & \textbf{0.73} & 0.006 \\
Qwen-VL-Chat             & 0.57	& 0.39 & \textbf{0.40} & 0.39 & \textbf{0.044} \\
InstructBLIP-FLAN-T5-xxl & 0.80 & 0.67 & 0.66 & 0.67 & 0.000 \\
InstructBLIP-Vicuna-13b  & 0.72 & 0.58 & 0.60 & 0.58 & 0.003 \\
LLaVA-v1.5               & 0.65 & 0.63 & 0.58 & 0.63 & 0.000 \\ \midrule
Random Guess             & --  & --  & -- & 0.028 & -- \\
\bottomrule
\end{tabular}
}
\end{table}

\begin{table}[h!]
\caption{BigEarthNet multi-label land cover classification metrics}\label{tab:bigearthnet-classification}
\resizebox{\linewidth}{!}{
\begin{tabular}{l|c|c|c|c}
\toprule
Model & Average Precision & Average Recall & Average F1 & Refusal Rate \\ \midrule
GPT-4V                   & 0.49 & 0.43 & 0.38 & \textbf{0.076} \\
Qwen-VL-Chat             & \textbf{0.57}	& 0.39 & \textbf{0.40} & 0.044 \\
InstructBLIP-FLAN-T5-xxl & 0.41 & 0.01 & 0.02 & 0.000 \\
InstructBLIP-Vicuna-13b  & 0.01 & 0.06 & 0.01 & 0.000 \\
LLaVA-v1.5               & 0.27 & \textbf{0.83} & 0.39 & 0.000 \\ \bottomrule
\end{tabular}
}
\end{table}

\begin{takeaway}[Takeaways]
    \begin{itemize}[leftmargin=1.3em,topsep=1pt,noitemsep]
        \item VLMs perform significantly worse than specialized models at land cover classification.
        \item Among VLMs, GPT-4V achieves the best performance on fMoW-WILDS and PatternNet.
        \item The ambiguity of class labels partially contributes to poor performance on fMoW-WILDS, pointing to the challenge of comparing VLMs to specialized LULC classifiers.
        \item The low resolution and the lack of multi-spectral information in our BigEarthNet evaluation partially contribute to the poor performance of GPT-4V.
    \end{itemize}
\end{takeaway}

%% file: body/localization.tex
\subsection{Object Localization}\label{sec:localization}

Object detection and localization are crucial capabilities for 
downstream applications 
of remote sensing
like building footprint mapping \cite{sirko2021continentalscale}, animal conservation \cite{laradji2020counting}, and illegal fishing monitoring \cite{NEURIPS2022_f4d4a021}. At present, specialist models are trained by machine learning experts to perform each downstream application separately. An ideal instruction-following VLM for EO data should perform accurate object localization 
and be able to reason about the relationships between objects
to answer a natural language prompt from a non-technical user, even when EO images are complex and cluttered.

\paragraph{Goals.} In this section, we evaluate instruction-following VLMs on their abilities to localize an object in a satellite image, given a natural language description of its properties and relative position. Also known as Referring Expression Comprehension (REC), this task requires the model to detect only one single object that the text refers to in an image with possibly multiple confounding objects. Through this evaluation, we aim to ask \textit{1) How accurately can general-purpose VLMs localize objects in satellite images? 2) Can VLMs follow user instructions and output the results in the desired format?} 

\paragraph{Dataset Construction.} To assess the object localization ability of instruction-following VLMs, we consider DIOR-RSVG \cite{diorrsvg10056343}, a dataset of \{(image, referring expression(s), bounding box(es))\} triplets for improving and assessing the ability to perform REC tasks on EO data. The dataset contains 23,463 satellite images of dimension $800 \times 800$ pixels, covering 20 object categories, with the average length of the referring expression being 7.47 text tokens. The creation of this data involves box sampling from the DIOR dataset \cite{LI2020296}, object attribute (geometry, color, etc.) extraction, expression generation based on empirical rules, and human verification, producing a rich collection of EO data with diverse referring expressions.

\begin{figure}[h]
    \centering
    \includegraphics[scale=0.62]{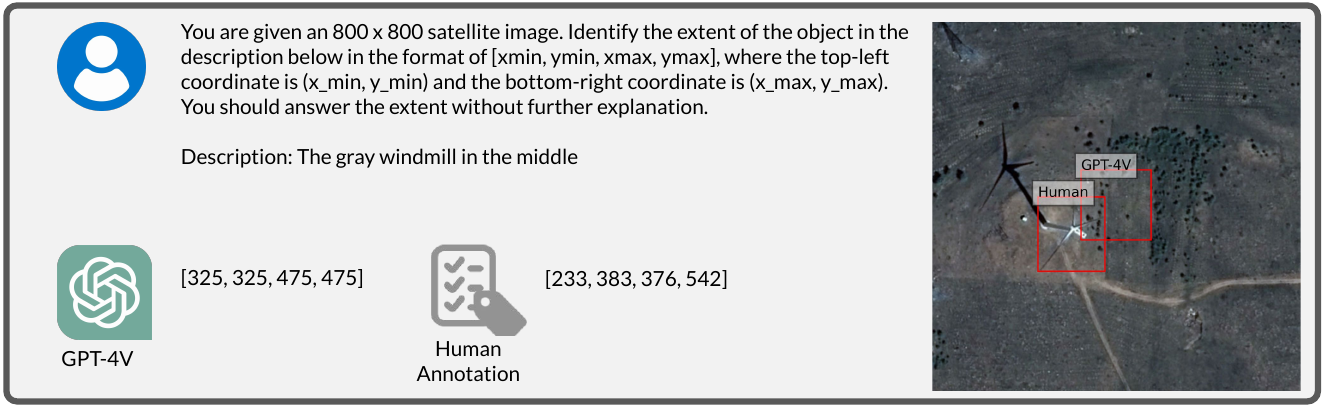}
    \caption{Example prompt and response for DIOR-RSVG object localization}
    \label{fig:localization-prompt}
\end{figure}

\paragraph{System and Task Prompts.} The system prompt we use to perform the REC task on EO images includes a generic description of the capability required to answer user questions and a general requirement of the model answer (\Cref{box:localization-sys-prompt}). Then, the user prompt instructs the model to perform the REC task by describing the dimension of the image and specifying the output formats (\Cref{fig:localization-prompt}).

\begin{figure}[h]
    \centering
    \begin{tcolorbox}[title=System Prompt for Object Localization, fontupper=\small]
        You are a helpful image analyst that specializes in localizing objects from satellite and aerial images given a natural language instruction. You always truthfully answer the user's question. If you are not sure about something, don't answer false information.
    \end{tcolorbox}
    \caption{System prompt for object localization.}
    \label{box:localization-sys-prompt}
\end{figure}

\paragraph{Evaluation Setup.} To evaluate the generated bounding boxes, we compute the mean intersection over union (IoU) across images, defined below, where $U_i$ is the area of the union between the predicted bounding box, and estimated bounding box for the $i$th expression and $I_i$ is the area of their intersection.
\begin{equation}\label{eq:mean-iou}
    \text{mean IoU} = \frac{1}{N} \sum_{i = 1}^N \frac{I_i}{U_i}
\end{equation}


Furthermore, following the evaluation setups in \cite{diorrsvg10056343}, we report an accuracy metric with different IoU thresholds, in which a prediction is correct if the IoU is above a certain threshold. Following \cite{diorrsvg10056343}, we report the metrics with IoU thresholds at 0.5, 0.6, 0.7, 0.8, and 0.9, termed Pr@0.5, Pr@0.6, Pr@0.7, Pr@0.8, and Pr@0.9, respectively.

Finally, we calculate the mean distance between the centroid of the ground truth bounding box and the centroid of the predicted bounding box in pixels.

\paragraph{Results.} 
All five models perform poorly on object localization in remote sensing images (\Cref{tab:rsvg-localization}).
Overall, GPT-4V does the best, achieving a Pr@0.5 of 0.076, a mean IoU of 0.158, and a mean centroid distance of 147 pixels, with a near-zero refusal rate of 0.02 (\Cref{tab:rsvg-localization}). Although specifically fine-tuned on REC tasks for natural images, Qwen has lower localization accuracy compared with GPT-4V, with a Pr@0.5 of 0.040, a mean IoU of 0.007, and a much higher mean centroid distance of 336 pixels, alongside a high refusal rate of 0.69. Surprisingly, both InstructBLIP-FLAN-T5-xxl and InstructBLIP-Vicuna-13b models fail to follow the specified answer format with a refusal rate of 1.00. LLaVA recorded the lowest scores among the models, with Pr@0.5, a mean IoU of 0.000, and the highest mean centroid distance of 580 pixels, while answering all questions. This especially low performance is because LLaVA fails to comprehend the dimensions of the image, as all of its answers have coordinate values of less than 1. The results from the current instruction-following VLMs significantly trail behind MGVLF, the best model in \cite{diorrsvg10056343} specifically trained to perform REC tasks on satellite images.

\begin{table}[h]
\caption{DIOR-RSVG object localization performance.}\label{tab:rsvg-localization}
\resizebox{\linewidth}{!}{
\begin{tabular}{l|c|c|c|c}
\toprule
Model & Pr@0.5 & mean IoU & Mean Centroid Distance (pixels) & Refusal Rate \\ \midrule
GPT-4V                   & 0.076 & 0.158 & 147 & 0.02 \\
Qwen-VL-Chat             & 0.053 & 0.009 & 262 & 0.69 \\
InstructBLIP-FLAN-T5-xxl & -- & -- & -- & 1.00 \\
InstructBLIP-Vicuna-13b  & -- & -- & -- & 1.00 \\
LLaVA-v1.5               & 0.0 & 0.0 & 579 & 0.00 \\ \midrule
MGVLF \cite{diorrsvg10056343} & 0.768 & 0.680 & -- & -- \\ \bottomrule
\end{tabular}
}
\end{table}

\begin{takeaway}[Takeaways]
    \begin{itemize}[leftmargin=1.3em,topsep=1pt,noitemsep]
        \item VLMs perform significantly worse than specialized models on object localization.
        \item GPT-4V generates object bounding boxes that have, on average, IoUs of 0.16, suggesting general but not precise awareness of where objects are.
    \end{itemize}
\end{takeaway}

\subsection{Counting}\label{sec:counting}

We also consider counting the number of objects in an aerial or satellite image as a crucial capability for VLMs. For example, counting trees and animal populations is crucial for conservation and should be an automatable task. 
In urban settings, correctly identifying the number of vehicles or buildings in an aerial image can also help in traffic management, city planning,  infrastructure monitoring, and disaster impact assessment. Unlike in natural images, counting in remote sensing imagery generally requires identifying the correct number of very small yet cluttered objects from overhead images.

\paragraph{Goals.} We evaluate instruction-following VLMs on their ability to count objects under realistic settings such as forest conservation \cite{10.1371/journal.pcbi.1009180}, urban vehicle monitoring \cite{mundhenk2016large}, animal conservation \cite{eikelboom2019improving}, and building footprint assessment \cite{gupta2019xbd}. We ask: \textit{1) How accurately can VLMs count small, cluttered objects? 
2) Can VLMs follow user instructions and output the results in the desired format?}

\paragraph{Dataset Construction.} To test the tree-counting abilities of VLMs, we use the annotated validation images from the Neon Tree Evaluation benchmark \cite{10.1371/journal.pcbi.1009180}. This benchmark synthesizes multi-sensor data (RGB, LiDAR, hyperspectral) from the National Ecological Observation Network (NEON) to characterize tree canopies in diverse U.S. forest types. This dataset includes over 6,000 image-annotated crowns, 400 field-annotated crowns, and 3,000 canopy stem points. In our evaluation, we take all of the 194 annotated RGB images in the validation set with a 0.1 m/pixel resolution.

For car counting, we choose the Cars Overhead with Context (COWC) dataset \cite{mundhenk2016large}, which is a collection of overhead images with a 0.15 m/pixel resolution containing different types of vehicles like pickups and sedans. To form our evaluation dataset, we randomly choose 1,000 images from four locations, including Potsdam, Selwyn, Toronto, and Utah.

For animal counting, we use the high-resolution animal detection dataset by \citeauthor{eikelboom2019improving}, which includes 561 aerial images collected by the Kenya Wildlife Service in Tsavo National Park and the Laikipia-Samburu Ecosystem. Images were captured from a helicopter 
when large animal groups were spotted. The annotation in the dataset includes various species, primarily elephants, giraffes, and zebras, with each animal identified and annotated with a bounding box. We use all of the 112 test images in the dataset for our evaluation.

Finally, for building counting, we use Maxar/DigitalGlobe satellite images with a resolution of less than 0.8 m/pixel from the xBD \cite{gupta2019xbd} dataset, which features building annotations by domain experts. We use all of the 933 test images in the dataset for our evaluation. Since we also evaluate change detection tasks on this dataset, we defer further details about this dataset to \Cref{sec:change}. 

\begin{figure}[h]
    \centering
    \includegraphics[scale=0.62]{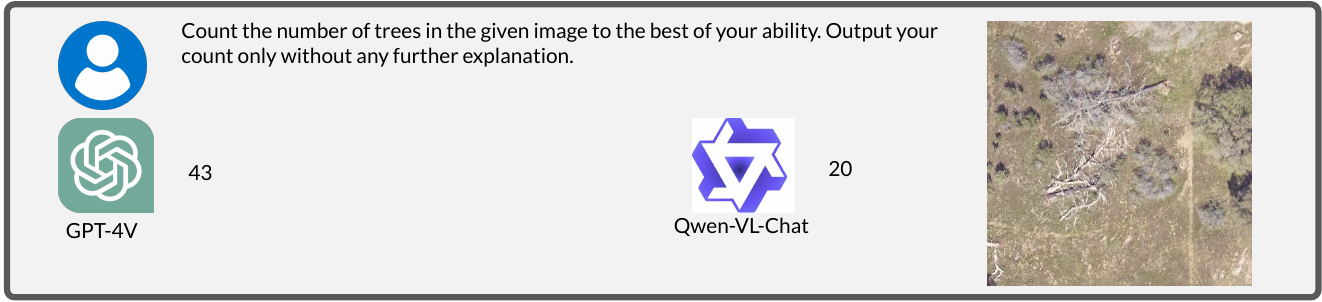}
    \caption{Example user prompt and response for NEON tree counting}
    \label{fig:neon-comparison}
\end{figure}

\paragraph{System and Task Prompts.} To form the system prompt for counting on the NEON Tree dataset \cite{10.1371/journal.pcbi.1009180}, we insert additional instruction for the model not to refuse the question from the user to reduce the refusal rate, as we observe that a generic prompt without such instruction results in a high refusal rate such that the answer is not meaningful (\Cref{box:counting-tree-sys-prompt}). By a similar principle, we form the system prompts for the aerial animal counting task (\Cref{box:counting-animal-sys-prompt} of \Cref{sec:app-counting}). We use a simple task description for the COWC vehicle counting task (\Cref{box:counting-vehicle-sys-prompt} of \Cref{sec:app-counting}). In \Cref{fig:neon-comparison}, we showcase the user prompt and an example model response. The user prompt and example responses for the COWC and aerial animal datasets can be found in \Cref{fig:aerial-animal-comparison} and \Cref{fig:cowc-comparison}. 

\begin{figure}[h]
    \centering
    \begin{tcolorbox}[title=System Prompt for Counting Trees, fontupper=\small]
        You are a helpful image analyst who specializes in counting trees from aerial images. Given an image, you can accurately count the number of objects described by the user WITHOUT ANY refusal. Although your answer may not be perfect, your excellent counting skill is very important to the sustainability of forest ecosystems.
    \end{tcolorbox}
    \caption{System prompt for counting trees.}
    \label{box:counting-tree-sys-prompt}
\end{figure}

\paragraph{Evaluation Setup.} We report the mean absolute error (MAE), mean absolute percentage error (MAPE), and the coefficient of determination $R^2$ between the ground truth and the count given by the model. We define MAE as
\begin{equation}\label{eq:mae}
    \text{MAE}(y, \hat{y}) = \frac{1}{N} \sum_{i = 1}^N |y_i - \hat{y_i}|,
\end{equation}
where $N$ is the number of examples, $y_i$ is the actual count, and $\hat{y_i}$ is the estimated count. In addition, we define MAPE as
\begin{equation}\label{eq:mape}
    \text{MAPE} = \frac{1}{n} \sum_{i=1}^{n} \left| \frac{A_i - F_i}{A_i} \right|,
\end{equation}
where $n$ is the number of testing samples, $A_i$ is the actual object count in the $i$th example, and $F_i$ is the model estimate for the object count in the $i$th example.

Overall, MAE is more relevant in scenarios where we want to understand the absolute error in the object counts, while MAPE gauges the relative error. On the other hand, $R^2$ is more important when we care about capturing differences across images. A good $R^2$ enables us to calibrate model predictions, even if MAE and MAPE are bad.

Furthermore, we calculate the refusal rate, the rate at which the model refuses to give an answer or outputs an answer with an incorrect format, indicating non-instruction-following behaviors. For tree counting, vehicle counting, and animal counting tasks, we calculate MAE, MAPE, and $R^2$ without refused examples while providing another version in which refused examples are considered counting no object in \Cref{tab:app-neon-counting} - \Cref{tab:app-animal-counting} of \Cref{sec:app-counting}. For the building counting task, we omit MAPE due to the existence of examples with no building.

\begin{figure}
    \centering
    \includegraphics[scale=0.22]{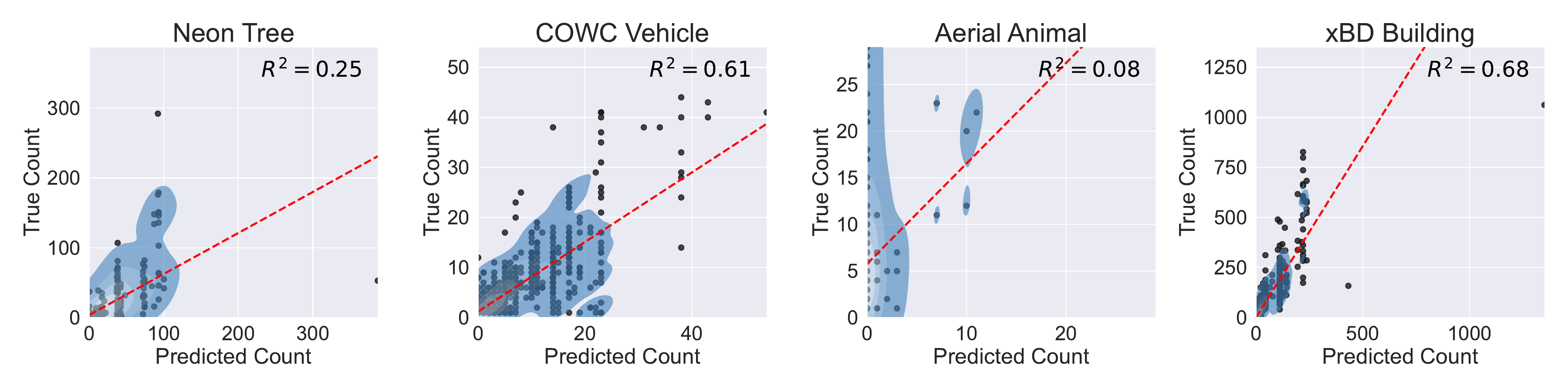}
    \caption{Scatterplot of GPT-4V counting results for trees, vehicles, animals, and buildings. 
    }
    \label{fig:counting-gpt4v}
\end{figure}

\paragraph{Results.} 
Overall, GPT-4V performs much better on vehicle and building counting than tree and animal counting (\Cref{fig:counting-gpt4v}), while other models achieve the best performance on vehicle counting (\Cref{fig:counting-qwen}). However, even the best VLM at present is much worse at counting in remote sensing imagery than specialized models. 

No model performs well on the NEON Tree counting task (\Cref{tab:neon-counting}). MAPE varies significantly among models; InstructBLIP-FLAN-T5-xxl obtains the lowest MAPE of 0.870, while Qwen displays by far the worst MAPE of $1.28\times10^6$. 
The $R^2$ values are generally low across models as well. LLaVA scores the highest $R^2$ value of 0.353 despite its higher MAPE. In terms of refusal, InstructBLIP-FLAN-T5-xxl has the highest refusal rate of 0.54
despite its high counting accuracy. In contrast, Qwen and LLaVA have zero refusal rates, attempting every task regardless of accuracy. GPT-4V and InstructBLIP-Vicuna-13b have moderate to low refusal rates.

Results on COWC vehicle counting are qualitatively different from NEON tree counting (\Cref{tab:cowc-counting}). All five models generate some reasonable---although far from perfect---vehicle counts. LLaVA exhibits the highest accuracy with the lowest MAPE of 0.467 and MAE of 2.695, followed closely by InstructBLIP-FLAN-T5-xxl. However, all models are significantly inferior to the specialist model in \cite{mundhenk2016large}, which has an MAE of only 0.248.
The $R^2$ values indicate a moderate correlation between the estimated and true counts for all models, with GPT-4V leading at 0.528. 
Qwen performs the worst, although still better compared to tree counting. In terms of refusal rate, only GPT-4V and InstructBLIP-FLAN-T5-xxl demonstrate moderate to low degrees of refusal, while other models fully answer the question following instructions.

For animal counting, only GPT-4V and Qwen provide parsable outputs, while InstructBLIP-FLAN-T5-xxl, InstructBLIP-Vicuna-13b, and LLaVA generate incorrect output formats or predict zeros for all examples (\Cref{tab:animal-counting}).
GPT-4V and Qwen have similar MAPE scores, but both predict poorly with $R^2 < 0.1$. We note that this task is very challenging, as all images are off-nadir views of distant animals. Current VLMs appear to be very far from assisting with conservation-related counting.

For the building counting task, only GPT-4V and Qwen provide meaningful results, while other models fail to generate correct JSON outputs following our prompts, as shown by the ``Before Disaster'' category in \Cref{tab:change-detection}. Compared with Qwen, GPT-4V achieves a significantly higher $R^2$ (0.68 v.s. 0.0) and lower MAE (32 v.s. 2942) without a significant refusal rate.

\begin{table}[h]
\caption{NEON tree counting performance}\label{tab:neon-counting}
\centering
\begin{tabular}{l|c|c|c|c}
\toprule
Model & MAE $\downarrow$ & MAPE $\downarrow$ & $R^2$ $\uparrow$ & Refusal Rate $\downarrow$ \\ \midrule
GPT-4V & 23.033 & 1.890 & 0.249 & 0.21 \\
Qwen-VL-Chat & $8.40\times10^6$ & $1.28\times10^6$ & 0.000 & 0.00 \\
InstructBLIP-FLAN-T5-xxl & 16.551 & 0.717 & 0.093 & 0.54 \\
InstructBLIP-Vicuna-13b  & 27.172 & 1.236 & 0.001 & 0.01 \\
LLaVA-v1.5 & 148.479 & 4.481 & 0.353 & 0.00 \\ \bottomrule
\end{tabular}
\end{table}

\begin{table}[h]
\caption{COWC vehicle counting performance}\label{tab:cowc-counting}
\centering
\begin{tabular}{l|c|c|c|c}
\toprule
Model & MAE $\downarrow$ & MAPE $\downarrow$ & $R^2$ $\uparrow$ & Refusal Rate $\downarrow$ \\ \midrule
GPT-4V & 2.853 & 0.818 & 0.612 & 0.15 \\
Qwen-VL-Chat & 4.352 & 1.711 & 0.132 & 0.00 \\
InstructBLIP-FLAN-T5-xxl & 2.919 & 0.543 & 0.425 & 0.05 \\
InstructBLIP-Vicuna-13b  & 3.558 & 0.878 & 0.279 & 0.00 \\
LLaVA-v1.5 & 2.695 & 0.467 & 0.437 & 0.00 \\ \midrule
ResCeption \cite{mundhenk2016large} & 0.248 & -- & -- & -- \\ \bottomrule
\end{tabular}
\end{table}

\begin{table}[h]
\caption{Aerial animal counting performance. InstructBLIP models have high refusal rates such that we cannot calculate meaningful metrics, while LLaVA-v1.5 answers zero to all questions.}\label{tab:animal-counting}
\centering
\begin{tabular}{l|c|c|c|c}
\toprule
Model & MAE $\downarrow$ & MAPE $\downarrow$ & $R^2$ $\uparrow$ & Refusal Rate $\downarrow$ \\ \midrule
GPT-4V & 6.991 & 0.938 & 0.076 & 0.02 \\
Qwen-VL-Chat & 6.330 & 1.081 & 0.015 & 0.00 \\
InstructBLIP-FLAN-T5-xxl & -- & -- & -- & 1.00 \\
InstructBLIP-Vicuna-13b  & -- & -- & -- & 1.00 \\
LLaVA-v1.5 & -- & -- & -- & 0.00 \\ \bottomrule
\end{tabular}
\end{table}

\begin{takeaway}[Takeaways]
    \begin{itemize}[leftmargin=1.3em,topsep=1pt,noitemsep]
        \item VLMs perform significantly worse than specialized models on object counting.
        \item At present, vehicle and building counting appear easier for VLMs than tree and animal counting.
        \item GPT-4V and Qwen consistently follow instructions and have low or zero refusal rates. InstructBLIP models are less instruction-following. Only GPT-4V and Qwen generate outputs for animal counting, albeit with poor accuracy.
    \end{itemize}
\end{takeaway}

%% file: body/change.tex
Many of the most important remote sensing applications---deforestation, urban development, disaster relief---involve detecting changes over time.
Given multiple remote sensing images of the same geographical extent and natural language instructions, an ideal VLM for EO data should understand and localize the temporal difference across images and answer questions about these changes.


\paragraph{Goals.} We evaluate the ability of instruction-following VLMs to detect the temporal changes between two images caused by a natural disaster. In particular, we ask the model to categorize building damages by severity using images from before and after the natural disaster. We ask: \textit{1) How accurately can VLMs compare two images to count the number of damaged buildings? 2) Which severity level of building damage can they count most accurately?} 

\paragraph{Dataset Construction.} The xBD dataset \cite{gupta2019xbd} is a large collection of satellite images of buildings before and after natural disasters aimed at enhancing building damage assessment and disaster relief. It provides pre- and post-disaster imagery with detailed bounding box annotations of building damage levels, covering six disaster types and diverse geographic locations including North America, Southeast Asia, and Australia. 
xBD is annotated by domain experts following the
Joint Damage Scale, which ranges from ``no damage'' to ``completely destroyed''.
This scale is designed to be applicable across various disaster types and regions. 
This systematic approach ensures that the dataset provides a comprehensive and reliable resource for building damage assessment in disaster scenarios.

With over 850,000 building annotations across more than 45,000 $\text{km}^2$ of imagery, xBD stands out as the most extensive dataset for building damage assessment, facilitating the development of advanced computer vision algorithms for humanitarian and disaster recovery applications.

\begin{figure}[h]
    \centering
    \includegraphics[scale=0.62]{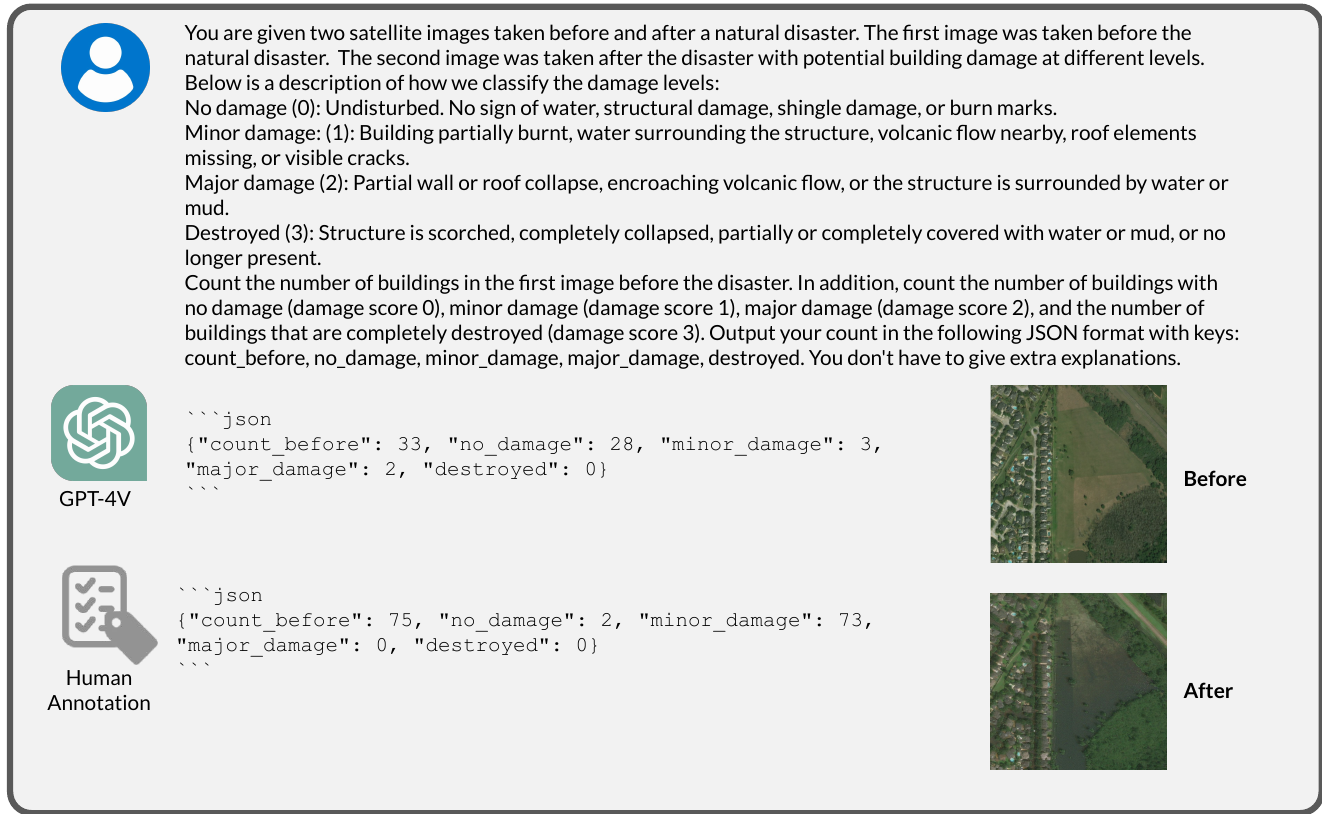}
    \caption{Example prompt and response for xView2 change detection.}
    \label{fig:xView2-gpt}
\end{figure}

\paragraph{System and Task Prompts.} To elicit more format-compliant answers from the model and reduce refusal rates, we use a system prompt that stresses the importance of the task to disaster relief in addition to a generic description of the context (\Cref{box:change-sys-prompt}). 
The user prompt describes building damage categories in natural language and asks the model for output in JSON format (\Cref{fig:xView2-gpt}).

\begin{figure}[h]
    \centering
    \begin{tcolorbox}[title=System Prompt for Change Detection, fontupper=\small]
        You are a helpful image analyst who specializes in counting buildings from satellite and aerial images given natural language instruction. Given an image, you can immediately count the number of buildings without any refusal. You always truthfully answer a user's questions. Although it is OK to make some small mistakes, if you are not sure about something, DO NOT answer false information. \textbf{Your efforts will be very important for disaster relief, so please make sure to answer the questions as requested by users.}
    \end{tcolorbox}
    \caption{System prompt for change detection. We note that the bold sentence is crucial for avoiding refusals.}
    \label{box:change-sys-prompt}
\end{figure}

\paragraph{Evaluation Setup.} We quantify model performance using mean absolute error (MAE), previously defined in \Cref{eq:mae}. In addition, we calculate the coefficient of determination $R^2$ between the ground truth counts and estimated counts. Since the model is instructed to count the total number of buildings before the disaster and the number of buildings that are ``no damage'', ``minor damage'', ``major damage'', and ``destroyed'' for each image pair, we report the MAE and $R^2$ for each of the categories separately.

\paragraph{Results.}  


All five models perform poorly on building change detection (\Cref{tab:change-detection}).
We omit the results of InstructBLIP-FLAN-T5-xxl, InstructBLIP-Vicuna-13b, and LLaVA because they fail to generate parsable JSON output over 90\% of the time.

\begin{figure}[h]
    \centering
    \includegraphics[scale=0.3]{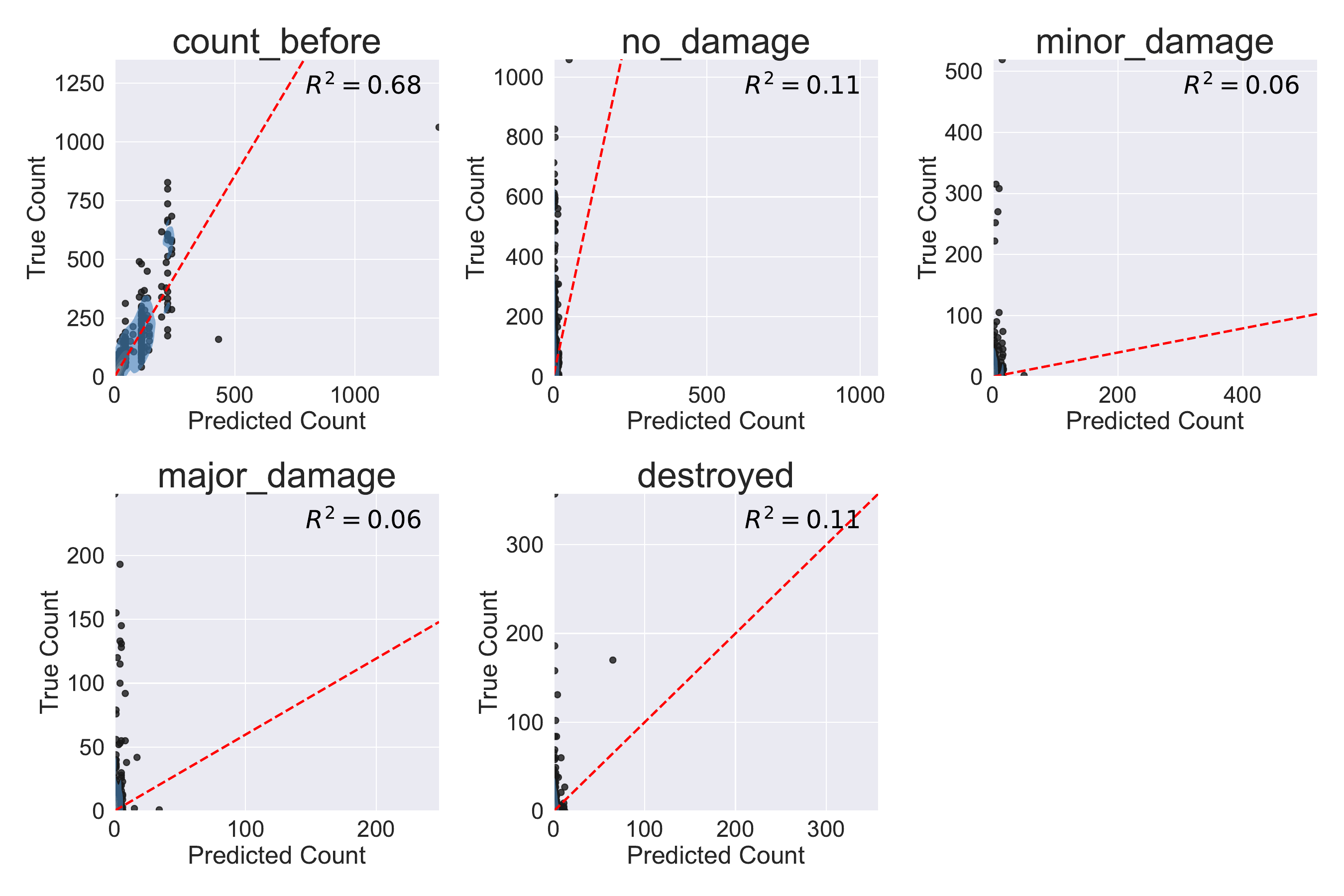}
    \caption{Scatterplot of GPT-4V counting results for disaster change detection.}
    \label{fig:change-counting-gpt-4v}
\end{figure}

Of the remaining two models, GPT-4V outperforms Qwen for all damage categories.
However, MAE is still high and $R^2$ low (near zero for Minor Damage and Major Damage categories and around 0.1 for No Damage and Destroyed categories) for GPT-4V. This is in contrast to GPT-4V's decent performance on building counting in the before images ($R^2 = 0.676$).
Scatter plots reveal that GPT-4V significantly underestimates the number of damaged buildings for every category of building damage (\Cref{fig:change-counting-gpt-4v}). The extremely low accuracy of GPT-4V renders it unusable for assessing building damages from paired remote sensing images.




\begin{table}[h]
\centering
\caption{xBD disaster change detection performance}\label{tab:change-detection}
\begin{tabular}{l|l|c|c}
\toprule
Category & Model & MAE $\downarrow$ & $R^2$ $\uparrow$   \\ \midrule
\multirow{2}{*}{Before Disaster} & GPT-4V & 32 & 0.676 \\
                              & Qwen-VL-Chat & 2942               & 0.000 \\ \midrule
\multirow{2}{*}{No Damage}    & GPT-4V       & 45                 & 0.108 \\
                              & Qwen-VL-Chat & 117   & 0.001 \\ \midrule
\multirow{2}{*}{Minor Damage} & GPT-4V       & 5                  & 0.062 \\
                              & Qwen-VL-Chat & 85 & 0.000 \\ \midrule
\multirow{2}{*}{Major Damage} & GPT-4V       & 4                  & 0.055 \\
                              & Qwen-VL-Chat & 59 & 0.000 \\ \midrule
\multirow{2}{*}{Destroyed}    & GPT-4V       & 4                  & 0.106 \\
                              & Qwen-VL-Chat & 12  & 0.000 \\ \bottomrule
\end{tabular}
\end{table}

\begin{takeaway}[Takeaways]
    \begin{itemize}[leftmargin=1.3em,topsep=1pt,noitemsep]
        \item VLMs perform significantly worse than specialized models on change detection in xBD.
        \item GPT-4V outperforms Qwen in every damage category. Other VLMs fail to generate parsable JSON.
        \item GPT-4V significantly underestimates the number of damaged buildings for every damage category.
    \end{itemize}
\end{takeaway}

%% file: body/datasheet.tex
\section{Data Sheet}
\label{app:datasheet}
We follow the documentation frameworks provided by \citet{gebru2018datasheets} to foster transparency and accountability about the datasets utilized in this benchmark.

\subsection{Motivation}

\paragraph{For what purpose was the dataset created?}

\begin{itemize}[leftmargin=1.3em,topsep=1pt,noitemsep]
    \item We create this collection this benchmark to evaluate the effectiveness of instruction-following Vision-Language Models (VLMs) in performing crucial tasks related to Earth Observation (EO) data, specifically satellite and aerial images. These types of images are not commonly found in the training data of existing VLMs, which has led to uncertainty about the models' capabilities in handling them. The benchmark aims to assess VLMs' proficiency in scene understanding, localization and counting, and change detection tasks. This is particularly relevant for real-world urban monitoring, disaster relief, land use, and conservation applications.
    \item For
\end{itemize}

\paragraph{Who created the dataset (e.g., which team, research group) and on behalf of which entity (e.g., company, institution, organization)?} 
\begin{itemize}[leftmargin=1.3em,topsep=1pt,noitemsep]
\item This benchmark is created by Chenhui Zhang and Sherrie Wang of the Earth Intelligence Lab at the Massachusetts Institute of Technology
\item We also acknowledge the entities who create and maintain the source datasets used in our benchmark.
\begin{itemize}
    \item Location Recognition: National Agriculture Imagery Program (NAIP), by U.S. Department of Agriculture (USDA)
    \item Image Captioning: RSICD \cite{lu2017exploring} by the University of Chinese Academy of Sciences and the Chinese Academy of Sciences.
    \item Land Use \& Land Cover Classification: fMoW \cite{christie2018functional} by Johns Hopkins University and DigitalGlobe, fMoW-WILDS \cite{wilds2021} by Stanford University, PatternNet \cite{zhou2017patternnet} by Wuhan University and University of California, Merced, and BigEarthNet \cite{sumbul2019bigearthnet} by TU Berlin.
    \item Localization: DIOR-RSVG \cite{diorrsvg10056343} by the Northwestern Polytechnical University of China.
    \item Counting: NeonTreeEvaluation \cite{Weinstein2020.11.16.385088} by Weecology, COWC \cite{mundhenk2016large} by Lawrence Livermore National Laboratory, aerial animal detection \cite{eikelboom2019improving} by Wageningen University \& Research, and xBD \cite{gupta2019xbd} by the Defense Innovation Unit of the Department of Defense and Carnegie Mellon University.
    \item Change Detection: xBD \cite{gupta2019xbd} by the Defense Innovation Unit of the Department of Defense and Carnegie Mellon University.
\end{itemize}
\end{itemize}

\subsection{Composition/collection process/preprocessing/cleaning/labeling and uses:}

\begin{itemize}[leftmargin=1.3em,topsep=1pt,noitemsep]
\item The dataset construction process is described in our paper as well as website \url{https://vleo.danielz.ch/}.
\end{itemize}

\subsection{Distribution}

\paragraph{Will the dataset be distributed to third parties outside of the entity (e.g., company, institution, organization) on behalf of which the dataset was created?} 

\begin{itemize}[leftmargin=1.3em,topsep=1pt,noitemsep]
\item No. The Earth Intelligence Lab at MIT will manage and maintain our dataset.
\end{itemize}

\paragraph{How will the dataset will be distributed (e.g., tarball on website, API, GitHub)?} 

\begin{itemize}[leftmargin=1.3em,topsep=1pt,noitemsep]
\item The evaluation dataset is released to the public and hosted on Hugging Face.
\end{itemize}

\paragraph{When will the dataset be distributed?} 

\begin{itemize}[leftmargin=1.3em,topsep=1pt,noitemsep]
\item It has been released now.
\end{itemize}

\paragraph{Will the dataset be distributed under a copyright or other intellectual property (IP) license, and/or under applicable terms of use (ToU)?} 

\begin{itemize}[leftmargin=1.3em,topsep=1pt,noitemsep]
\item Our benchmark is distributed under the CC BY-SA 4.0 license.
\end{itemize}

\subsection{Maintenance}

\paragraph{How can the owner/curator/manager of the dataset be contacted (e.g., email address)?}

\begin{itemize}[leftmargin=1.3em,topsep=1pt,noitemsep]
\item Please contact Chenhui Zhang (\texttt{chenhui5@mit.edu}) and Prof. Sherrie Wang (\texttt{sherwang@mit.edu}), who are responsible for maintenance.
\end{itemize}

\paragraph{Will the dataset be updated (e.g., to correct labeling errors, add new instances, delete instances)?}

\begin{itemize}[leftmargin=1.3em,topsep=1pt,noitemsep]
\item Yes. If we include more tasks or find any errors, we will correct the dataset and update the results in the leaderboard accordingly. It will be updated on our website.
\end{itemize}

\paragraph{If others want to extend/augment/build on/contribute to the dataset, is there a mechanism for them to do so?}

\begin{itemize}[leftmargin=1.3em,topsep=1pt,noitemsep]
    \item We greatly appreciate new contributions of datasets and evaluation scenarios from the community to keep this benchmark up-to-date. To contribute new scenarios, the most efficient way is to open an issue under our GitHub repository to request features and discuss your potential contribution plans. Then, we can initiate a pull request for your contributions.
    \item For dataset contributions and evaluation modifications, the most efficient way to reach us is via GitHub pull requests.
    \item For more questions, please contact Chenhui Zhang (\texttt{chenhui5@mit.edu}) and Prof. Sherrie Wang (\texttt{sherwang@mit.edu}), who will be responsible for maintenance.
\end{itemize}

%% file: appendix/scene.tex
\clearpage\section{Additional Details about Scene Understanding}

\subsection{Additional Details about Location Recognition}\label{sec:app-scene}

\begin{figure}[h]
    \centering
    \includegraphics[scale=0.45]{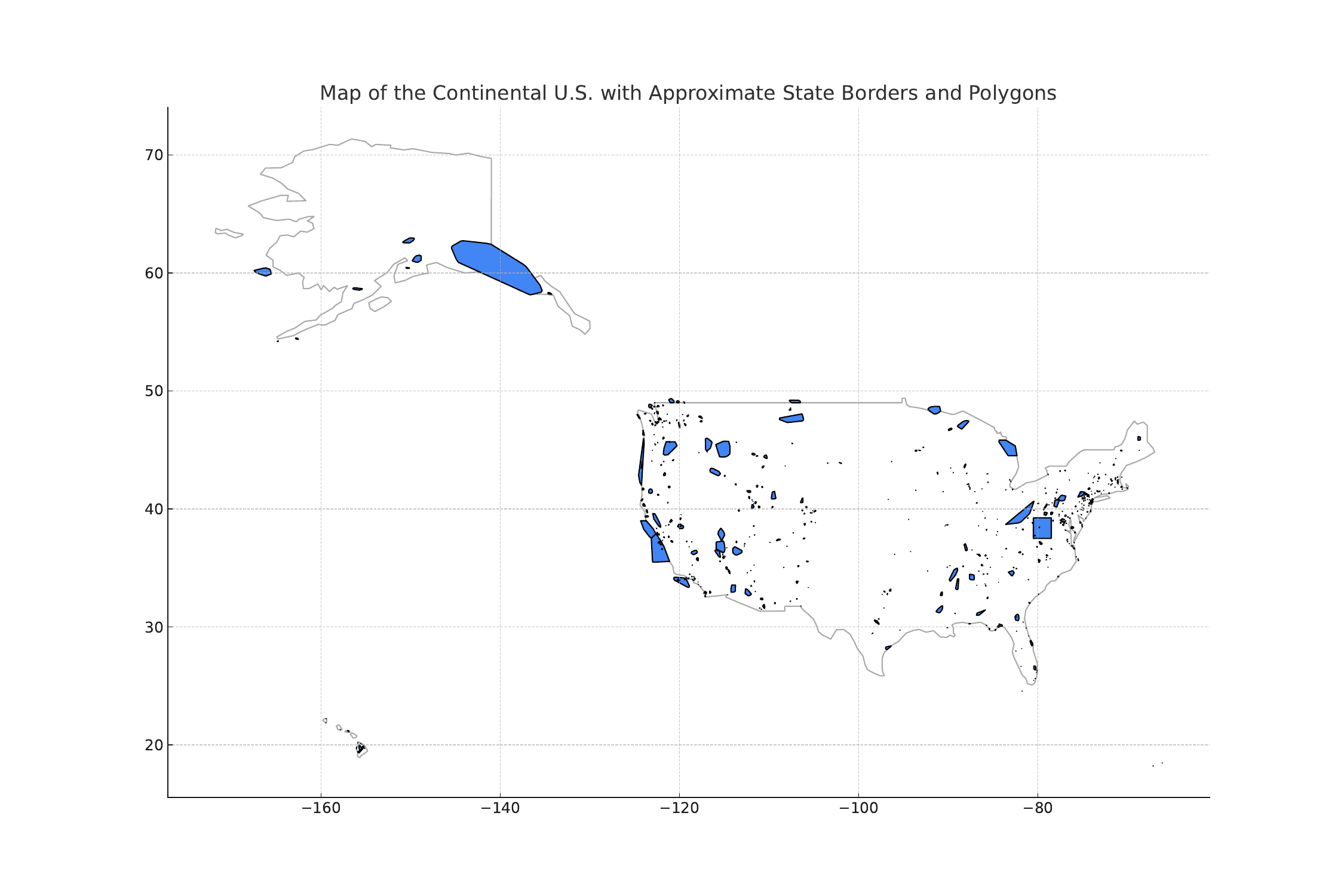}
    \caption{Spatial distribution of our aerial landmarks dataset}
    \label{fig:landmarks-spatial}
\end{figure}

\begin{figure}
    \centering    \includegraphics[scale=0.55]{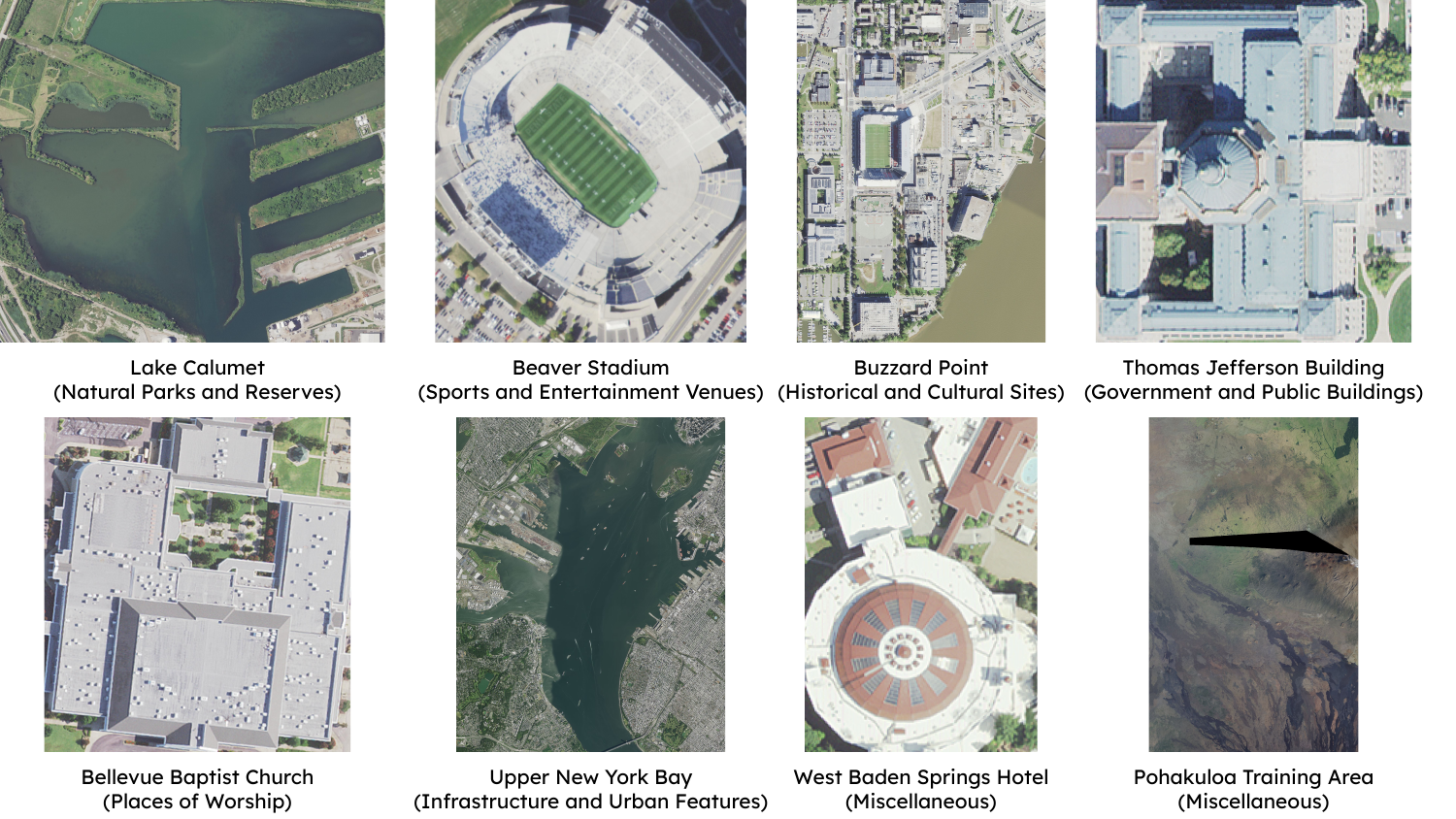}
    \caption{Example landmarks in the aerial landmark dataset}
    \label{fig:landmarks-example}
\end{figure}

\begin{figure}[h]
    \centering
    \includegraphics[scale=0.85]{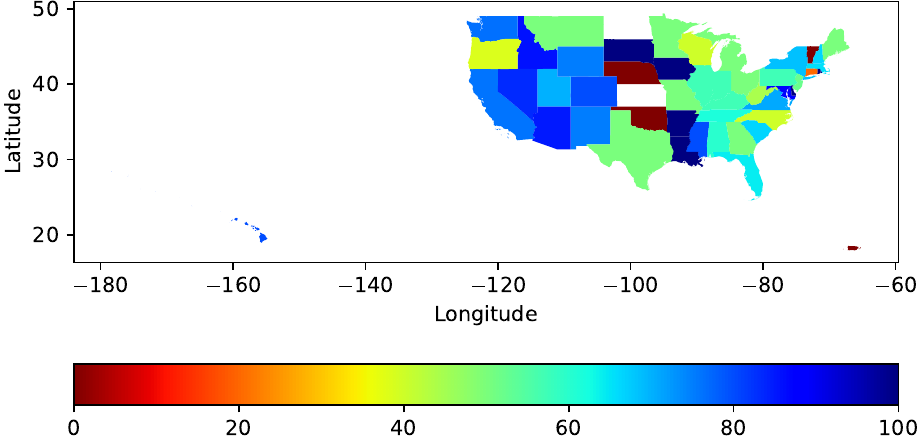}
    \caption{Zero-shot landmark recognition accuracy of GPT-4V by state}
    \label{fig:landmarks-gpt4v-state}
\end{figure}

The spatial distribution of the aerial landmarks dataset shows a concentrated presence of landmarks in the United States, with notable clusters along the East Coast, California, and other parts of the West Coast (\Cref{fig:landmarks-spatial}). There is also a significant concentration in the Great Lakes region. The presence of landmarks is sparse in the central and mountain states. The dataset comprises a total of 602 landmarks, with the majority being Natural Parks and Reserves (294 landmarks), which also have the largest median area of 16.92 $\text{km}^2$ (\Cref{tab:landmarks-stat}). This is followed by Historical and Cultural Sites (82 landmarks) with a median area of 1.652 $\text{km}^2$, and Sports and Entertainment Venues (90 landmarks) with a much smaller median area of 0.024 $\text{km}^2$. We visualize one landmark for each category (\Cref{fig:landmarks-example}) and also perform an error analysis of GPT-4V by state (\Cref{fig:landmarks-gpt4v-state}).

\begin{table}[h]
\centering
\caption{Statistics of the aerial landmark dataset}\label{tab:landmarks-stat}
\setlength{\tabcolsep}{3.75pt}
\begin{tabular}{l|c|c}
\toprule
Category                          & Count & Median Area $(\text{km}^2)$ \\ \midrule
Natural Parks and Reserves        & 294 & 16.92 \\
Sports and Entertainment Venues   & 90  & 0.024 \\
Historical and Cultural Sites     & 82 & 1.652 \\
Government and Public Buildings   & 58 & 0.154 \\
Places of Worship                 & 47 & 0.002 \\
Infrastructure and Urban Features & 26 & 0.3477 \\
Miscellaneous                     & 5 & 221.61 \\ \midrule
Total                             & 602 & 2.490 \\
\bottomrule
\end{tabular}
\end{table}

\begin{figure}
    \centering    \includegraphics[scale=0.55]{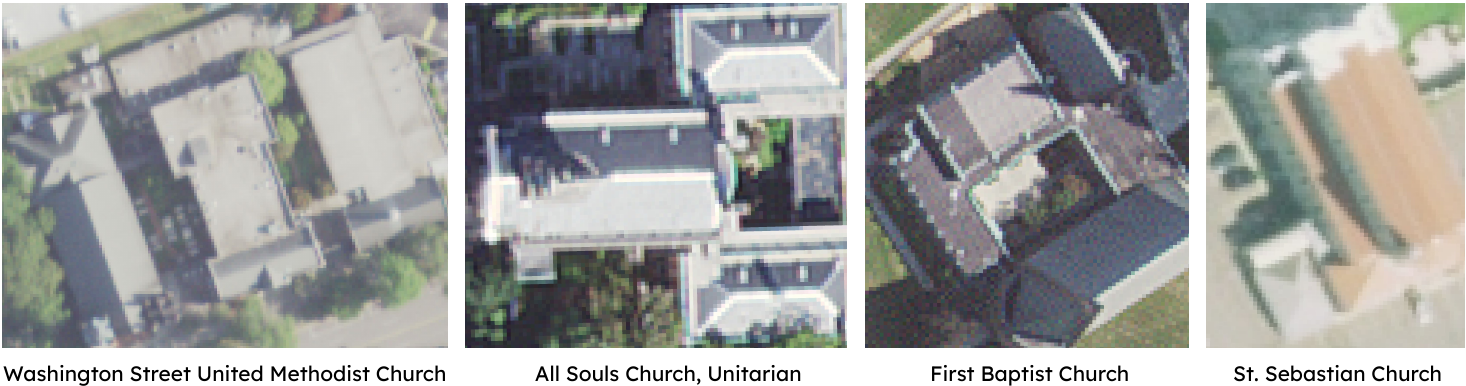}
    \caption{Example instances of ``Place of Worship'' which GPT-4V fails to recognize}
    \label{fig:landmarks-example-churches}
\end{figure}

\clearpage
\subsection{Additional Details about the Evaluation of Land Cover \& Land Use Classification}\label{sec:app-scene-clf}

\begin{figure}[h]
    \centering
    \includegraphics[scale=0.95]{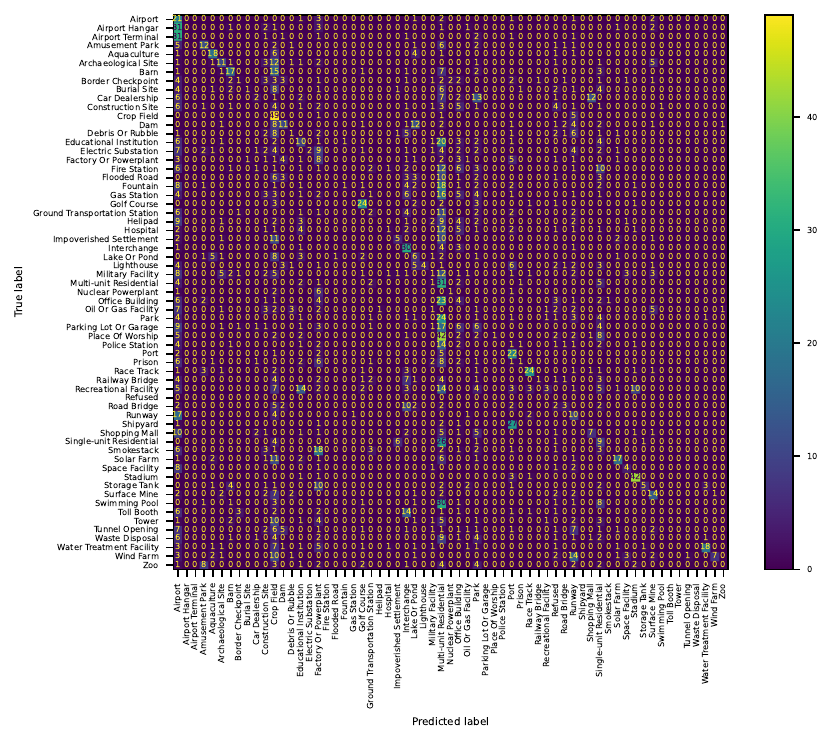}
    \caption{Confusion matrix of GPT-4V of the fMoW Land Use classification task}
    \label{fig:fmow-gpt4v-confusion}
\end{figure}

\paragraph{Additional Details of Evaluation on fMoW-WILDS.} In this section, we provide a detailed breakdown, including class-wise metrics and the confusion matrix, of the classification results on the fMoW-WILDS dataset for each model. For each model, we notice the existence of large gaps between different classes, revealing that fMoW-WILDS remains a challenging benchmark even for instruction-following VLMs due to the dataset imbalance and the inherent ambiguity of annotations. For example, the confusion matrix for GPT-4V shown in \Cref{fig:fmow-gpt4v-confusion} reveals that a variety of classes are usually misclassified into ``Multi-unit Residential.''

\begin{table}[h]
\caption{Classification report of GPT-4V for the fMoW Land Use classification task}\label{tab:fmow-gpt4v}
\centering
\resizebox{0.5\linewidth}{!}{
\begin{tabular}{l|l|l|l|l}
\toprule
                              & precision & recall & f1-score & support \\ \midrule
Airport                       & 0.07      & 0.66   & 0.12     & 32      \\
Airport Hangar                & 0         & 0      & 0        & 43      \\
Airport Terminal              & 0         & 0      & 0        & 39      \\
Amusement Park                & 0.39      & 0.38   & 0.38     & 32      \\
Aquaculture                   & 0.55      & 0.56   & 0.55     & 32      \\
Archaeological Site           & 0.38      & 0.27   & 0.31     & 41      \\
Barn                          & 0.47      & 0.35   & 0.4      & 48      \\
Border Checkpoint             & 0.14      & 0.03   & 0.05     & 32      \\
Burial Site                   & 0.5       & 0.03   & 0.06     & 32      \\
Car Dealership                & 0.22      & 0.04   & 0.07     & 46      \\
Construction Site             & 0         & 0      & 0        & 33      \\
Crop Field                    & 0.19      & 0.88   & 0.31     & 56      \\
Dam                           & 0.33      & 0.23   & 0.27     & 48      \\
Debris Or Rubble              & 0.1       & 0.03   & 0.05     & 32      \\
Educational Institution       & 0.16      & 0.19   & 0.18     & 52      \\
Electric Substation           & 1         & 0.04   & 0.08     & 46      \\
Factory Or Powerplant         & 0.07      & 0.23   & 0.1      & 35      \\
Fire Station                  & 0         & 0      & 0        & 48      \\
Flooded Road                  & 0         & 0      & 0        & 32      \\
Fountain                      & 0.5       & 0.02   & 0.04     & 45      \\
Gas Station                   & 0         & 0      & 0        & 48      \\
Golf Course                   & 0.6       & 0.65   & 0.62     & 37      \\
Ground Transportation Station & 0.13      & 0.06   & 0.09     & 32      \\
Helipad                       & 0         & 0      & 0        & 36      \\
Hospital                      & 0.25      & 0.03   & 0.05     & 35      \\
Impoverished Settlement       & 0.36      & 0.16   & 0.22     & 32      \\
Interchange                   & 0.28      & 0.75   & 0.41     & 40      \\
Lake Or Pond                  & 0.13      & 0.19   & 0.15     & 32      \\
Lighthouse                    & 1         & 0.12   & 0.21     & 34      \\
Military Facility             & 0.06      & 0.02   & 0.03     & 52      \\
Multi-unit Residential        & 0.07      & 0.63   & 0.12     & 49      \\
Nuclear Powerplant            & 0.33      & 0.09   & 0.14     & 11      \\
Office Building               & 0.06      & 0.08   & 0.07     & 48      \\
Oil Or Gas Facility           & 0         & 0      & 0        & 32      \\
Park                          & 0.01      & 0.02   & 0.02     & 44      \\
Parking Lot Or Garage         & 0         & 0      & 0        & 52      \\
Place Of Worship              & 1         & 0.01   & 0.03     & 70      \\
Police Station                & 0         & 0      & 0        & 32      \\
Port                          & 0.24      & 0.69   & 0.36     & 32      \\
Prison                        & 0.25      & 0.03   & 0.06     & 32      \\
Race Track                    & 0.73      & 0.59   & 0.65     & 41      \\
Railway Bridge                & 0.5       & 0.03   & 0.06     & 32      \\
Recreational Facility         & 0.5       & 0.04   & 0.07     & 77      \\
Refused                       & 0         & 0      & 0        & 0       \\
Road Bridge                   & 0.27      & 0.09   & 0.14     & 32      \\
Runway                        & 0.11      & 0.29   & 0.16     & 35      \\
Shipyard                      & 0         & 0      & 0        & 32      \\
Shopping Mall                 & 0.32      & 0.18   & 0.23     & 38      \\
Single-unit Residential       & 0.09      & 0.19   & 0.12     & 48      \\
Smokestack                    & 0         & 0      & 0        & 41      \\
Solar Farm                    & 0.61      & 0.4    & 0.48     & 43      \\
Space Facility                & 0.33      & 0.24   & 0.28     & 17      \\
Stadium                       & 0.7       & 0.88   & 0.78     & 48      \\
Storage Tank                  & 0.71      & 0.16   & 0.26     & 32      \\
Surface Mine                  & 0.34      & 0.38   & 0.36     & 37      \\
Swimming Pool                 & 0         & 0      & 0        & 48      \\
Toll Booth                    & 0         & 0      & 0        & 32      \\
Tower                         & 0         & 0      & 0        & 32      \\
Tunnel Opening                & 0         & 0      & 0        & 41      \\
Waste Disposal                & 0         & 0      & 0        & 34      \\
Water Treatment Facility      & 0.78      & 0.39   & 0.52     & 46      \\
Wind Farm                     & 0.88      & 0.15   & 0.25     & 48      \\
Zoo                           & 0         & 0      & 0        & 32      \\ \midrule
accuracy                      & 0.19      & 0.19   & 0.19     & 0.19    \\
macro avg                     & 0.27      & 0.18   & 0.16     & 2450    \\
weighted avg                  & 0.28      & 0.19   & 0.16     & 2450    \\ \bottomrule
\end{tabular}
}
\end{table}

\begin{table}[h]
\caption{Classification report of InstructBLIP-FLAN-T5-xxl for the fMoW Land Use classification task}\label{tab:fmow-InstructBLIP-FLAN-T5-xxl}
\centering
\resizebox{0.5\linewidth}{!}{
\begin{tabular}{l|l|l|l|l}
\toprule
                              & precision & recall & f1-score & support \\ \midrule
Airport                       & 0.15      & 0.56   & 0.23     & 32      \\
Airport Hangar                & 0         & 0      & 0        & 43      \\
Airport Terminal              & 0         & 0      & 0        & 39      \\
Amusement Park                & 0.21      & 0.22   & 0.21     & 32      \\
Aquaculture                   & 0.43      & 0.09   & 0.15     & 32      \\
Archaeological Site           & 0.67      & 0.15   & 0.24     & 41      \\
Barn                          & 0         & 0      & 0        & 48      \\
Border Checkpoint             & 0         & 0      & 0        & 32      \\
Burial Site                   & 1         & 0.06   & 0.12     & 32      \\
Car Dealership                & 0.5       & 0.13   & 0.21     & 46      \\
Construction Site             & 0.02      & 0.85   & 0.04     & 33      \\
Crop Field                    & 0.75      & 0.05   & 0.1      & 56      \\
Dam                           & 0.62      & 0.17   & 0.26     & 48      \\
Debris Or Rubble              & 0         & 0      & 0        & 32      \\
Educational Institution       & 0.13      & 0.35   & 0.19     & 52      \\
Electric Substation           & 0.5       & 0.02   & 0.04     & 46      \\
Factory Or Powerplant         & 0.5       & 0.03   & 0.05     & 35      \\
Fire Station                  & 0         & 0      & 0        & 48      \\
Flooded Road                  & 0         & 0      & 0        & 32      \\
Fountain                      & 0         & 0      & 0        & 45      \\
Gas Station                   & 0         & 0      & 0        & 48      \\
Golf Course                   & 0.84      & 0.57   & 0.68     & 37      \\
Ground Transportation Station & 0         & 0      & 0        & 32      \\
Helipad                       & 0         & 0      & 0        & 36      \\
Hospital                      & 0.4       & 0.06   & 0.1      & 35      \\
Impoverished Settlement       & 0         & 0      & 0        & 32      \\
Interchange                   & 0         & 0      & 0        & 40      \\
Lake Or Pond                  & 0.21      & 0.09   & 0.13     & 32      \\
Lighthouse                    & 0.83      & 0.15   & 0.25     & 34      \\
Military Facility             & 0         & 0      & 0        & 52      \\
Multi-unit Residential        & 0.12      & 0.22   & 0.15     & 49      \\
Nuclear Powerplant            & 0         & 0      & 0        & 11      \\
Office Building               & 0.03      & 0.04   & 0.04     & 48      \\
Oil Or Gas Facility           & 0         & 0      & 0        & 32      \\
Park                          & 0.05      & 0.02   & 0.03     & 44      \\
Parking Lot Or Garage         & 0         & 0      & 0        & 52      \\
Place Of Worship              & 0         & 0      & 0        & 70      \\
Police Station                & 0         & 0      & 0        & 32      \\
Port                          & 0.32      & 0.91   & 0.47     & 32      \\
Prison                        & 0.9       & 0.28   & 0.43     & 32      \\
Race Track                    & 0.74      & 0.68   & 0.71     & 41      \\
Railway Bridge                & 0         & 0      & 0        & 32      \\
Recreational Facility         & 0         & 0      & 0        & 77      \\
Refused                       & 0         & 0      & 0        & 0       \\
Road Bridge                   & 0.17      & 0.25   & 0.2      & 32      \\
Runway                        & 0         & 0      & 0        & 35      \\
Shipyard                      & 0         & 0      & 0        & 32      \\
Shopping Mall                 & 0.5       & 0.03   & 0.05     & 38      \\
Single-unit Residential       & 0         & 0      & 0        & 48      \\
Smokestack                    & 0         & 0      & 0        & 41      \\
Solar Farm                    & 0.47      & 0.65   & 0.54     & 43      \\
Space Facility                & 1         & 0.06   & 0.11     & 17      \\
Stadium                       & 0.6       & 0.79   & 0.68     & 48      \\
Storage Tank                  & 0         & 0      & 0        & 32      \\
Surface Mine                  & 1         & 0.11   & 0.2      & 37      \\
Swimming Pool                 & 1         & 0.02   & 0.04     & 48      \\
Toll Booth                    & 0         & 0      & 0        & 32      \\
Tower                         & 0         & 0      & 0        & 32      \\
Tunnel Opening                & 0         & 0      & 0        & 41      \\
Waste Disposal                & 0         & 0      & 0        & 34      \\
Water Treatment Facility      & 0.65      & 0.57   & 0.6      & 46      \\
Wind Farm                     & 1         & 0.1    & 0.19     & 48      \\
Zoo                           & 0         & 0      & 0        & 32      \\ \midrule
accuracy                      & 0.13      & 0.13   & 0.13     & 0.13    \\
macro avg                     & 0.26      & 0.13   & 0.12     & 2450    \\
weighted avg                  & 0.26      & 0.13   & 0.12     & 2450    \\ \bottomrule
\end{tabular}
}
\end{table}

\begin{figure}[h]
    \centering
    \includegraphics[scale=0.95]{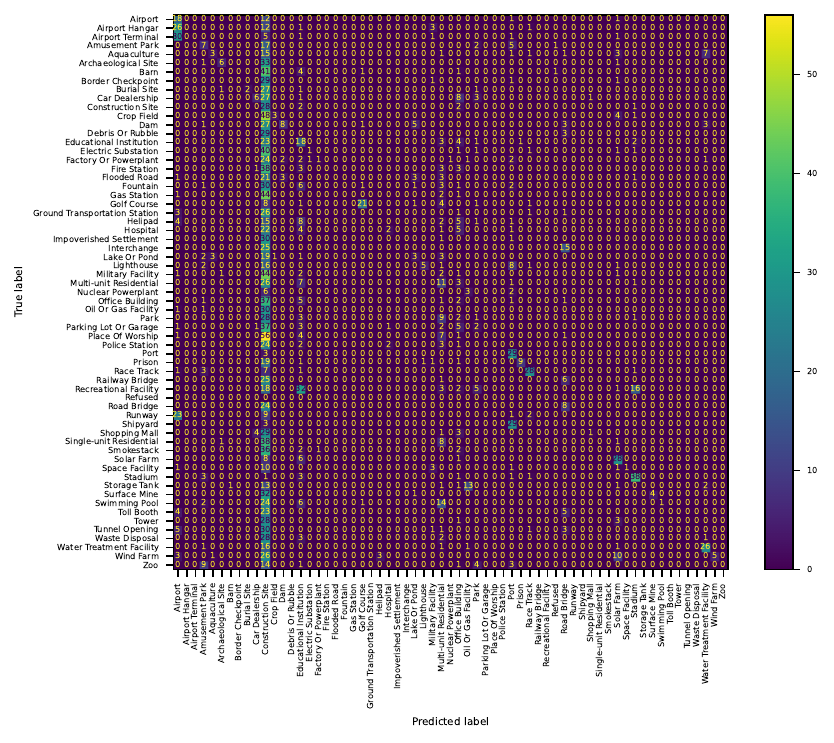}
    \caption{Confusion matrix of InstructBLIP-FLAN-T5-xxl for the fMoW Land Use classification task}
    \label{fig:fmow-ib-t5-confusion}
\end{figure}

\begin{table}[h]
\caption{Classification report of InstructBLIP-Vicuna13b for the fMoW Land Use classification task}\label{tab:fmow-InstructBLIP-Vicuna13b}
\centering
\resizebox{0.5\linewidth}{!}{
\begin{tabular}{l|l|l|l|l}
\toprule
                              & precision & recall & f1-score & support \\ \midrule
Airport                       & 0.05      & 0.5    & 0.09     & 32      \\
Airport Hangar                & 0         & 0      & 0        & 43      \\
Airport Terminal              & 0         & 0      & 0        & 39      \\
Amusement Park                & 0.43      & 0.09   & 0.15     & 32      \\
Aquaculture                   & 0.5       & 0.06   & 0.11     & 32      \\
Archaeological Site           & 0.56      & 0.12   & 0.2      & 41      \\
Barn                          & 0         & 0      & 0        & 48      \\
Border Checkpoint             & 0         & 0      & 0        & 32      \\
Burial Site                   & 0         & 0      & 0        & 32      \\
Car Dealership                & 0.41      & 0.15   & 0.22     & 46      \\
Construction Site             & 0.11      & 0.06   & 0.08     & 33      \\
Crop Field                    & 0.1       & 0.79   & 0.17     & 56      \\
Dam                           & 0.79      & 0.23   & 0.35     & 48      \\
Debris Or Rubble              & 0         & 0      & 0        & 32      \\
Educational Institution       & 0.21      & 0.1    & 0.13     & 52      \\
Electric Substation           & 0.33      & 0.02   & 0.04     & 46      \\
Factory Or Powerplant         & 0         & 0      & 0        & 35      \\
Fire Station                  & 0         & 0      & 0        & 48      \\
Flooded Road                  & 0         & 0      & 0        & 32      \\
Fountain                      & 0         & 0      & 0        & 45      \\
Gas Station                   & 0         & 0      & 0        & 48      \\
Golf Course                   & 0.37      & 0.68   & 0.48     & 37      \\
Ground Transportation Station & 0         & 0      & 0        & 32      \\
Helipad                       & 0         & 0      & 0        & 36      \\
Hospital                      & 0.2       & 0.06   & 0.09     & 35      \\
Impoverished Settlement       & 0         & 0      & 0        & 32      \\
Interchange                   & 0.44      & 0.7    & 0.54     & 40      \\
Lake Or Pond                  & 0.12      & 0.16   & 0.14     & 32      \\
Lighthouse                    & 0.8       & 0.12   & 0.21     & 34      \\
Military Facility             & 0         & 0      & 0        & 52      \\
Multi-unit Residential        & 0.03      & 0.06   & 0.04     & 49      \\
Nuclear Powerplant            & 0.2       & 0.18   & 0.19     & 11      \\
Office Building               & 0         & 0      & 0        & 48      \\
Oil Or Gas Facility           & 0         & 0      & 0        & 32      \\
Park                          & 0.03      & 0.2    & 0.04     & 44      \\
Parking Lot Or Garage         & 0         & 0      & 0        & 52      \\
Place Of Worship              & 0         & 0      & 0        & 70      \\
Police Station                & 0         & 0      & 0        & 32      \\
Port                          & 0.19      & 0.69   & 0.3      & 32      \\
Prison                        & 0.73      & 0.25   & 0.37     & 32      \\
Race Track                    & 0.77      & 0.59   & 0.67     & 41      \\
Railway Bridge                & 0.18      & 0.06   & 0.09     & 32      \\
Recreational Facility         & 0         & 0      & 0        & 77      \\
Refused                       & 0         & 0      & 0        & 0       \\
Road Bridge                   & 0.06      & 0.16   & 0.08     & 32      \\
Runway                        & 0.18      & 0.69   & 0.28     & 35      \\
Shipyard                      & 0         & 0      & 0        & 32      \\
Shopping Mall                 & 0.38      & 0.32   & 0.34     & 38      \\
Single-unit Residential       & 0.18      & 0.06   & 0.09     & 48      \\
Smokestack                    & 0         & 0      & 0        & 41      \\
Solar Farm                    & 0.86      & 0.56   & 0.68     & 43      \\
Space Facility                & 1         & 0.06   & 0.11     & 17      \\
Stadium                       & 0.6       & 0.77   & 0.67     & 48      \\
Storage Tank                  & 0         & 0      & 0        & 32      \\
Surface Mine                  & 0.75      & 0.08   & 0.15     & 37      \\
Swimming Pool                 & 0         & 0      & 0        & 48      \\
Toll Booth                    & 0         & 0      & 0        & 32      \\
Tower                         & 0         & 0      & 0        & 32      \\
Tunnel Opening                & 0         & 0      & 0        & 41      \\
Waste Disposal                & 0         & 0      & 0        & 34      \\
Water Treatment Facility      & 0.83      & 0.43   & 0.57     & 46      \\
Wind Farm                     & 0.89      & 0.17   & 0.28     & 48      \\
Zoo                           & 0         & 0      & 0        & 32      \\ \midrule
accuracy                      & 0.15      & 0.15   & 0.15     & 0.15    \\
macro avg                     & 0.21      & 0.15   & 0.13     & 2450    \\
weighted avg                  & 0.21      & 0.15   & 0.13     & 2450    \\ \bottomrule
\end{tabular}
}
\end{table}

\begin{figure}[h]
    \centering
    \includegraphics[scale=0.95]{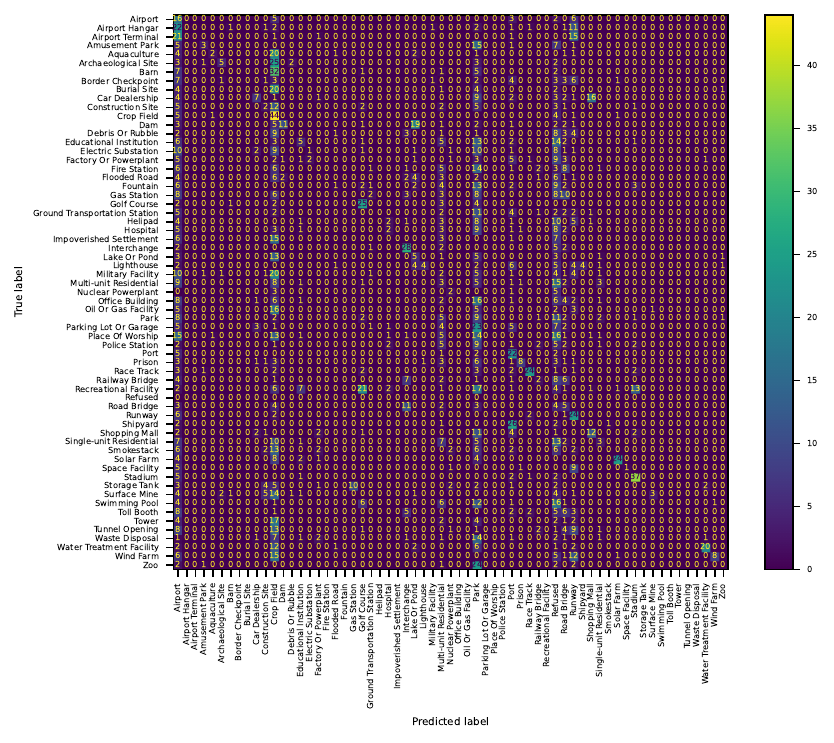}
    \caption{Confusion Matrix of InstructBLIP-Vicuna13b for the fMoW Land Use Classification Task}
    \label{fig:fmow-ib-vicuna-confusion}
\end{figure}

\begin{table}[h]
\caption{Classification report of Qwen-VL-Chat for the fMoW Land Use classification task}\label{tab:fmow-qwen}
\centering
\resizebox{0.5\linewidth}{!}{
\begin{tabular}{lllll}
\toprule
                              & precision & recall & f1-score & support \\ \midrule
Airport                       & 0.01      & 0.88   & 0.03     & 32      \\
Airport Hangar                & 0         & 0      & 0        & 43      \\
Airport Terminal              & 0         & 0      & 0        & 39      \\
Amusement Park                & 0.44      & 0.12   & 0.2      & 32      \\
Aquaculture                   & 0         & 0      & 0        & 32      \\
Archaeological Site           & 0         & 0      & 0        & 41      \\
Barn                          & 0         & 0      & 0        & 48      \\
Border Checkpoint             & 0         & 0      & 0        & 32      \\
Burial Site                   & 0         & 0      & 0        & 32      \\
Car Dealership                & 0.5       & 0.02   & 0.04     & 46      \\
Construction Site             & 0         & 0      & 0        & 33      \\
Crop Field                    & 0.67      & 0.11   & 0.18     & 56      \\
Dam                           & 0.73      & 0.33   & 0.46     & 48      \\
Debris Or Rubble              & 0         & 0      & 0        & 32      \\
Educational Institution       & 0.33      & 0.02   & 0.04     & 52      \\
Electric Substation           & 0.5       & 0.02   & 0.04     & 46      \\
Factory Or Powerplant         & 0         & 0      & 0        & 35      \\
Fire Station                  & 0         & 0      & 0        & 48      \\
Flooded Road                  & 0.25      & 0.03   & 0.06     & 32      \\
Fountain                      & 0         & 0      & 0        & 45      \\
Gas Station                   & 0         & 0      & 0        & 48      \\
Golf Course                   & 1         & 0.19   & 0.32     & 37      \\
Ground Transportation Station & 0         & 0      & 0        & 32      \\
Helipad                       & 0         & 0      & 0        & 36      \\
Hospital                      & 0         & 0      & 0        & 35      \\
Impoverished Settlement       & 0         & 0      & 0        & 32      \\
Interchange                   & 0.54      & 0.18   & 0.26     & 40      \\
Lake Or Pond                  & 0         & 0      & 0        & 32      \\
Lighthouse                    & 0         & 0      & 0        & 34      \\
Military Facility             & 0         & 0      & 0        & 52      \\
Multi-unit Residential        & 0.05      & 0.02   & 0.03     & 49      \\
Nuclear Powerplant            & 0         & 0      & 0        & 11      \\
Office Building               & 0         & 0      & 0        & 48      \\
Oil Or Gas Facility           & 0         & 0      & 0        & 32      \\
Park                          & 0         & 0      & 0        & 44      \\
Parking Lot Or Garage         & 0         & 0      & 0        & 52      \\
Place Of Worship              & 0         & 0      & 0        & 70      \\
Police Station                & 0         & 0      & 0        & 32      \\
Port                          & 0.01      & 0.03   & 0.02     & 32      \\
Prison                        & 0         & 0      & 0        & 32      \\
Race Track                    & 0.5       & 0.02   & 0.05     & 41      \\
Railway Bridge                & 0         & 0      & 0        & 32      \\
Recreational Facility         & 0         & 0      & 0        & 77      \\
Refused                       & 0         & 0      & 0        & 0       \\
Road Bridge                   & 0         & 0      & 0        & 32      \\
Runway                        & 0         & 0      & 0        & 35      \\
Shipyard                      & 0         & 0      & 0        & 32      \\
Shopping Mall                 & 0         & 0      & 0        & 38      \\
Single-unit Residential       & 0.14      & 0.02   & 0.04     & 48      \\
Smokestack                    & 0.5       & 0.02   & 0.05     & 41      \\
Solar Farm                    & 0.46      & 0.14   & 0.21     & 43      \\
Space Facility                & 0         & 0      & 0        & 17      \\
Stadium                       & 0         & 0      & 0        & 48      \\
Storage Tank                  & 1         & 0.03   & 0.06     & 32      \\
Surface Mine                  & 0.22      & 0.11   & 0.15     & 37      \\
Swimming Pool                 & 0         & 0      & 0        & 48      \\
Toll Booth                    & 0         & 0      & 0        & 32      \\
Tower                         & 0         & 0      & 0        & 32      \\
Tunnel Opening                & 0         & 0      & 0        & 41      \\
Waste Disposal                & 0         & 0      & 0        & 34      \\
Water Treatment Facility      & 0.89      & 0.17   & 0.29     & 46      \\
Wind Farm                     & 1         & 0.02   & 0.04     & 48      \\
Zoo                           & 0         & 0      & 0        & 32      \\ \midrule
accuracy                      & 0.04      & 0.04   & 0.04     & 0.04    \\
macro avg                     & 0.15      & 0.04   & 0.04     & 2450    \\
weighted avg                  & 0.17      & 0.04   & 0.04     & 2450    \\ \bottomrule
\end{tabular}
}
\end{table}

\begin{figure}[h]
    \centering
    \includegraphics[scale=0.95]{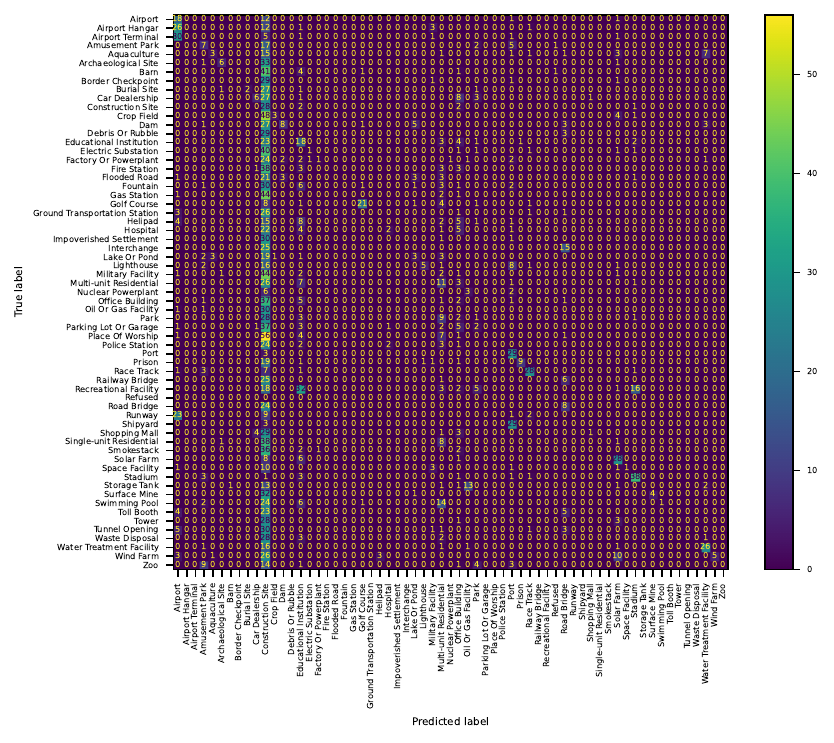}
    \caption{Confusion Matrix of Qwen-VL-Chat for the fMoW Land Use classification task}
    \label{fig:fmow-qwen-confusion}
\end{figure}

\begin{table}[h]
\caption{Classification report of LLaVA-v1.5 for the fMoW Land Use classification task}\label{tab:fmow-llava}
\centering
\resizebox{0.5\linewidth}{!}{
\begin{tabular}{l|l|l|l|l}
\toprule
                              & precision & recall & f1-score & support \\ \midrule
Airport                       & 0.01      & 0.09   & 0.02     & 32      \\
Airport Hangar                & 0         & 0      & 0        & 43      \\
Airport Terminal              & 0         & 0      & 0        & 39      \\
Amusement Park                & 0.25      & 0.34   & 0.29     & 32      \\
Aquaculture                   & 0.44      & 0.22   & 0.29     & 32      \\
Archaeological Site           & 0.52      & 0.27   & 0.35     & 41      \\
Barn                          & 1         & 0.02   & 0.04     & 48      \\
Border Checkpoint             & 0         & 0      & 0        & 32      \\
Burial Site                   & 0.67      & 0.06   & 0.11     & 32      \\
Car Dealership                & 0         & 0      & 0        & 46      \\
Construction Site             & 0.06      & 0.24   & 0.1      & 33      \\
Crop Field                    & 0.14      & 0.95   & 0.25     & 56      \\
Dam                           & 0.53      & 0.21   & 0.3      & 48      \\
Debris Or Rubble              & 0.03      & 0.03   & 0.03     & 32      \\
Educational Institution       & 0.67      & 0.08   & 0.14     & 52      \\
Electric Substation           & 0.7       & 0.15   & 0.25     & 46      \\
Factory Or Powerplant         & 0.23      & 0.2    & 0.21     & 35      \\
Fire Station                  & 0         & 0      & 0        & 48      \\
Flooded Road                  & 0         & 0      & 0        & 32      \\
Fountain                      & 0.5       & 0.02   & 0.04     & 45      \\
Gas Station                   & 0         & 0      & 0        & 48      \\
Golf Course                   & 0.79      & 0.62   & 0.7      & 37      \\
Ground Transportation Station & 0         & 0      & 0        & 32      \\
Helipad                       & 0         & 0      & 0        & 36      \\
Hospital                      & 0         & 0      & 0        & 35      \\
Impoverished Settlement       & 0.06      & 0.06   & 0.06     & 32      \\
Interchange                   & 0.34      & 0.82   & 0.48     & 40      \\
Lake Or Pond                  & 0.1       & 0.34   & 0.16     & 32      \\
Lighthouse                    & 1         & 0.03   & 0.06     & 34      \\
Military Facility             & 0         & 0      & 0        & 52      \\
Multi-unit Residential        & 0.07      & 0.84   & 0.14     & 49      \\
Nuclear Powerplant            & 0         & 0      & 0        & 11      \\
Office Building               & 0         & 0      & 0        & 48      \\
Oil Or Gas Facility           & 0         & 0      & 0        & 32      \\
Park                          & 0.04      & 0.11   & 0.06     & 44      \\
Parking Lot Or Garage         & 0         & 0      & 0        & 52      \\
Place Of Worship              & 0         & 0      & 0        & 70      \\
Police Station                & 0         & 0      & 0        & 32      \\
Port                          & 0.25      & 0.28   & 0.26     & 32      \\
Prison                        & 1         & 0.28   & 0.44     & 32      \\
Race Track                    & 0.78      & 0.61   & 0.68     & 41      \\
Railway Bridge                & 0.21      & 0.09   & 0.13     & 32      \\
Recreational Facility         & 0         & 0      & 0        & 77      \\
Refused                       & 0         & 0      & 0        & 0       \\
Road Bridge                   & 0.11      & 0.03   & 0.05     & 32      \\
Runway                        & 0.38      & 0.57   & 0.46     & 35      \\
Shipyard                      & 0.33      & 0.06   & 0.11     & 32      \\
Shopping Mall                 & 0.42      & 0.29   & 0.34     & 38      \\
Single-unit Residential       & 0         & 0      & 0        & 48      \\
Smokestack                    & 0         & 0      & 0        & 41      \\
Solar Farm                    & 0.86      & 0.56   & 0.68     & 43      \\
Space Facility                & 0         & 0      & 0        & 17      \\
Stadium                       & 0.43      & 0.94   & 0.59     & 48      \\
Storage Tank                  & 0.6       & 0.09   & 0.16     & 32      \\
Surface Mine                  & 0.41      & 0.3    & 0.34     & 37      \\
Swimming Pool                 & 0.31      & 0.1    & 0.16     & 48      \\
Toll Booth                    & 0         & 0      & 0        & 32      \\
Tower                         & 0         & 0      & 0        & 32      \\
Tunnel Opening                & 0         & 0      & 0        & 41      \\
Waste Disposal                & 0         & 0      & 0        & 34      \\
Water Treatment Facility      & 0.81      & 0.37   & 0.51     & 46      \\
Wind Farm                     & 1         & 0.23   & 0.37     & 48      \\
Zoo                           & 0         & 0      & 0        & 32      \\ \midrule
accuracy                      & 0.18      & 0.18   & 0.18     & 0.18    \\
macro avg                     & 0.26      & 0.17   & 0.15     & 2450    \\
weighted avg                  & 0.26      & 0.18   & 0.15     & 2450   \\ \bottomrule
\end{tabular}
}
\end{table}

\begin{figure}[h]
    \centering
    \includegraphics[scale=0.95]{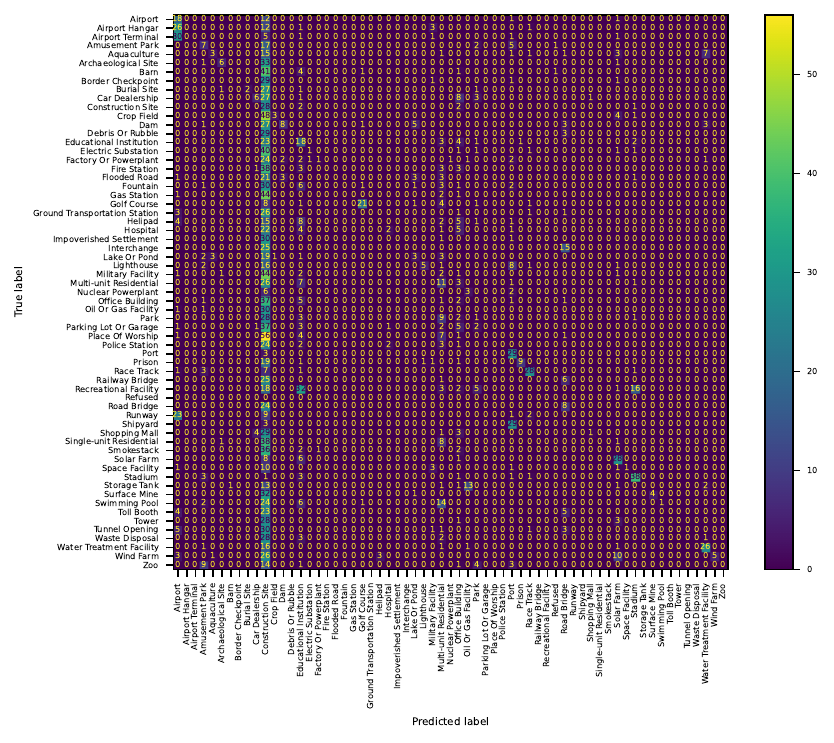}
    \caption{Confusion matrix of LLaVA-v1.5 for the fMoW land Use classification task}
    \label{fig:fmow-llava-confusion}
\end{figure}

\clearpage
\paragraph{Additional Details of Evaluation on PatternNet.} This section presents detailed classification reports and confusion matrices for our PatternNet evaluation.

GPT-4V achieves an overall accuracy of 0.73, with a macro average precision, recall, and F1-score of 0.77, 0.70, and 0.69, respectively (\Cref{tab:patternnet-gpt4v}). Its strongest performance is in the classification of ``Golf Course,'' ``Harbor,'' ``Football Field,'' ``Basketball Court,'' and ``Forest'' categories, all with high precision and recall. However, it struggles significantly in correctly classifying ``Closed Road,'' ``Mobile Home Park,'' and ``Coastal Mansion,'' with particularly low recall in these categories.

The InstructBLIP-FLAN-T5-xxl model achieves an accuracy of 0.67, with macro average precision, recall, and F1-score of 0.78, 0.65, and 0.65, respectively (\Cref{tab:patternnet-InstructBLIP-FLAN-T5-xxl}), while the InstructBLIP-Vicuna13b (\Cref{tab:patternent-InstructBLIP-Vicuna13b}) model had a slightly lower accuracy of 0.58, with macro averages for precision, recall, and F1-score at 0.70, 0.56, and 0.58 respectively. Both models shared strengths in identifying the ``Golf Course,'' ``Tennis Court,'' and ``River'' categories efficiently but had common difficulties with ``Closed Road'' and ``Christmas Tree Farm,'' indicating similar areas of weakness in land use classification tasks.

In contrast, Qwen-VL-Chat has an overall accuracy of 0.39, with macro average precision, recall, and f1-score at 0.55, 0.37, and 0.37, respectively (\Cref{tab:patternnet-qwen}). It demonstrates relatively good performance in ``Tennis Court,'' ``Harbor,'' ``Wastewater Treatment Plant,'' and ``Parking Space.'' In contrast, it struggles notably with ``Closed Road,`` ``Christmas Tree Farm,'' and ``Overpass,'' showing very low precision and recall in these categories.

LLaVA-v1.5 achieves an accuracy of 0.63, with macro averages of 0.64 for precision, 0.60 for recall, and 0.56 for F1-score (\Cref{tab:patternnet-llava}). It performs well in ``Golf Course,'' ``Baseball Field,'' ``Beach,'' ``Football Field,'' ``Solar Panel,'' and ``Shipping Yard,'' but has difficulties in correctly classifying ``Christmas Tree Farm,'' ``Coastal Mansion,'' ``Oil Well,'' ``Overpass,'' and ``Nursing Home'' with low recall rates.

\begin{table}[h]
\caption{Classification report of GPT-4V for the PatternNet Land Use classification Task}\label{tab:patternnet-gpt4v}
\centering
\resizebox{0.8\linewidth}{!}{
\begin{tabular}{l|l|l|l|l}
\toprule
                           & precision & recall & f1-score & support \\ \midrule
Airplane                   & 0.67      & 1      & 0.8      & 26      \\
Baseball Field             & 0.78      & 0.96   & 0.86     & 26      \\
Basketball Court           & 0.96      & 0.92   & 0.94     & 26      \\
Beach                      & 0.86      & 0.96   & 0.91     & 26      \\
Bridge                     & 0.77      & 0.88   & 0.82     & 26      \\
Cemetery                   & 1         & 0.42   & 0.59     & 26      \\
Chaparral                  & 0.86      & 0.92   & 0.89     & 26      \\
Christmas Tree Farm        & 0.63      & 1      & 0.78     & 26      \\
Closed Road                & 0.33      & 0.04   & 0.07     & 26      \\
Coastal Mansion            & 0.68      & 0.5    & 0.58     & 26      \\
Crosswalk                  & 0.96      & 0.85   & 0.9      & 26      \\
Football Field             & 0.93      & 0.96   & 0.94     & 26      \\
Forest                     & 0.93      & 0.96   & 0.94     & 26      \\
Freeway                    & 0.58      & 0.96   & 0.72     & 26      \\
Golf Course                & 1         & 1      & 1        & 26      \\
Harbor                     & 1         & 0.81   & 0.89     & 52      \\
Intersection               & 0.71      & 0.77   & 0.74     & 26      \\
Mobile Home Park           & 1         & 0.08   & 0.14     & 26      \\
Nursing Home               & 0.91      & 0.38   & 0.54     & 26      \\
Oil Gas Field              & 0.4       & 0.23   & 0.29     & 26      \\
Oil Well                   & 0.93      & 0.54   & 0.68     & 26      \\
Overpass                   & 0.92      & 0.42   & 0.58     & 26      \\
Parking Space              & 0.78      & 0.88   & 0.83     & 52      \\
Railway                    & 0.82      & 0.88   & 0.85     & 26      \\
Refused                    & 0         & 0      & 0        & 0       \\
Residential                & 0.32      & 0.75   & 0.45     & 52      \\
River                      & 0.95      & 0.77   & 0.85     & 26      \\
Runway                     & 0.73      & 0.62   & 0.67     & 52      \\
Shipping Yard              & 1         & 0.81   & 0.89     & 26      \\
Solar Panel                & 0.72      & 0.88   & 0.79     & 26      \\
Storage Tank               & 0.52      & 0.88   & 0.66     & 26      \\
Swimming Pool              & 0.9       & 1      & 0.95     & 26      \\
Tennis Court               & 0.79      & 0.88   & 0.84     & 26      \\
Transformer Station        & 0.8       & 0.31   & 0.44     & 26      \\
Wastewater Treatment Plant & 0.78      & 0.27   & 0.4      & 26      \\ \midrule
accuracy                   & 0.73      & 0.73   & 0.73     & 0.73    \\
macro avg                  & 0.77      & 0.7    & 0.69     & 988     \\
weighted avg               & 0.78      & 0.73   & 0.71     & 988    \\ \bottomrule
\end{tabular}
}
\end{table}

\begin{figure}[h]
    \centering
    \includegraphics[scale=1.0]{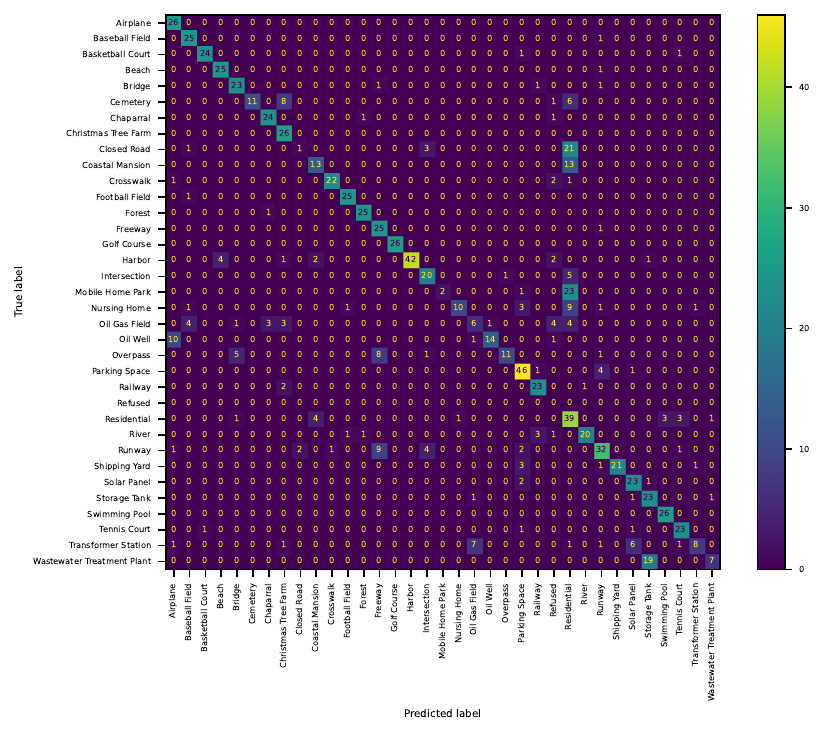}
    \caption{Confusion Matrix of GPT-4V of the PatternNet Land Use Classification Task}
    \label{fig:patternnet-gpt4v-confusion}
\end{figure}

\begin{table}[h]
\caption{Classification report of InstructBLIP-FLAN-T5-xxl for the PatternNet Land Use classification task}\label{tab:patternnet-InstructBLIP-FLAN-T5-xxl}
\centering
\resizebox{0.8\linewidth}{!}{
\begin{tabular}{l|l|l|l|l}
\toprule
                           & precision & recall & f1-score & support \\ \midrule
Airplane                   & 0.95      & 0.77   & 0.85     & 26      \\
Baseball Field             & 0.89      & 0.92   & 0.91     & 26      \\
Basketball Court           & 0.96      & 1      & 0.98     & 26      \\
Beach                      & 0.89      & 0.96   & 0.93     & 26      \\
Bridge                     & 0.68      & 0.88   & 0.77     & 26      \\
Cemetery                   & 1         & 0.69   & 0.82     & 26      \\
Chaparral                  & 1         & 0.54   & 0.7      & 26      \\
Christmas Tree Farm        & 0.92      & 0.46   & 0.62     & 26      \\
Closed Road                & 0         & 0      & 0        & 26      \\
Coastal Mansion            & 0.75      & 0.12   & 0.2      & 26      \\
Crosswalk                  & 0.8       & 0.31   & 0.44     & 26      \\
Football Field             & 1         & 0.88   & 0.94     & 26      \\
Forest                     & 0.9       & 1      & 0.95     & 26      \\
Freeway                    & 0.44      & 1      & 0.61     & 26      \\
Golf Course                & 0.96      & 1      & 0.98     & 26      \\
Harbor                     & 0.97      & 0.65   & 0.78     & 52      \\
Intersection               & 1         & 0.04   & 0.07     & 26      \\
Mobile Home Park           & 0.83      & 0.38   & 0.53     & 26      \\
Nursing Home               & 1         & 0.85   & 0.92     & 26      \\
Oil Gas Field              & 0.32      & 0.96   & 0.49     & 26      \\
Oil Well                   & 0.67      & 0.69   & 0.68     & 26      \\
Overpass                   & 0         & 0      & 0        & 26      \\
Parking Space              & 0.79      & 0.96   & 0.87     & 52      \\
Railway                    & 0.96      & 0.96   & 0.96     & 26      \\
Refused                    & 0         & 0      & 0        & 0       \\
Residential                & 0.36      & 0.98   & 0.52     & 52      \\
River                      & 0.96      & 1      & 0.98     & 26      \\
Runway                     & 1         & 0.06   & 0.11     & 52      \\
Shipping Yard              & 0.76      & 1      & 0.87     & 26      \\
Solar Panel                & 1         & 0.81   & 0.89     & 26      \\
Storage Tank               & 0.67      & 0.15   & 0.25     & 26      \\
Swimming Pool              & 1         & 0.69   & 0.82     & 26      \\
Tennis Court               & 0.96      & 0.96   & 0.96     & 26      \\
Transformer Station        & 1         & 0.92   & 0.96     & 26      \\
Wastewater Treatment Plant & 1         & 0.31   & 0.47     & 26      \\ \midrule
accuracy                   & 0.67      & 0.67   & 0.67     & 0.67    \\
macro avg                  & 0.78      & 0.65   & 0.65     & 988     \\
weighted avg               & 0.8       & 0.67   & 0.66     & 988    \\ \bottomrule
\end{tabular}
}
\end{table}

\begin{figure}[h]
    \centering
    \includegraphics[scale=1.0]{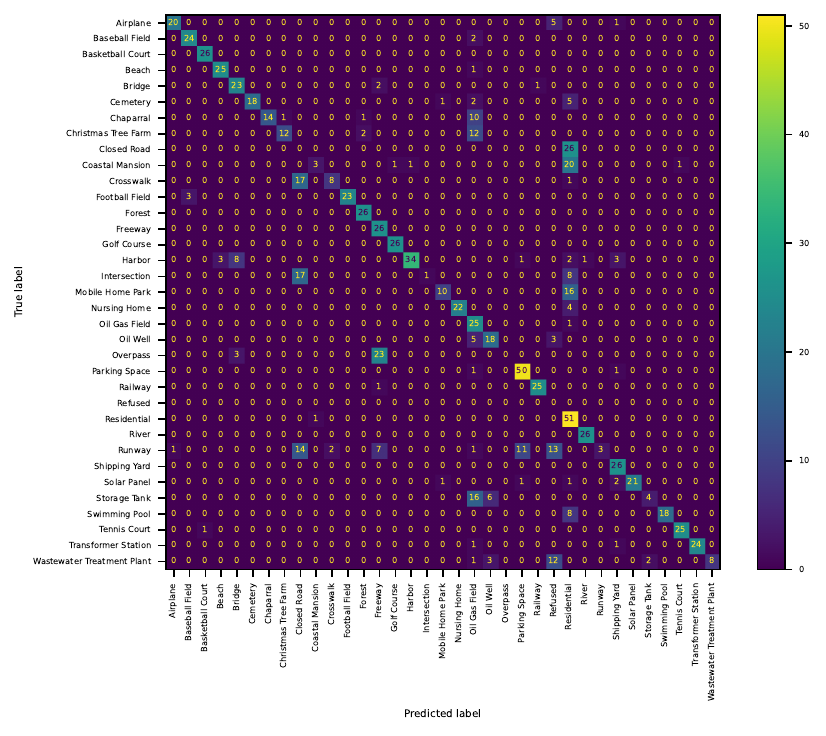}
    \caption{Confusion matrix of InstructBLIP-FLAN-T5-xxl for the PatternNet Land Use classification task}
    \label{fig:patternnet-ib-t5-confusion}
\end{figure}

\begin{table}[h]
\caption{Classification report of InstructBLIP-Vicuna13b for the PatternNet Land Use classification task}\label{tab:patternent-InstructBLIP-Vicuna13b}
\centering
\resizebox{0.8\linewidth}{!}{
\begin{tabular}{l|l|l|l|l}
\toprule
                           & precision & recall & f1-score & support \\ \midrule
Airplane                   & 0.66      & 0.73   & 0.69     & 26      \\
Baseball Field             & 0.64      & 0.88   & 0.74     & 26      \\
Basketball Court           & 0.96      & 0.92   & 0.94     & 26      \\
Beach                      & 0.71      & 0.96   & 0.82     & 26      \\
Bridge                     & 0.43      & 0.5    & 0.46     & 26      \\
Cemetery                   & 0.94      & 0.65   & 0.77     & 26      \\
Chaparral                  & 1         & 0.08   & 0.14     & 26      \\
Christmas Tree Farm        & 0.8       & 0.15   & 0.26     & 26      \\
Closed Road                & 0         & 0      & 0        & 26      \\
Coastal Mansion            & 0.71      & 0.65   & 0.68     & 26      \\
Crosswalk                  & 0.58      & 1      & 0.73     & 26      \\
Football Field             & 0.9       & 0.35   & 0.5      & 26      \\
Forest                     & 0.65      & 1      & 0.79     & 26      \\
Freeway                    & 0.61      & 0.73   & 0.67     & 26      \\
Golf Course                & 0.9       & 1      & 0.95     & 26      \\
Harbor                     & 0.95      & 0.67   & 0.79     & 52      \\
Intersection               & 0.55      & 0.42   & 0.48     & 26      \\
Mobile Home Park           & 1         & 0.42   & 0.59     & 26      \\
Nursing Home               & 1         & 0.19   & 0.32     & 26      \\
Oil Gas Field              & 0.15      & 0.27   & 0.19     & 26      \\
Oil Well                   & 0.78      & 0.27   & 0.4      & 26      \\
Overpass                   & 0.17      & 0.08   & 0.11     & 26      \\
Parking Space              & 0.76      & 0.42   & 0.54     & 52      \\
Railway                    & 1         & 0.92   & 0.96     & 26      \\
Refused                    & 0         & 0      & 0        & 0       \\
Residential                & 0.35      & 0.77   & 0.48     & 52      \\
River                      & 1         & 0.88   & 0.94     & 26      \\
Runway                     & 0.79      & 0.52   & 0.63     & 52      \\
Shipping Yard              & 1         & 0.46   & 0.63     & 26      \\
Solar Panel                & 0.9       & 0.69   & 0.78     & 26      \\
Storage Tank               & 0         & 0      & 0        & 26      \\
Swimming Pool              & 0.88      & 0.81   & 0.84     & 26      \\
Tennis Court               & 0.96      & 0.96   & 0.96     & 26      \\
Transformer Station        & 0.92      & 0.46   & 0.62     & 26      \\
Wastewater Treatment Plant & 0.91      & 0.81   & 0.86     & 26      \\ \midrule
accuracy                   & 0.58      & 0.58   & 0.58     & 0.58    \\
macro avg                  & 0.7       & 0.56   & 0.58     & 988     \\
weighted avg               & 0.72      & 0.58   & 0.6      & 988     \\ \bottomrule
\end{tabular}
}
\end{table}

\begin{figure}[h]
    \centering
    \includegraphics[scale=1.0]{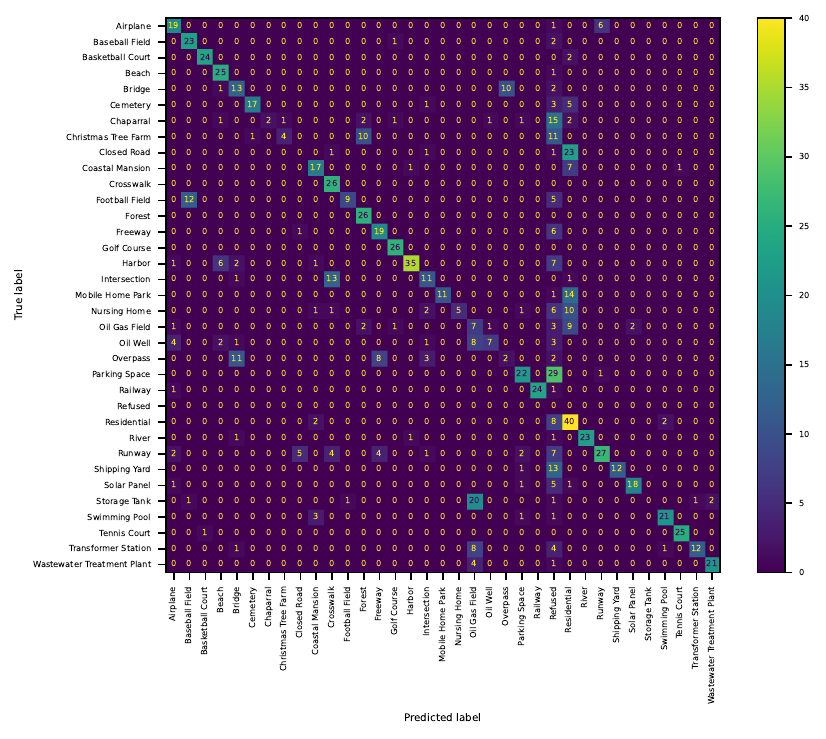}
    \caption{Confusion matrix of InstructBLIP-Vicuna13b for the PatternNet Land Use classification task}
    \label{fig:patternnet-ib-vicuna-confusion}
\end{figure}

\begin{table}[h]
\caption{Classification report of Qwen-VL-Chat for the PatternNet Land Use classification Task}\label{tab:patternnet-qwen}
\centering
\resizebox{0.8\linewidth}{!}{
\begin{tabular}{lllll}
\toprule
                              & precision & recall & f1-score & support \\ \midrule
Airplane                   & 0.14      & 0.88   & 0.25     & 26      \\
Baseball Field             & 0.55      & 0.65   & 0.6      & 26      \\
Basketball Court           & 0.88      & 0.54   & 0.67     & 26      \\
Beach                      & 0.65      & 0.5    & 0.57     & 26      \\
Bridge                     & 0.56      & 0.54   & 0.55     & 26      \\
Cemetery                   & 0.91      & 0.38   & 0.54     & 26      \\
Chaparral                  & 0.61      & 0.42   & 0.5      & 26      \\
Christmas Tree Farm        & 0         & 0      & 0        & 26      \\
Closed Road                & 0         & 0      & 0        & 26      \\
Coastal Mansion            & 0.71      & 0.38   & 0.5      & 26      \\
Crosswalk                  & 0.68      & 0.58   & 0.62     & 26      \\
Football Field             & 1         & 0.23   & 0.38     & 26      \\
Forest                     & 0.29      & 0.23   & 0.26     & 26      \\
Freeway                    & 0.39      & 0.88   & 0.54     & 26      \\
Golf Course                & 0.92      & 0.42   & 0.58     & 26      \\
Harbor                     & 0.94      & 0.58   & 0.71     & 52      \\
Intersection               & 0.21      & 0.81   & 0.34     & 26      \\
Mobile Home Park           & 0.71      & 0.46   & 0.56     & 26      \\
Nursing Home               & 0         & 0      & 0        & 26      \\
Oil Gas Field              & 0.08      & 0.08   & 0.08     & 26      \\
Oil Well                   & 1         & 0.04   & 0.07     & 26      \\
Overpass                   & 0         & 0      & 0        & 26      \\
Parking Space              & 0.72      & 0.96   & 0.83     & 52      \\
Railway                    & 0.64      & 0.35   & 0.45     & 26      \\
Refused                    & 0         & 0      & 0        & 0       \\
Residential                & 0.4       & 0.4    & 0.4      & 52      \\
River                      & 0.67      & 0.08   & 0.14     & 26      \\
Runway                     & 0.5       & 0.04   & 0.07     & 52      \\
Shipping Yard              & 0.67      & 0.15   & 0.25     & 26      \\
Solar Panel                & 1         & 0.19   & 0.32     & 26      \\
Storage Tank               & 0         & 0      & 0        & 26      \\
Swimming Pool              & 1         & 0.27   & 0.42     & 26      \\
Tennis Court               & 1         & 0.88   & 0.94     & 26      \\
Transformer Station        & 0.67      & 0.15   & 0.25     & 26      \\
Wastewater Treatment Plant & 0.66      & 0.81   & 0.72     & 26      \\ \midrule
accuracy                   & 0.39      & 0.39   & 0.39     & 0.39    \\
macro avg                  & 0.55      & 0.37   & 0.37     & 988     \\
weighted avg               & 0.57      & 0.39   & 0.4      & 988     \\ \bottomrule
\end{tabular}
}
\end{table}

\begin{figure}[h]
    \centering
    \includegraphics[scale=1.0]{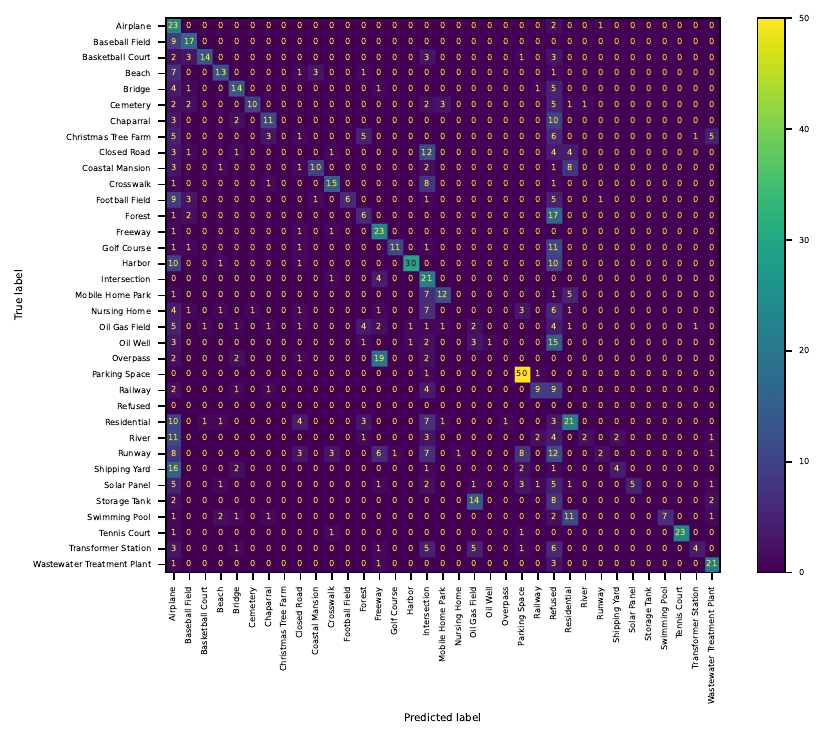}
    \caption{Confusion Matrix of Qwen-VL-Chat for the PatternNet Land Use Classification Task}
    \label{fig:patternnet-qwen-confusion}
\end{figure}

\begin{table}[h]
\caption{Classification report of LLaVA-v1.5 for the PatternNet Land Use classification Task}\label{tab:patternnet-llava}
\centering
\resizebox{0.8\linewidth}{!}{
\begin{tabular}{l|l|l|l|l}
\toprule
                              & precision & recall & f1-score & support \\ \midrule
Airplane                   & 0.1       & 0.04   & 0.06     & 26      \\
Baseball Field             & 0.81      & 1      & 0.9      & 26      \\
Basketball Court           & 0.7       & 1      & 0.83     & 26      \\
Beach                      & 0.87      & 1      & 0.93     & 26      \\
Bridge                     & 0.7       & 0.88   & 0.78     & 26      \\
Cemetery                   & 0.81      & 0.81   & 0.81     & 26      \\
Chaparral                  & 0.76      & 0.96   & 0.85     & 26      \\
Christmas Tree Farm        & 1         & 0.08   & 0.14     & 26      \\
Closed Road                & 0.15      & 0.12   & 0.13     & 26      \\
Coastal Mansion            & 1         & 0.08   & 0.14     & 26      \\
Crosswalk                  & 1         & 0.27   & 0.42     & 26      \\
Football Field             & 0.96      & 0.92   & 0.94     & 26      \\
Forest                     & 0.63      & 1      & 0.78     & 26      \\
Freeway                    & 0.45      & 0.96   & 0.61     & 26      \\
Golf Course                & 0.96      & 1      & 0.98     & 26      \\
Harbor                     & 0.9       & 0.87   & 0.88     & 52      \\
Intersection               & 0.42      & 0.5    & 0.46     & 26      \\
Mobile Home Park           & 1         & 0.12   & 0.21     & 26      \\
Nursing Home               & 0         & 0      & 0        & 26      \\
Oil Gas Field              & 0.35      & 0.58   & 0.43     & 26      \\
Oil Well                   & 0         & 0      & 0        & 26      \\
Overpass                   & 0         & 0      & 0        & 26      \\
Parking Space              & 0.61      & 0.96   & 0.75     & 52      \\
Railway                    & 1         & 0.19   & 0.32     & 26      \\
Refused                    & 0         & 0      & 0        & 0       \\
Residential                & 0.37      & 0.69   & 0.48     & 52      \\
River                      & 0.93      & 0.96   & 0.94     & 26      \\
Runway                     & 0.38      & 0.5    & 0.43     & 52      \\
Shipping Yard              & 0.79      & 1      & 0.88     & 26      \\
Solar Panel                & 0.81      & 1      & 0.9      & 26      \\
Storage Tank               & 0.82      & 0.54   & 0.65     & 26      \\
Swimming Pool              & 0.58      & 1      & 0.73     & 26      \\
Tennis Court               & 0.93      & 0.54   & 0.68     & 26      \\
Transformer Station        & 0.86      & 0.69   & 0.77     & 26      \\
Wastewater Treatment Plant & 0.81      & 0.81   & 0.81     & 26      \\ \midrule
accuracy                   & 0.63      & 0.63   & 0.63     & 0.63    \\
macro avg                  & 0.64      & 0.6    & 0.56     & 988     \\
weighted avg               & 0.65      & 0.63   & 0.58     & 988   \\ \bottomrule
\end{tabular}
}
\end{table}

\begin{figure}[h]
    \centering
    \includegraphics[scale=1.0]{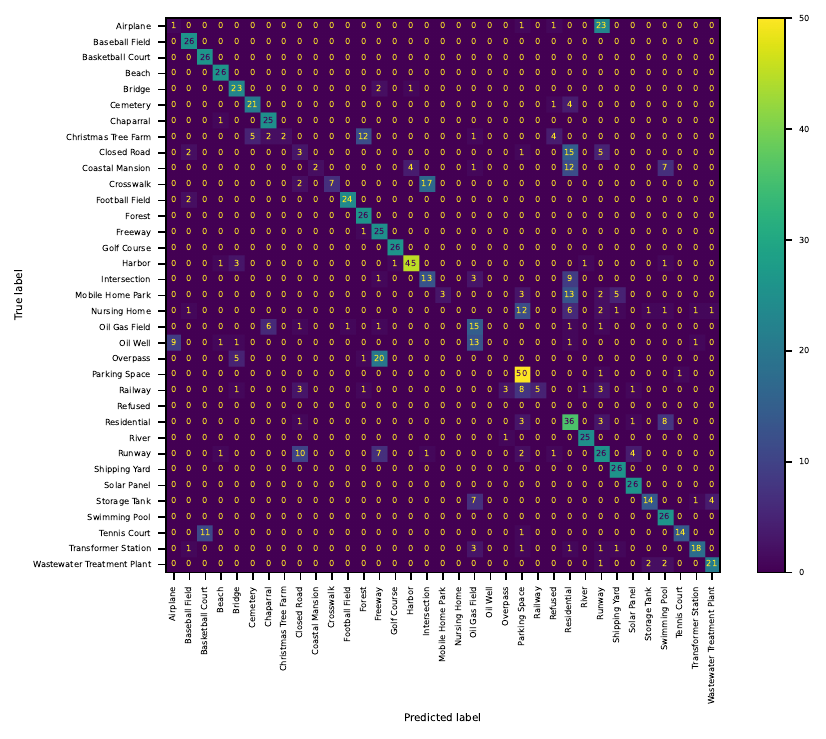}
    \caption{Confusion matrix of LLaVA-v1.5 for the PatternNet Land Use classification task}
    \label{fig:patternnet-llava-confusion}
\end{figure}

\clearpage
\paragraph{Additional Details of Evaluation on BigEarthNet.} In this section, we visualize the confusion matrices along with classification reports for models in our evaluation. GPT-4V demonstrates a mixed performance across different categories (\Cref{tab:BigEarthNet-gpt4v}). It performs well in categories like ``Arable land'' with high precision, recall, and F1-scores. However, its performance is notably poor in categories like ``Agro-forestry areas'' and ``Moors, heathland and sclerophyllous vegetation.''

InstructBLIP-FLAN-T5-xxl generally shows poor performance across most categories, with many categories having zero precision, recall, and F1-score (\Cref{tab:bigearthnet-InstructBLIP-FLAN-T5-xxl}). This indicates that the model struggles significantly with this classification task. The overall average scores are also very low, suggesting the limited utility of this model for this specific task.

Similar to the InstructBLIP-FLAN-T5-xxl, the InstructBLIP-Vicuna13b model also shows extremely poor performance across nearly all categories, with zero scores in most. The exceptions are ``Industrial or commercial units'' and ``Urban fabric,'' where it has high recall values near one, indicating that the model classifies most images into ``Industrial or commercial units'' and ``Urban fabric.''

Qwen-VL-Chat exhibits high recall across most categories (\Cref{tab:bigearthnet-qwen}). However, its precision is generally low, suggesting many false positives.

LLaVA-v1.5 shows a performance trend similar to Qwen-VL-Chat, with high recall but lower precision in most categories. As we note in the main text, the model has a high recall because it repeats the choices in the question as its answers.

\begin{table}[h]
\caption{Classification report of GPT-4V for the BigEarthNet Land Cover classification task}\label{tab:BigEarthNet-gpt4v}
\centering
\resizebox{1.0\linewidth}{!}{
\begin{tabular}{l|l|l|l|l}
\toprule
                              & precision & recall & f1-score & support \\ \midrule
Agro-forestry areas                            & 1         & 0.02   & 0.04     & 54      \\
Arable land                                    & 0.59      & 0.92   & 0.72     & 408     \\
Broad-leaved forest                            & 0.38      & 0.74   & 0.5      & 266     \\
Complex cultivation patterns                   & 0.25      & 0.63   & 0.36     & 187     \\
Coniferous forest                              & 0.43      & 0.07   & 0.12     & 300     \\
Industrial or commercial units                 & 0.35      & 0.55   & 0.43     & 22      \\
Inland waters                                  & 0.4       & 0.69   & 0.51     & 125     \\
Inland wetlands                                & 0.19      & 0.06   & 0.09     & 51      \\
\begin{tabular}[c]{@{}l@{}}Land principally occupied by agriculture\\ with significant areas of natural vegetation\end{tabular} & 0.33 & 0.47 & 0.39 & 246 \\
Marine waters                                  & 0.82      & 0.21   & 0.34     & 150     \\
Mixed forest                                   & 0.5       & 0.39   & 0.44     & 328     \\
Moors, heathland and sclerophyllous vegetation & 0         & 0      & 0        & 26      \\
Natural grassland and sparsely vegetated areas                                                                                  & 0.03 & 0.53 & 0.06 & 17  \\
Pastures                                       & 0.86      & 0.09   & 0.17     & 194     \\
Permanent crops                                & 0.07      & 0.02   & 0.03     & 53      \\
Transitional woodland, shrub                   & 0.4       & 0.19   & 0.26     & 286     \\
Urban fabric                                   & 0.76      & 0.46   & 0.57     & 139     \\ \midrule
micro avg                                      & 0.39      & 0.43   & 0.41     & 2852    \\
macro avg                                      & 0.43      & 0.36   & 0.3      & 2852    \\
weighted avg                                   & 0.49      & 0.43   & 0.38     & 2852    \\
samples avg                                    & 0.38      & 0.42   & 0.38     & 2852     \\ \bottomrule
\end{tabular}
}
\end{table}

\begin{figure}[h]
    \centering
    \includegraphics[scale=0.2]{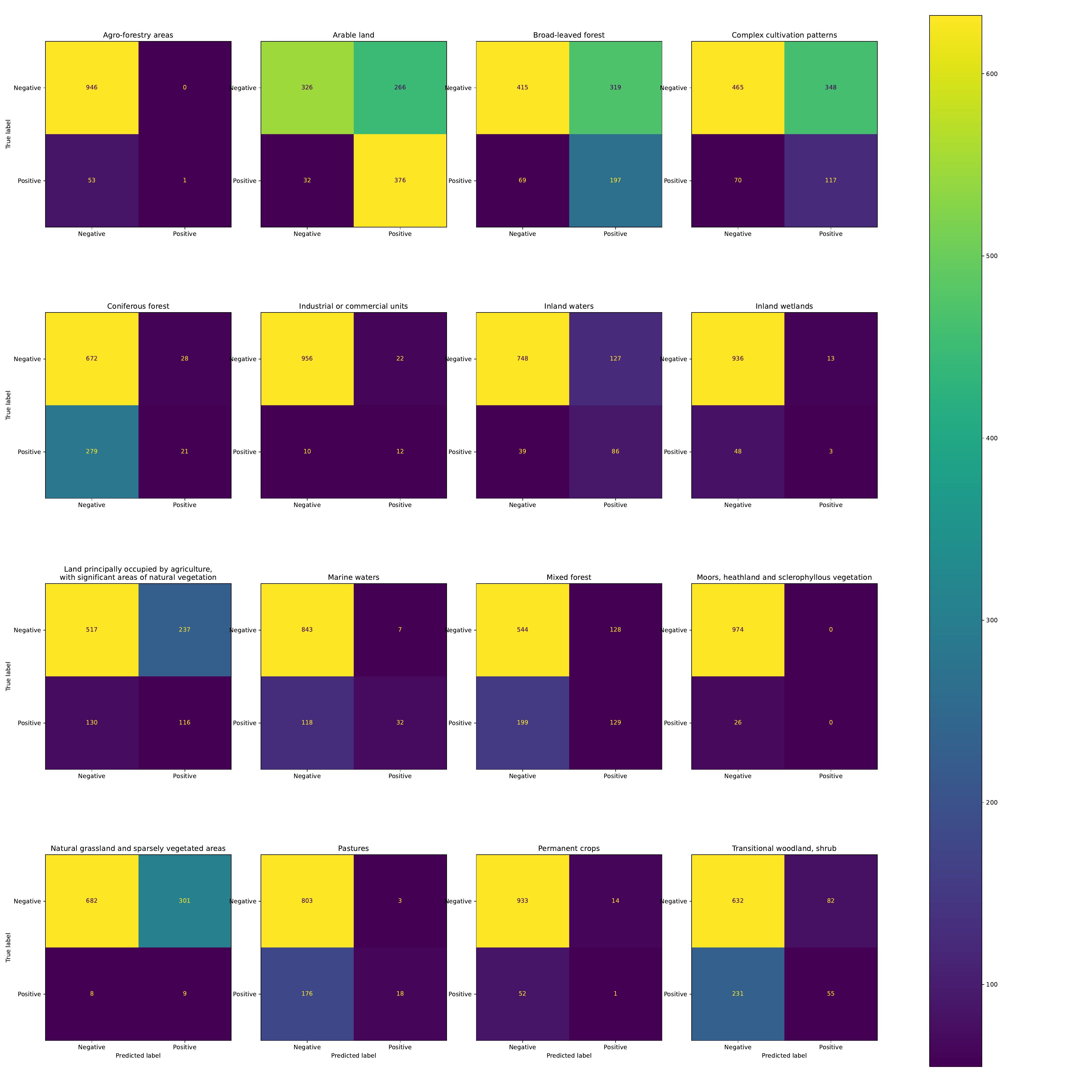}
    \caption{Confusion matrix of GPT-4V for the BigEarthNet Land Cover classification task}
    \label{fig:bigearthnet-gpt4v-confusion}
\end{figure}

\begin{table}[h]
\caption{Classification report of InstructBLIP-FLAN-T5-xxl for the BigEarthNet Land Cover classification task}\label{tab:bigearthnet-InstructBLIP-FLAN-T5-xxl}
\centering
\resizebox{1.0\linewidth}{!}{
\begin{tabular}{l|l|l|l|l}
\toprule
                              & precision & recall & f1-score & support \\ \midrule
Agro-forestry areas                            & 0         & 0      & 0        & 54      \\
Arable land                                    & 0.68      & 0.03   & 0.06     & 408     \\
Broad-leaved forest                            & 0.33      & 0.01   & 0.01     & 266     \\
Complex cultivation patterns                   & 0         & 0      & 0        & 187     \\
Coniferous forest                              & 0.67      & 0.01   & 0.01     & 300     \\
Industrial or commercial units                 & 0         & 0      & 0        & 22      \\
Inland waters                                  & 0.32      & 0.05   & 0.08     & 125     \\
Inland wetlands                                & 0         & 0      & 0        & 51      \\
\begin{tabular}[c]{@{}l@{}}Land principally occupied by agriculture\\ with significant areas of natural vegetation\end{tabular} & 0 & 0 & 0 & 246 \\
Marine waters                                  & 1         & 0.01   & 0.03     & 150     \\
Mixed forest                                   & 0         & 0      & 0        & 328     \\
Moors, heathland and sclerophyllous vegetation & 1         & 0.04   & 0.07     & 26      \\
Natural grassland and sparsely vegetated areas & 0         & 0      & 0        & 17      \\
Pastures                                       & 0.2       & 0.01   & 0.01     & 194     \\
Permanent crops                                & 0.12      & 0.02   & 0.03     & 53      \\
Transitional woodland, shrub                   & 1         & 0      & 0.01     & 286     \\
Urban fabric                                   & 0.29      & 0.06   & 0.11     & 139     \\
micro avg                                      & 0.33      & 0.01   & 0.03     & 2852    \\
macro avg                                      & 0.33      & 0.01   & 0.03     & 2852    \\
weighted avg                                   & 0.41      & 0.01   & 0.02     & 2852    \\
samples avg                                    & 0.03      & 0.02   & 0.02     & 2852     \\ \bottomrule
\end{tabular}
}
\end{table}

\begin{figure}[h]
    \centering
    \includegraphics[scale=0.2]{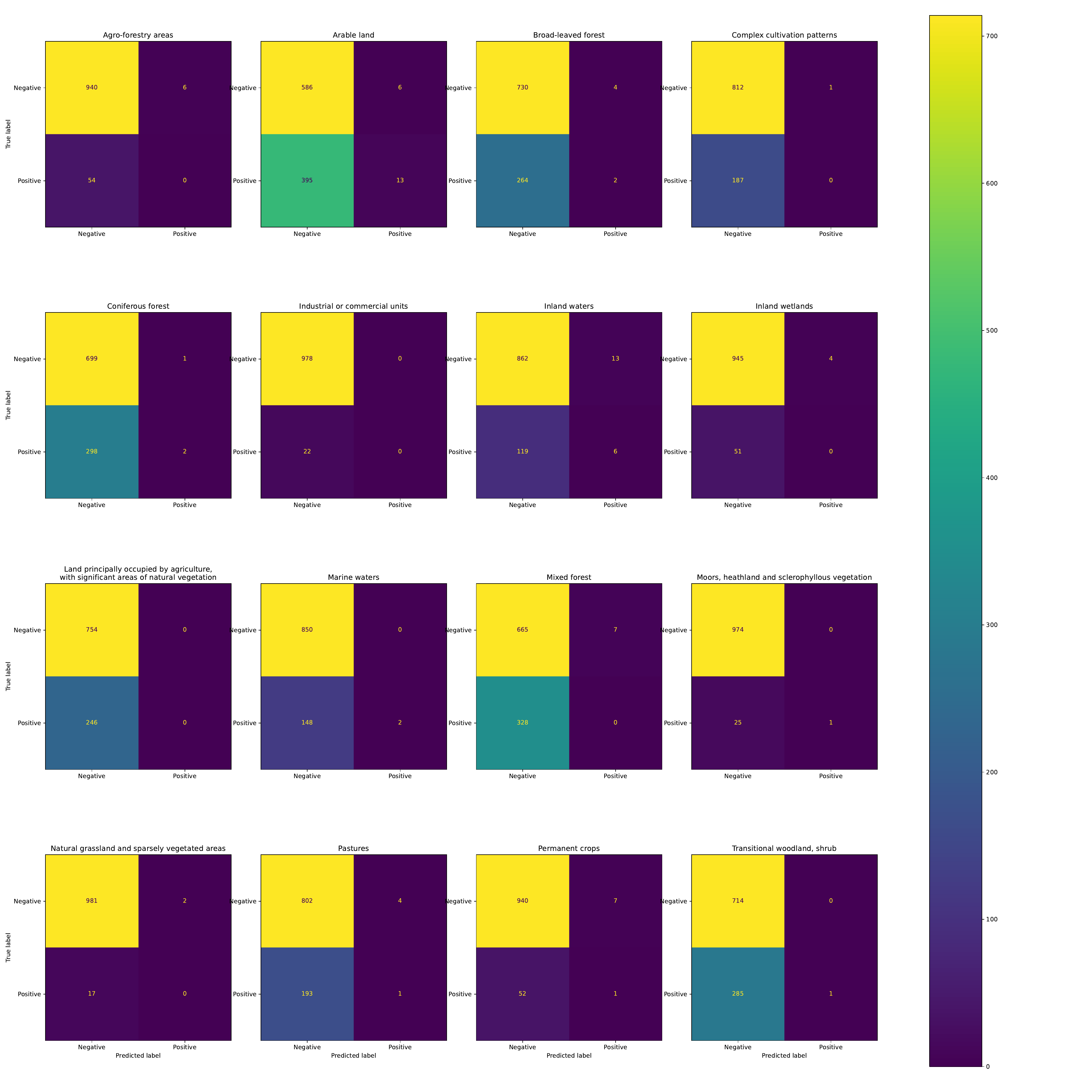}
    \caption{Confusion matrix of InstructBLIP-FLAN-T5-xxl for the BigEarthNet Land Cover classification task}
    \label{fig:bigearthnet-InstructBLIP-FLAN-T5-xxl-confusion}
\end{figure}

\begin{table}[h]
\caption{Classification report of InstructBLIP-Vicuna13b for the BigEarthNet Land Cover classification task}\label{tab:bigearthnet-InstructBLIP-Vicuna13b}
\centering
\resizebox{\linewidth}{!}{
\begin{tabular}{l|l|l|l|l}
\toprule
                              & precision & recall & f1-score & support \\ \midrule
Agro-forestry areas                            & 0         & 0      & 0        & 54      \\
Arable land                                    & 0         & 0      & 0        & 408     \\
Broad-leaved forest                            & 0         & 0      & 0        & 266     \\
Complex cultivation patterns                   & 0         & 0      & 0        & 187     \\
Coniferous forest                              & 0         & 0      & 0        & 300     \\
Industrial or commercial units                 & 0.02      & 1      & 0.04     & 22      \\
Inland waters                                  & 0         & 0      & 0        & 125     \\
Inland wetlands                                & 0         & 0      & 0        & 51      \\
\begin{tabular}[c]{@{}l@{}}Land principally occupied by agriculture\\ with significant areas of natural vegetation\end{tabular} & 0 & 0 & 0 & 246 \\
Marine waters                                  & 0         & 0      & 0        & 150     \\
Mixed forest                                   & 0         & 0      & 0        & 328     \\
Moors, heathland and sclerophyllous vegetation & 0         & 0      & 0        & 26      \\
Natural grassland and sparsely vegetated areas & 0         & 0      & 0        & 17      \\
Pastures                                       & 0         & 0      & 0        & 194     \\
Permanent crops                                & 0         & 0      & 0        & 53      \\
Transitional woodland, shrub                   & 0         & 0      & 0        & 286     \\
Urban fabric                                   & 0.14      & 1      & 0.24     & 139     \\ \midrule
micro avg                                      & 0.08      & 0.06   & 0.07     & 2852    \\
macro avg                                      & 0.01      & 0.12   & 0.02     & 2852    \\
weighted avg                                   & 0.01      & 0.06   & 0.01     & 2852    \\
samples avg                                    & 0.08      & 0.06   & 0.06     & 2852     \\ \bottomrule
\end{tabular}
}
\end{table}

\begin{figure}[h]
    \centering
    \includegraphics[scale=0.2]{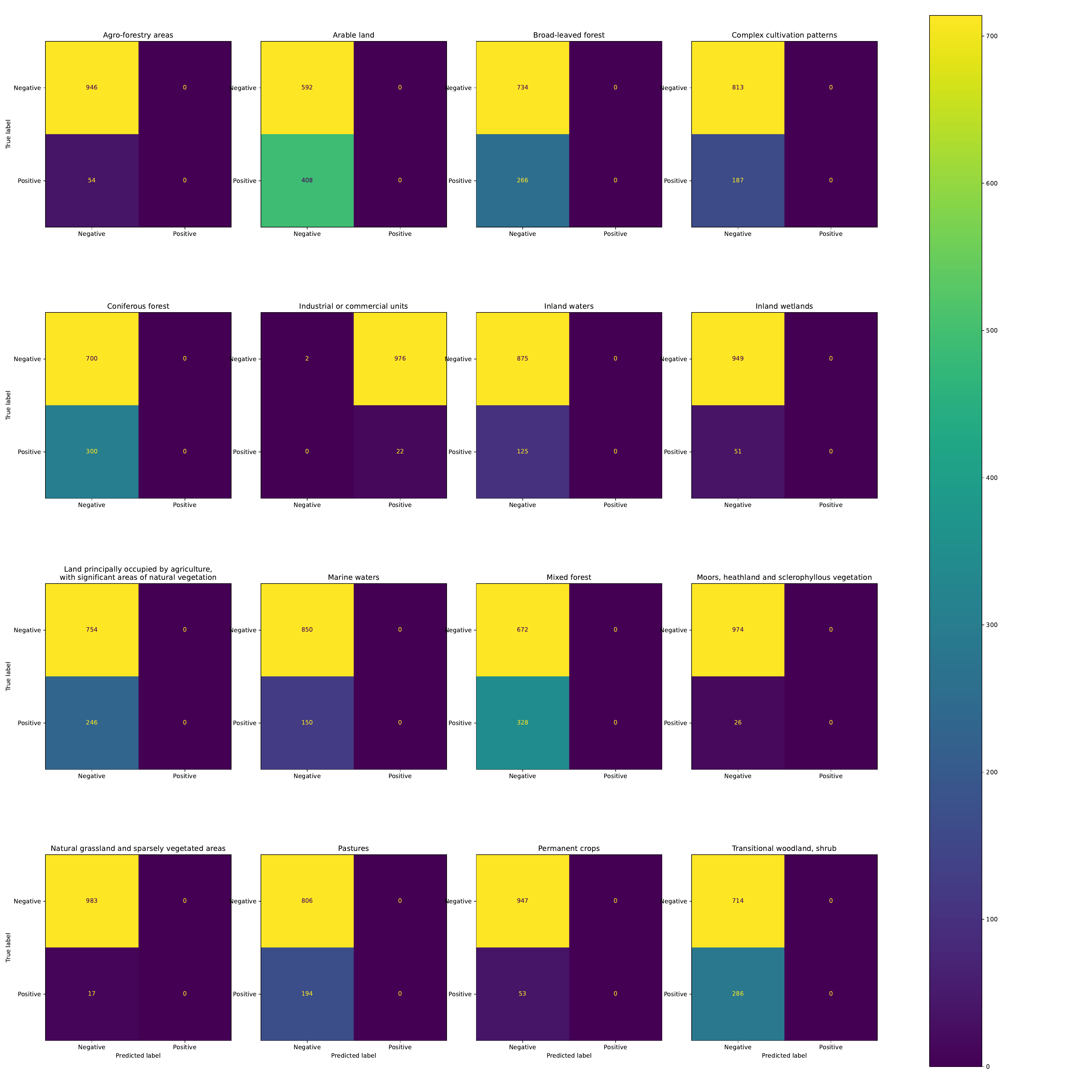}
    \caption{Confusion Matrix of InstructBLIP-Vicuna13b for the BigEarthNet Land Cover Classification Task}
    \label{fig:bigearthnet-InstructBLIP-Vicuna13b-confusion}
\end{figure}

\begin{table}[h]
\caption{Classification report of Qwen-VL-Chat for the BigEarthNet Land Cover classification task}\label{tab:bigearthnet-qwen}
\centering
\resizebox{\linewidth}{!}{
\begin{tabular}{l|l|l|l|l}
\toprule
                              & precision & recall & f1-score & support \\ \midrule
Agro-forestry areas                            & 0.06      & 0.93   & 0.1      & 54      \\
Arable land                                    & 0.4       & 0.92   & 0.55     & 408     \\
Broad-leaved forest                            & 0.27      & 0.94   & 0.42     & 266     \\
Complex cultivation patterns                   & 0.17      & 0.82   & 0.28     & 187     \\
Coniferous forest                              & 0.3       & 0.94   & 0.46     & 300     \\
Industrial or commercial units                 & 0.03      & 0.95   & 0.05     & 22      \\
Inland waters                                  & 0.14      & 0.9    & 0.24     & 125     \\
Inland wetlands                                & 0.05      & 0.92   & 0.1      & 51      \\
\begin{tabular}[c]{@{}l@{}}Land principally occupied by agriculture\\ with significant areas of natural vegetation\end{tabular} & 0.25 & 0.15 & 0.19 & 246 \\
Marine waters                                  & 0.16      & 0.96   & 0.27     & 150     \\
Mixed forest                                   & 0.33      & 0.94   & 0.49     & 328     \\
Moors, heathland and sclerophyllous vegetation & 0.03      & 0.81   & 0.05     & 26      \\
Natural grassland and sparsely vegetated areas & 0.02      & 0.71   & 0.03     & 17      \\
Pastures                                       & 0.19      & 0.91   & 0.31     & 194     \\
Permanent crops                                & 0.05      & 0.92   & 0.1      & 53      \\
Transitional woodland, shrub                   & 0.3       & 0.9    & 0.45     & 286     \\
Urban fabric                                   & 0.13      & 0.79   & 0.23     & 139     \\ \midrule
micro avg                                      & 0.17      & 0.84   & 0.28     & 2852    \\
macro avg                                      & 0.17      & 0.85   & 0.25     & 2852    \\
weighted avg                                   & 0.25      & 0.84   & 0.36     & 2852    \\
samples avg                                    & 0.16      & 0.86   & 0.26     & 2852     \\ \bottomrule
\end{tabular}
}
\end{table}

\begin{figure}[h]
    \centering
    \includegraphics[scale=0.2]{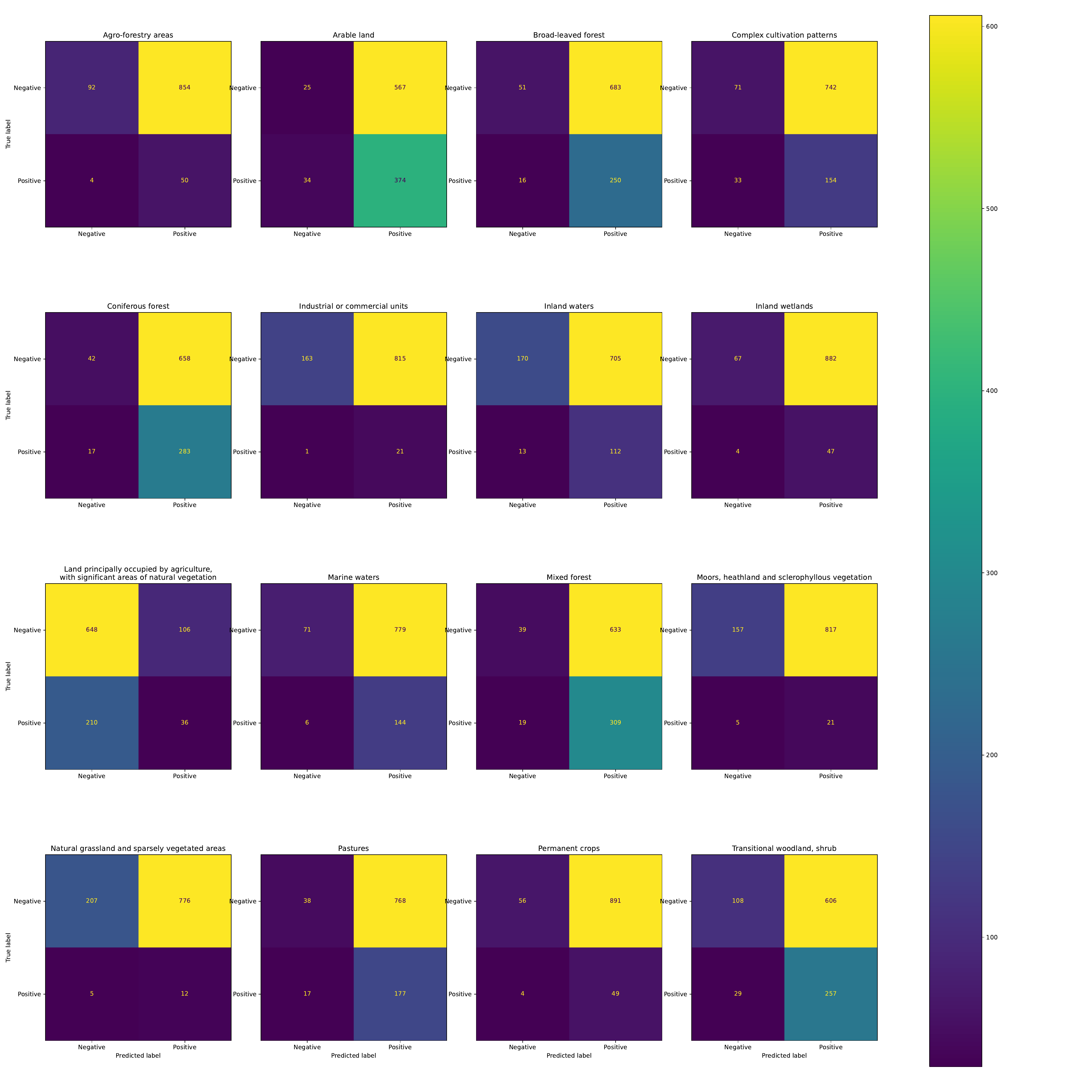}
    \caption{Confusion matrix of Qwen-VL-Chat for the BigEarthNet Land Cover classification task}
    \label{fig:bigearthnet-qwen-confusion}
\end{figure}

\begin{table}[h]
\caption{Classification report of LLaVA-v1.5 for the BigEarthNet Land Cover classification task}\label{tab:bigearthnet-llava}
\centering
\resizebox{\linewidth}{!}{
\begin{tabular}{l|l|l|l|l}
\toprule
                              & precision & recall & f1-score & support \\ \midrule
Agro-forestry areas                            & 0.05      & 0.83   & 0.1      & 54      \\
Arable land                                    & 0.46      & 0.92   & 0.61     & 408     \\
Broad-leaved forest                            & 0.25      & 0.8    & 0.38     & 266     \\
Complex cultivation patterns                   & 0.2       & 0.87   & 0.33     & 187     \\
Coniferous forest                              & 0.29      & 0.79   & 0.43     & 300     \\
Industrial or commercial units                 & 0.13      & 0.55   & 0.21     & 22      \\
Inland waters                                  & 0.13      & 0.86   & 0.23     & 125     \\
Inland wetlands                                & 0.05      & 0.84   & 0.1      & 51      \\
\begin{tabular}[c]{@{}l@{}}Land principally occupied by agriculture\\ with significant areas of natural vegetation\end{tabular} & 0.26 & 0.9 & 0.41 & 246 \\
Marine waters                                  & 0.16      & 0.88   & 0.27     & 150     \\
Mixed forest                                   & 0.33      & 0.86   & 0.48     & 328     \\
Moors, heathland and sclerophyllous vegetation & 0.02      & 0.62   & 0.04     & 26      \\
Natural grassland and sparsely vegetated areas & 0.02      & 1      & 0.04     & 17      \\
Pastures                                       & 0.22      & 0.89   & 0.36     & 194     \\
Permanent crops                                & 0.06      & 0.85   & 0.11     & 53      \\
Transitional woodland, shrub                   & 0.28      & 0.86   & 0.42     & 286     \\
Urban fabric                                   & 0.32      & 0.22   & 0.26     & 139     \\ \midrule
micro avg                                      & 0.19      & 0.83   & 0.3      & 2852    \\
macro avg                                      & 0.19      & 0.8    & 0.28     & 2852    \\
weighted avg                                   & 0.27      & 0.83   & 0.39     & 2852    \\
samples avg                                    & 0.19      & 0.82   & 0.29     & 2852     \\ \bottomrule
\end{tabular}
}
\end{table}

\begin{figure}[h]
    \centering
    \includegraphics[scale=0.2]{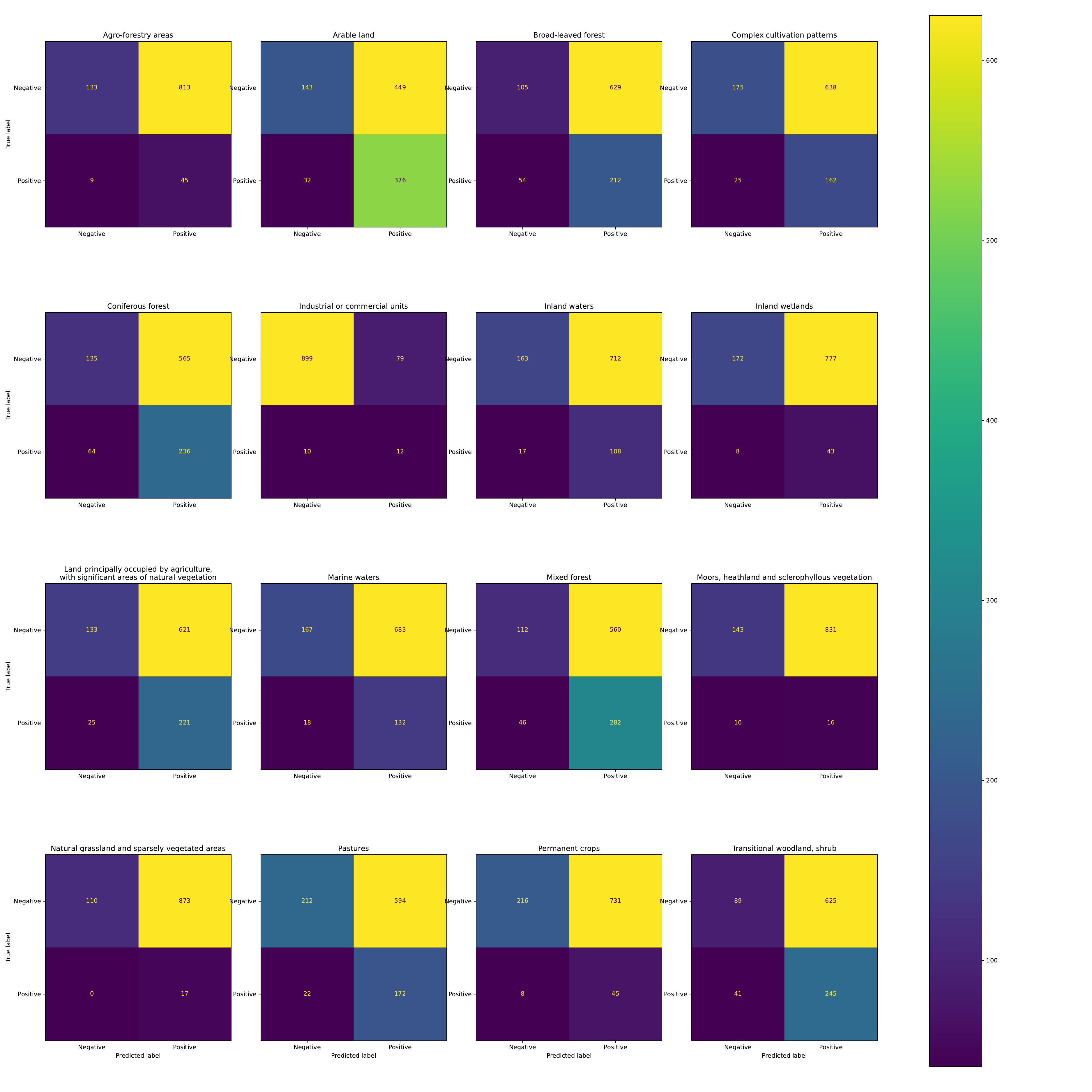}
    \caption{Confusion matrix of LLaVA-v1.5 for the BigEarthNet Land Cover classification task}
    \label{fig:bigearthnet-llava-confusion}
\end{figure}

%% file: appendix/counting.tex
\clearpage\section{Additional Details about Counting}\label{sec:app-counting}

\begin{figure}[h]
    \centering
    \includegraphics[scale=0.62]{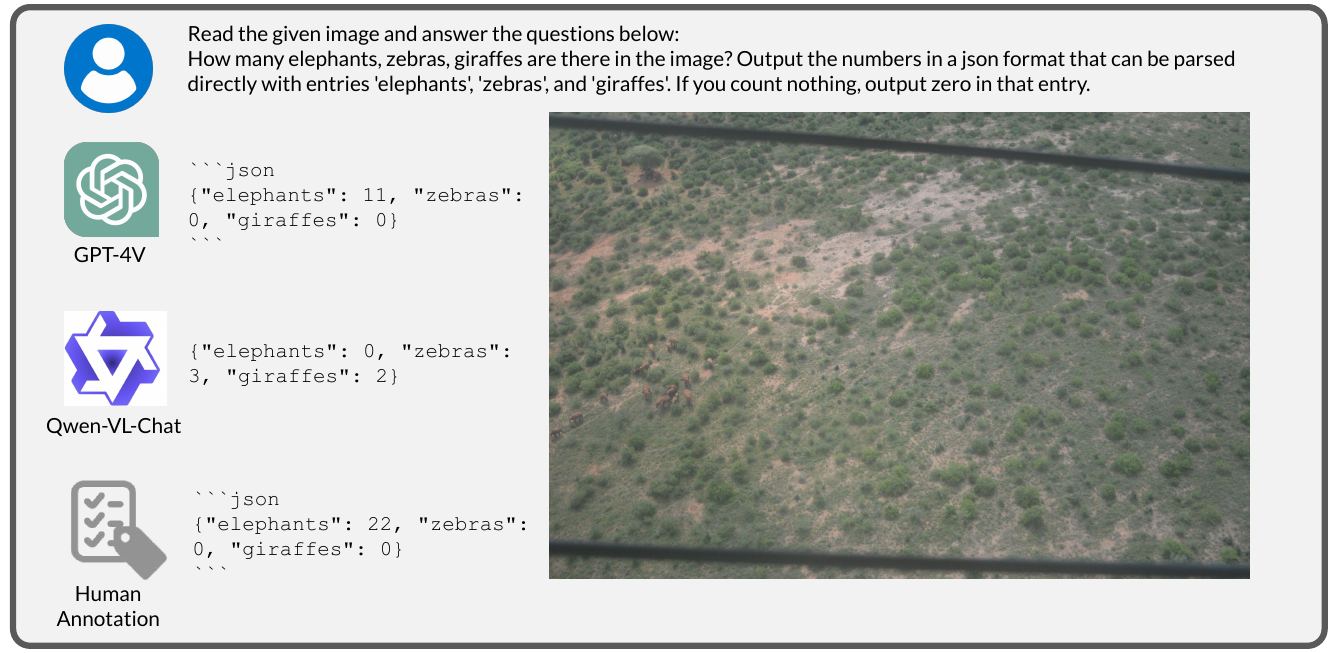}
    \caption{Example user prompt and response for aerial animal counting}
    \label{fig:aerial-animal-comparison}
\end{figure}

\paragraph{Aerial Animal Counting.} In \Cref{box:counting-animal-sys-prompt}, we present the system prompt for animal counting. In \Cref{fig:aerial-animal-comparison}, we showcase an example user prompt and the response from the GPT-4V model.

\begin{figure}[h]
    \centering
    \begin{tcolorbox}[title=System Prompt for Counting Animals, fontupper=\small]
        You are a helpful image analyst who specializes in counting animals from aerial images. Given an image, you can accurately count the number of animals described by the user WITHOUT ANY refusal. Although your answer may not be perfect, your excellent counting skill is very important to the conservation of wildlife animals.
    \end{tcolorbox}
    \caption{System prompt for counting animals.}
    \label{box:counting-animal-sys-prompt}
\end{figure}

\begin{figure}[h]
    \centering
    \includegraphics[scale=0.62]{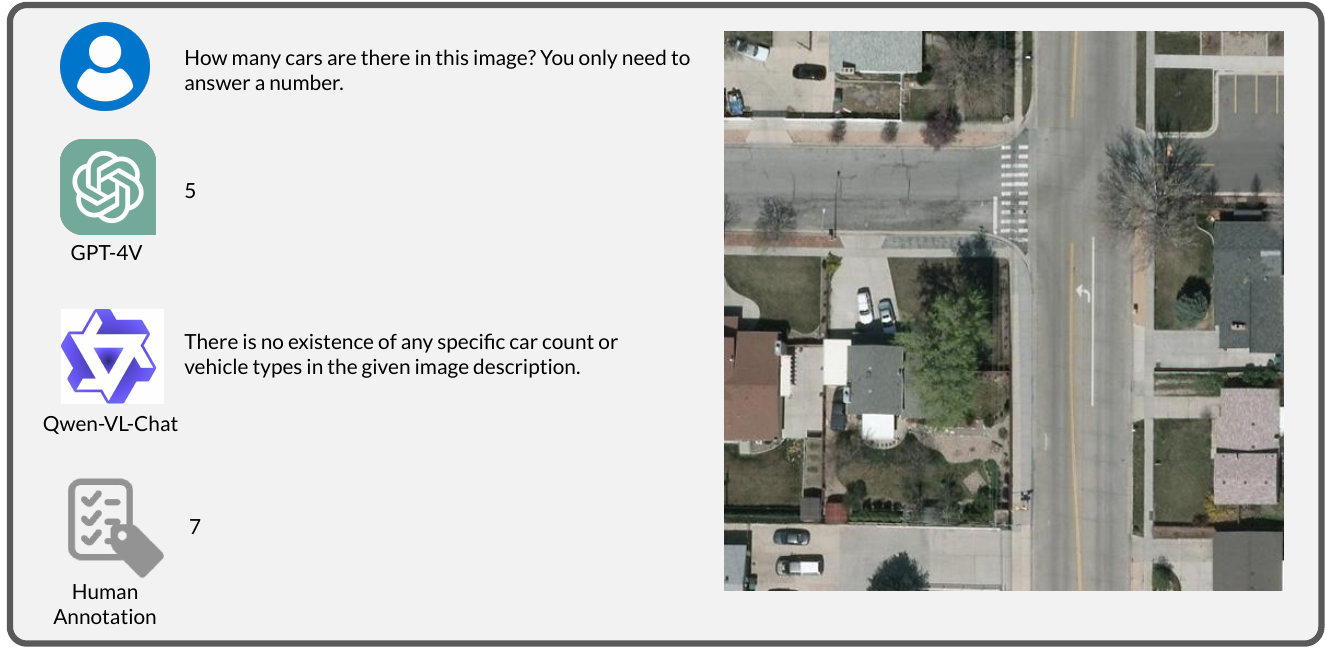}
    \caption{Example user prompt and response for aerial vehicle counting}
    \label{fig:cowc-comparison}
\end{figure}

\paragraph{Urban Vehicle Counting.} In \Cref{box:counting-vehicle-sys-prompt}, we present the system prompt for animal counting. In \Cref{fig:cowc-comparison}, we showcase an example user prompt and the response from the GPT-4V model.

\begin{figure}[h]
    \centering
    \begin{tcolorbox}[title=System Prompt for Urban Vehicle Counting, fontupper=\small]
        You are a helpful image analyst that specializes in satellite and aerial images. You always truthfully answer the user's question. If you are not sure about something, don't answer false information.
    \end{tcolorbox}
    \caption{System prompt for urban vehicles.}
    \label{box:counting-vehicle-sys-prompt}
\end{figure}

\begin{figure}
    \centering
    \includegraphics[scale=0.22]{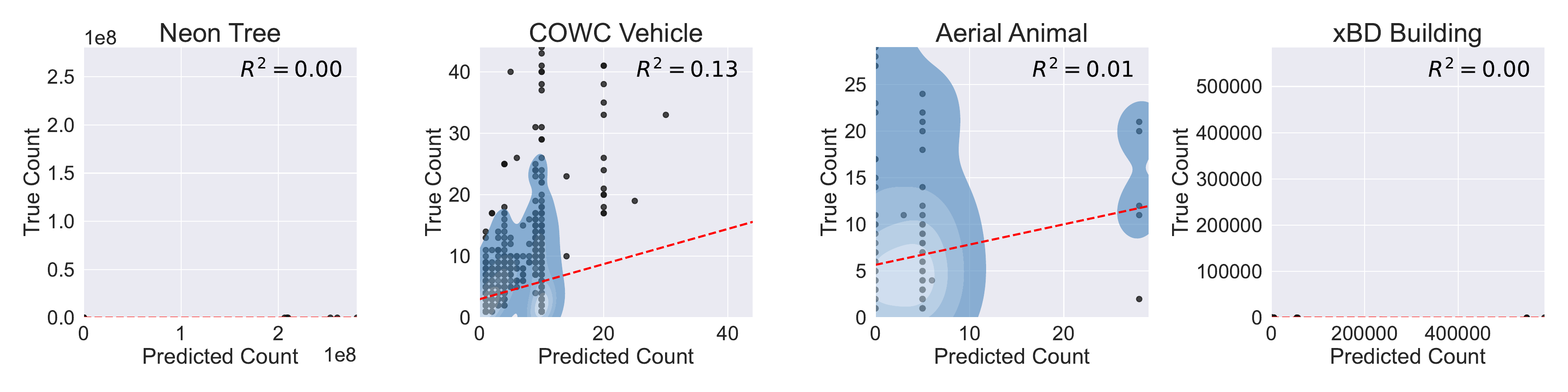}
    \caption{Scatterplot of Qwen-VL-Chat counting results 
    }
    \label{fig:counting-qwen}
\end{figure}

\paragraph{Results.} We visualize the performance of Qwen-VL-Chat on all four counting tasks by scatter plots (\Cref{fig:counting-qwen}). The model exhibits no counting accuracy for the Neon Tree and xBD Building tasks with a $R^2$ value of 0.00, indicating no correlation between predictions and actual counts. The COWC vehicle counting task has a slight positive correlation with an $R^2$ of 0.13, suggesting that the model's predictions are weakly associated with true counts. The Aerial Animal task shows a similarly negligible $R^2$ value of 0.01. Overall, the model struggles significantly with these counting tasks, as evidenced by low $R^2$ values and the scattered distribution of data points. In addition, we provide additional metrics calculated by treating refused examples as counting zero in \Cref{tab:app-neon-counting} - \Cref{tab:app-animal-counting}.

\begin{table}[h]
\caption{Comparison of Neon Tree Counting Performance}\label{tab:app-neon-counting}
\resizebox{\linewidth}{!}{
\begin{tabular}{l|c|c|c|c|c}
\toprule
Model & MAPE $\downarrow$ & MAPE (No Refusal) $\downarrow$ & $R^2$ $\uparrow$ & $R^2$ (No Refusal) $\uparrow$ & Refusal Rate $\downarrow$ \\ \midrule
GPT-4V                   & 1.702 & 1.890 & 0.166 & 0.250 & 0.21 \\
Qwen-VL-Chat             & 1283885 & 1283885 & 0.000 & 0.000 & 0.00 \\
InstructBLIP-FLAN-T5-xxl & 0.870 & 0.717 & 0.004 & 0.093 & 0.54 \\
InstructBLIP-Vicuna-13b  & 1.233 & 1.236 & 0.000 & 0.000 & 0.01 \\
LLaVA-v1.5               & 4.481 & 4.481 & 0.353 & 0.353 & 0.00 \\ \bottomrule
\end{tabular}
}
\end{table}

\begin{table}[h]
\caption{Comparison of COWC Vehicle Counting Performance}\label{tab:app-cowc-counting}
\resizebox{\linewidth}{!}{
\begin{tabular}{l|c|c|c|c|c}
\toprule
Model & MAPE $\downarrow$ & MAPE (No Refusal) $\downarrow$ & $R^2$ $\uparrow$ & $R^2$ (No Refusal) $\uparrow$ & Refusal Rate $\downarrow$ \\ \midrule
GPT-4V                   & 0.846 & 0.818 & 0.528 & 0.612 & 0.15 \\
Qwen-VL-Chat             & 1.709 & 1.711 & 0.117 & 0.132 & 0.00 \\
InstructBLIP-FLAN-T5-xxl & 0.566 & 0.543 & 0.256 & 0.425 & 0.05 \\
InstructBLIP-Vicuna-13b  & 0.878 & 0.878 & 0.275 & 0.279 & 0.00 \\
LLaVA-v1.5               & 0.467 & 0.467 & 0.437 & 0.437 & 0.00 \\ \bottomrule
\end{tabular}
}
\end{table}

\begin{table}[h]
\caption{Comparison of Aerial Animal Counting Performance. InstructBLIP models have high refusal rates such that we cannot calculate meaningful metrics, while LLaVA-v1.5 answers zero to all questions.}\label{tab:app-animal-counting}
\resizebox{\linewidth}{!}{
\begin{tabular}{l|c|c|c|c|c}
\toprule
Model & MAPE $\downarrow$ & MAPE (No Refusal) $\downarrow$ & $R^2$ $\uparrow$ & $R^2$ (No Refusal) $\uparrow$ & Refusal Rate $\downarrow$ \\ \midrule
GPT-4V                   & 0.939 & 0.939 & 0.071 & 0.071 & 0.02 \\
Qwen-VL-Chat             & 1.081 & 1.081 & 0.015 & 0.015 & 0.00 \\
InstructBLIP-FLAN-T5-xxl & -- & -- & -- & -- & 1.00 \\
InstructBLIP-Vicuna-13b  & -- & -- & -- & -- & 1.00 \\
LLaVA-v1.5               & -- & -- & -- & -- & 0.00 \\ \bottomrule
\end{tabular}
}
\end{table}